\documentclass[10pt,twocolumn,letterpaper]{article}

\usepackage{cvpr}              

\usepackage[utf8]{inputenc} 
\usepackage[T1]{fontenc}    
\usepackage{hyperref}       
\usepackage{url}            
\usepackage{amsfonts}       
\usepackage{nicefrac}       
\usepackage{microtype}      

\usepackage{enumitem}
\usepackage{xspace}

\usepackage{subcaption}

\usepackage{multirow}

\usepackage{graphicx}
\usepackage{subcaption}

\usepackage{wrapfig}
\usepackage{adjustbox}

\usepackage{epsfig}
\usepackage{pgfplots}
    \pgfplotsset{compat=1.18}
\usepackage{caption}
\usepgfplotslibrary{groupplots}

\usepackage{bm}
\usepackage{bbding}

\usepackage{pifont}
\usepackage{pgfplots}
\usepackage{algorithm}
\usepackage[noend]{algpseudocode}
\usepackage{caption} 
\usepackage{arydshln}

\usepackage{soul}

\usepackage[accsupp]{axessibility}  

\usepackage{multirow}

\usepackage{colortbl}

\usepackage{orcidlink}
\usepackage{circledsteps}
\usepackage{pict2e}
\usepackage{wrapfig,lipsum,booktabs}
\newsavebox{\bigimage}

\usepackage{layouts}
\usepackage{listings}
\usepackage{pifont}
\usepackage{wasysym}

\usepackage{tkz-kiviat}
\usepackage{xfp}

\usepackage{array,multirow,graphicx}
\usepackage{wrapfig}
\definecolor{purple}{rgb}{1,0,1}
\newcolumntype{a}{>{\columncolor{lightblue}}c}
\newcommand{\kibitz}[2]{\ifnum\Comments=1\textcolor{#1}{#2}\fi}

\usepackage{tkz-kiviat}
\usepackage{xfp}

\usepackage{dblfloatfix}
\usepackage{epsfig}
\usepackage{pgfplots}
    \pgfplotsset{compat=1.18}
\usepackage{caption}
\usepgfplotslibrary{groupplots}

\usepackage{array,multirow,graphicx}
\usepackage{wrapfig}
\definecolor{purple}{rgb}{1,0,1}

\usepackage{tkz-kiviat}
\usepackage{xfp}
\usepackage{multirow}

\usepackage{epsfig}
\usepackage{pgfplots}
    \pgfplotsset{compat=1.18}
\usepackage{caption}
\usepgfplotslibrary{groupplots}

\definecolor{demphcolor}{RGB}{144,144,144}
\newcommand{\demph}[1]{\textcolor{demphcolor}{#1}}

\newcommand{\customsubsection}[1]{%
  \par
  \pagebreak[2]%
  \refstepcounter{subsection}%
    \everypar={%
      {\setbox0=\lastbox}
      \addcontentsline{toc}{subsection}{%
        {\protect\makebox[0.3in][r]{\thesubsubsection.} \hspace*{3pt}#1}}%
      \textbf{\thesubsection\space\space{#1}\space\newline}%
      \everypar={}%
    }%
  \ignorespaces
}

\usepackage{pifont}
\usepackage{algorithm}
\usepackage{algpseudocode}

\usepackage{lipsum}

\definecolor{cvprblue}{rgb}{0.21,0.49,0.74}
\definecolor{babyblueeyes}{rgb}{0.63, 0.79, 0.95}
\definecolor{darkturquoise}{rgb}{0.0, 0.81, 0.82}
\definecolor{deepskyblue}{rgb}{0.0, 0.75, 1.0}
\definecolor{dodgerblue}{rgb}{0.12, 0.56, 1.0}
\definecolor{turquoiseblue}{rgb}{0.0, 1.0, 0.94}

\hypersetup{
    colorlinks,
    linkcolor={red},
    citecolor={deepskyblue}
}
\usepackage{pifont}
\usepackage{algorithm}
\usepackage{algpseudocode}

\usepackage{caption}
\usepackage{cuted} 

\definecolor{Light}{HTML}{f6fae4}
\definecolor{Light}{HTML}{fafced}
\definecolor{Light}{HTML}{f8fbe9}

\definecolor{6ec2b5}{HTML}{b8d65e}
\definecolor{FFABA8}{HTML}{807d7d}
\colorlet{Dark}{6ec2b5}
\colorlet{Salmon}{FFABA8}

\definecolor{forestg}{HTML}{34bf34}
\colorlet{ForestGreen}{forestg}
\colorlet{OrangeRed}{red}

\definecolor{citecolor}{HTML}{0071bc}
\definecolor{color_ao}{gray}{0.5}
\definecolor{color_our}{HTML}{e6f2c2}
\definecolor{color_pre}{rgb}{0.52,0.59,0.69}
\definecolor{Gray}{gray}{0.9}
\definecolor{LighterGray}{gray}{0.93}
\definecolor{LightGrayForTableRule}{gray}{0.92}
\definecolor{DarkGray}{gray}{0.5}
\definecolor{Black}{rgb}{0.0, 0.0, 0.0}
\definecolor{NiceBlue}{rgb}{0.11764705882352941, 0.5647058823529412, 1.0}
\definecolor{NiceGreen}{HTML}{b6c783}
\usepackage{diagbox}
\definecolor{gray}{HTML}{b0aeae}
\definecolor{diffgray}{HTML}{878787}
\definecolor{NiceGray}{HTML}{696969}
\definecolor{applegreen}{rgb}{0.55, 0.71, 0.0}

\definecolor{F08080}{HTML}{F08080}
\colorlet{LinePlot1}{F08080}

\definecolor{D2B48C}{HTML}{D2B48C}
\colorlet{LinePlot2}{D2B48C}

\definecolor{B8D65E}{HTML}{B8D65E}
\colorlet{LinePlot3}{B8D65E}

\definecolor{b5b3b3}{HTML}{b5b3b3}
\colorlet{LinePlot4}{b5b3b3}

\newcommand{\CC}[1]{\cellcolor{Light}}

\definecolor{d8d9ed}{HTML}{d8d9ed}
\definecolor{f5f8ff}{HTML}{f5f8ff}
\colorlet{ThemeColor}{f5f8ff}

\definecolor{increase}{HTML}{FCE4D6}

\def\ourbenchmark{\textsc{AVTrustBench}\xspace}

\def\adversarialsuite{Adversarial attack\xspace}
\def\compositionalsuite{Compositional reasoning\xspace}
\def\missingmodalitysuite{Modality-specific dependency\xspace}
\def\adversarialaad{MCIT\xspace}
\def\adversarialiasd{ICIT\xspace}
\def\adversarialivqd{MVIT\xspace}
\def\adversarialiaqd{MAIT\xspace}
\def\compositionalstitching{COT-Stitch\xspace}
\def\compositionalswapping{COT-Swap\xspace}
\def\compositionalattribute{CAT\xspace}
\def\missingaudio{MAT\xspace}
\def\missingvideo{MVT\xspace}
\def\totalbenchmarksize{600K\xspace}
\def\totalbenchmarktestsize{181K\xspace}
\def\totaltaskcount{9\xspace}

\usepackage{tcolorbox}
\newtcolorbox{abox}{colback=TakeAwayColor6,colframe=Black}

\definecolor{97C8DB}{HTML}{97C8DB}
\colorlet{TakeAwayColor1}{97C8DB}

\definecolor{AECEE8}{HTML}{AECEE8}
\colorlet{TakeAwayColor2}{AECEE8}

\definecolor{4B9CDE}{HTML}{4B9CDE}
\colorlet{TakeAwayColor3}{4B9CDE}

\definecolor{256bdb}{HTML}{256bdb}
\colorlet{TakeAwayColor4}{256bdb}

\definecolor{D1D8E3}{HTML}{D1D8E3}
\colorlet{TakeAwayColor5}{D1D8E3}

\definecolor{C8C8CC}{HTML}{C8C8CC}
\colorlet{TakeAwayColor6}{C8C8CC}

\definecolor{D4D3DB}{HTML}{D4D3DB}
\colorlet{TakeAwayColor7}{D4D3DB}

\definecolor{e5e1fa}{HTML}{e5e1fa}
\colorlet{Adv}{e5e1fa}

\definecolor{faf0f4}{HTML}{faf0f4}
\colorlet{Comp}{faf0f4}

\definecolor{ebf7eb}{HTML}{ebf7eb}
\colorlet{Mis}{ebf7eb}

\definecolor{nicegreen}{rgb}{0.1, 0.6, 0.2}

\usepackage{circledsteps}
\usepackage{pict2e}
\usepackage{wrapfig,lipsum,booktabs}

\usepackage{tkz-kiviat}
\usepackage{xfp}

\usepackage{array,multirow,graphicx}
\usepackage{wrapfig}
\definecolor{purple}{rgb}{1,0,1}

\newcommand\blfootnote[1]{%
  \begingroup
  \renewcommand\thefootnote{}\footnote{#1}%
  \addtocounter{footnote}{-1}%
  \endgroup
}

\usepackage{tkz-kiviat}
\usepackage{xfp}

\usepackage{dblfloatfix}
\usepackage{epsfig}
\usepackage{pgfplots}
    \pgfplotsset{compat=1.18}
\usepackage{caption}
\usepgfplotslibrary{groupplots}

\usepackage{array,multirow,graphicx}
\usepackage{wrapfig}
\definecolor{purple}{rgb}{1,0,1}

\usepackage{tkz-kiviat}
\usepackage{xfp}
\usepackage{multirow}

\usepackage{epsfig}
\usepackage{pgfplots}
    \pgfplotsset{compat=1.18}
\usepackage{caption}
\usepgfplotslibrary{groupplots}

\definecolor{demphcolor}{RGB}{144,144,144}

\usepackage{lipsum}

\definecolor{cvprblue}{rgb}{0.21,0.49,0.74}
\definecolor{babyblueeyes}{rgb}{0.63, 0.79, 0.95}
\definecolor{darkturquoise}{rgb}{0.0, 0.81, 0.82}
\definecolor{deepskyblue}{rgb}{0.0, 0.75, 1.0}
\definecolor{dodgerblue}{rgb}{0.12, 0.56, 1.0}
\definecolor{turquoiseblue}{rgb}{0.0, 1.0, 0.94}

\hypersetup{
    colorlinks,
    linkcolor={red},
    citecolor={deepskyblue}
}
\usepackage{pifont}
\usepackage{algorithm}
\usepackage{algpseudocode}
\definecolor{Light}{HTML}{f6fae4}
\definecolor{Light}{HTML}{fafced}
\definecolor{Light}{HTML}{f8fbe9}

\definecolor{6ec2b5}{HTML}{b8d65e}
\definecolor{FFABA8}{HTML}{807d7d}
\colorlet{Dark}{6ec2b5}
\colorlet{Salmon}{FFABA8}

\definecolor{forestg}{HTML}{34bf34}
\colorlet{ForestGreen}{forestg}
\colorlet{OrangeRed}{red}

\definecolor{citecolor}{HTML}{0071bc}
\definecolor{color_ao}{gray}{0.5}
\definecolor{color_our}{HTML}{e6f2c2}
\definecolor{color_pre}{rgb}{0.52,0.59,0.69}
\definecolor{Gray}{gray}{0.9}
\definecolor{LighterGray}{gray}{0.93}
\definecolor{LightGrayForTableRule}{gray}{0.92}
\definecolor{DarkGray}{gray}{0.5}
\definecolor{Black}{rgb}{0.0, 0.0, 0.0}
\definecolor{NiceBlue}{rgb}{0.11764705882352941, 0.5647058823529412, 1.0}
\definecolor{NiceGreen}{HTML}{b6c783}
\usepackage{diagbox}
\definecolor{gray}{HTML}{b0aeae}
\definecolor{diffgray}{HTML}{878787}
\definecolor{NiceGray}{HTML}{696969}
\definecolor{applegreen}{rgb}{0.55, 0.71, 0.0}

\definecolor{F08080}{HTML}{F08080}
\colorlet{LinePlot1}{F08080}

\definecolor{D2B48C}{HTML}{D2B48C}
\colorlet{LinePlot2}{D2B48C}

\definecolor{B8D65E}{HTML}{B8D65E}
\colorlet{LinePlot3}{B8D65E}

\definecolor{b5b3b3}{HTML}{b5b3b3}
\colorlet{LinePlot4}{b5b3b3}

\definecolor{cvprblue}{rgb}{0.21,0.49,0.74}

\def\paperID{13976} 
\def\confName{CVPR}
\def\confYear{2025}

\title{\ourbenchmark: Assessing and Enhancing Reliability and Robustness in Audio-Visual LLMs}

\author{Sanjoy Chowdhury$^{*1}$ $\quad$ Sayan Nag$^{*2}$ $\quad$ Subhrajyoti Dasgupta$^3$ \\ Yaoting Wang$^4$ $\quad$ Mohamed Elhoseiny$^4$$^\dagger$ $\quad$ Ruohan Gao$^1$$^\dagger$ $\quad$ Dinesh Manocha$^1$$^\dagger$ 
\vspace{1mm} 
\\ 
$^{1}$University of Maryland, College Park $\quad$ $^{2}$University of Toronto  
\\
 $^{3}$Mila and Université de Montréal $\quad$ $^{4}$KAUST 
 \vspace{-0.5mm}
 \\ 
\tt\footnotesize \{sanjoyc, rhgao, dmanocha\}@umd.edu $\quad$ sayan.nag@mail.utoronto.ca 
$\quad$ 
 \tt\footnotesize subhrajyoti.dasgupta@umontreal.ca 
}

\makeatletter

\begin{document}



\twocolumn[{%
\renewcommand\twocolumn[1][]{#1}%
\maketitle
\begin{center}
    \centering
    \captionsetup{type=figure}
    \vspace{-2mm}
    \includegraphics[width=0.97\textwidth]{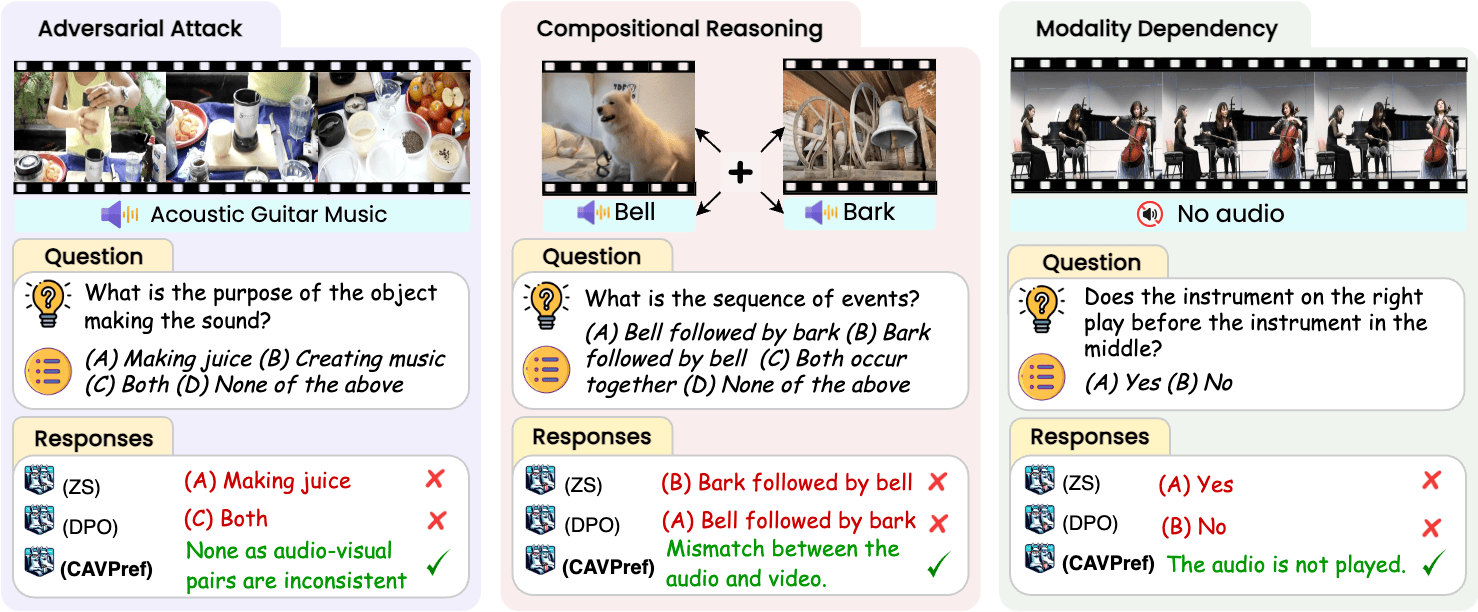}
    \captionof{figure}{\textbf{Introducing \ourbenchmark and CAVPref.} We present \ourbenchmark, a new benchmark comprising three challenging yet unexplored axes, i.e., \colorbox{Adv}{Adversarial Attack}, \colorbox{Comp}{Compositional Reasoning}, and \colorbox{Mis}{Modality Dependency}, and evaluate SOTA Audio-Visual LLMs (AVLLMs) on this benchmark. We observe that these models demonstrate poor performances under these settings. To alleviate these limitations, we propose a novel AVLLM-agnostic preference optimization strategy \textbf{CAVPref}, which substantially improves the reliability and robustness of these models over existing solutions such as DPO. 
    \raisebox{-0.7mm}
    {\includegraphics[height=3.2mm]{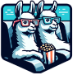}}
    : VideoLLaMA2 model.}
    \label{fig:main_teaser}
\end{center}%
}]




\begin{abstract}

\vspace{-0.06in}

\blfootnote{$^*$Equal contribution. $^\dagger$Equal advising.}

With the rapid advancement of Multi-modal Large Language Models (MLLMs), several diagnostic benchmarks have recently been developed to assess these models' multi-modal reasoning proficiency. However, these benchmarks are restricted to assessing primarily the visual aspect and do not examine the holistic audio-visual (AV) understanding. Moreover, currently, there are no benchmarks that investigate the capabilities of AVLLMs to calibrate their responses when presented with perturbed inputs. To this end, we introduce \textbf{A}udio-\textbf{V}isual \textbf{Trust}worthiness assessment \textbf{Bench}mark (\textbf{\ourbenchmark}), comprising \textbf{\totalbenchmarksize} samples spanning over \textbf{\totaltaskcount} meticulously crafted tasks, evaluating the capabilities of AVLLMs across three distinct dimensions: \textit{\adversarialsuite}, \textit{\compositionalsuite}, and \textit{\missingmodalitysuite}. Using our benchmark, we extensively evaluate \textbf{13} state-of-the-art AVLLMs. The findings reveal that the majority of existing models fall significantly short of achieving human-like comprehension, offering valuable insights for future research directions. To alleviate the limitations in the existing approaches, we further propose a robust, model-agnostic calibrated audio-visual preference optimization-based training strategy \textbf{CAVPref}, obtaining a gain up to \textit{30.19\%} across all 9 tasks. We will publicly release our code and benchmark to facilitate future research in this direction.





\end{abstract}

\begin{table*}[!t]
\centering
\renewcommand{\arraystretch}{1.0}
\resizebox{\textwidth}{!}{%
\begin{tabular}{l|c|c|c|c|c|c|c|c|c}
\hline

\multirow{2}{*}{\textbf{Benchmark}} &
\textbf{Visual} &
\textbf{Benchmark} &
\textbf{Answer} &
\textbf{Evaluation Type} &
\textbf{Temporal} &
\multirow{2}{*}{\textbf{Adversarial?}} &
\multirow{2}{*}{\textbf{Compositionality?}} &
\textbf{Modality} &
\textbf{Audio-visual} \\

 &
\textbf{modality} &
\textbf{size} &
\textbf{Type} &
\textbf{(Human / GPT)} &
\textbf{order?} &
 &
 &
\textbf{dependency?} &
\textbf{reasoning?} \\

\hline
MVBench \cite{mvbench} &
  Image + Video &
  1.9M & 
  MCQ &
  GPT &
  \textcolor{ForestGreen}{\ding{51}} &
  \textcolor{OrangeRed}{\ding{55}} &
  \textcolor{OrangeRed}{\ding{55}} &
  \textcolor{OrangeRed}{\ding{55}} &
  \textcolor{OrangeRed}{\ding{55}} \\
SEED-bench \cite{seedbench} &
  Image + Video &
  19K & 
  MCQ &
  Heuristics-based &
  \textcolor{ForestGreen}{\ding{51}} &
  \textcolor{OrangeRed}{\ding{55}} &
  \textcolor{OrangeRed}{\ding{55}} &
  \textcolor{OrangeRed}{\ding{55}} &
  \textcolor{OrangeRed}{\ding{55}} \\
MMBench \cite{mmbench} &
  Image &
  3.2K & 
  Free-form &
  GPT &
  \textcolor{OrangeRed}{\ding{55}} &
  \textcolor{OrangeRed}{\ding{55}} &
  \textcolor{OrangeRed}{\ding{55}} &
  \textcolor{OrangeRed}{\ding{55}} &
  \textcolor{OrangeRed}{\ding{55}} \\
LVLM-eHub \cite{lvlmehub} &
  Image &
  -- & 
  Free-form &
  Human &
  \textcolor{OrangeRed}{\ding{55}} &
  \textcolor{OrangeRed}{\ding{55}} &
  \textcolor{OrangeRed}{\ding{55}} &
  \textcolor{OrangeRed}{\ding{55}} &
  \textcolor{OrangeRed}{\ding{55}} \\
LAMM \cite{lamm} &
  Image + Point-cloud &
  186K & 
  Free-form &
  GPT &
  \textcolor{OrangeRed}{\ding{55}} &
  \textcolor{OrangeRed}{\ding{55}} &
  \textcolor{OrangeRed}{\ding{55}} &
  \textcolor{OrangeRed}{\ding{55}} &
  \textcolor{OrangeRed}{\ding{55}} \\
MME \cite{videollmsurvey} &
  Image &
  -- & 
  Y/N &
  -- &
  \textcolor{OrangeRed}{\ding{55}} &
  \textcolor{OrangeRed}{\ding{55}} &
  \textcolor{OrangeRed}{\ding{55}} &
  \textcolor{OrangeRed}{\ding{55}} &
  \textcolor{OrangeRed}{\ding{55}} \\
Video-Bench \cite{videobench} &
  Video &
  15K & 
  MCQ &
  GPT &
  \textcolor{ForestGreen}{\ding{51}} &
  \textcolor{OrangeRed}{\ding{55}} &
  \textcolor{OrangeRed}{\ding{55}} &
  \textcolor{OrangeRed}{\ding{55}} &
  \textcolor{OrangeRed}{\ding{55}} \\
  HallusionBench \cite{hallusionbench} &
  Image &
  1.1K & 
  Free-form &
  GPT &
  \textcolor{OrangeRed}{\ding{55}} &
  \textcolor{OrangeRed}{\ding{55}} &
  \textcolor{OrangeRed}{\ding{55}} &
  \textcolor{OrangeRed}{\ding{55}} &
  \textcolor{OrangeRed}{\ding{55}} \\
\rowcolor{ThemeColor} 
\textbf{\ourbenchmark\ (ours)} &
  Audio + Video &
  \textbf{\totalbenchmarksize} & 
  MCQ &
  Heuristics + GPT &
  \textcolor{ForestGreen}{\ding{51}} &
  \textcolor{ForestGreen}{\ding{51}} &
  \textcolor{ForestGreen}{\ding{51}} &
  \textcolor{ForestGreen}{\ding{51}} &
  \textcolor{ForestGreen}{\ding{51}} \\
  \hline
\end{tabular}%
}
\vspace{-0.05in}
\caption{\textbf{Comparison with existing benchmarks for MLLMs}. \ourbenchmark is the first to study the robustness and reliability of AVLLMs under 3 critical yet unexplored dimensions: \textit{\adversarialsuite}, \textit{\compositionalsuite}, \textit{\missingmodalitysuite}.}
\label{tab:comparison_overview}
\vspace{-2mm}
\end{table*}


\section{Introduction}
\label{introduction}



In recent years, Large Language Models (LLMs) \cite{chung2024scaling, gpt4, llama, llama2} have demonstrated remarkable capabilities to understand, reason, and generate text across a variety of tasks. Leveraging LLMs, recent efforts extend to other modalities beyond text (e.g., image, video, audio, etc.) through Multi-modal Large Language Models (MLLMs) \cite{blip2, minigpt4, llava, mplugowl, instructblip, pandagpt, kosmos, vistallm, shikra, minigptv2, gpt4roi, videochat, videochatgpt, valley, macawllm, gemini}. However, with the introduction of these more powerful models comes the increasing need of assessing the reliability and robustness of their output when deployed in real-world settings. While we humans can easily identify the discrepancies and act accordingly when encountering a ``wrong'' question, in most cases, current AVLLMs assume the validity of the question and have a propensity towards responding with a hallucinated answer.

Of late, a number of benchmarks have been proposed \cite{videollmsurvey, li2023evaluating, mmbench, lvlmehub, mmvet, seedbench} to evaluate MLLMs under a typical Question-Answer (QA) set-up (free form or multiple-choice) to investigate its performance under various reasoning and perception tasks. We identify two major limitations in the existing benchmarks: \textit{(i)} current benchmarks are primarily restricted to the visual modality and \textit{ignore} other modalities such as `audio', an extremely critical component in comprehensive video understanding; \textit{(ii)} existing benchmarks \textit{do not evaluate the reliability and robustness} of AVLLMs' response under critical aspects such as adversarial attack, compositional understanding capabilities, and their ability to extract synchronous information from the constituent modalities.



Recent works \cite{videollmsurvey, lamm, lvlmehub, mmbench} develop benchmarks to evaluate MLLMs for images and videos as shown in Tab.~\ref{tab:comparison_overview}. LVLM-eHub \cite{lvlmehub} and LAMM \cite{lamm} employ human annotators to assess the model's performance. This introduces subjectivity and compromises efficiency. MME \cite{videollmsurvey} and MMBench \cite{mmbench} improve objective evaluation of MLLMs by constructing True / False or Multiple-Choice questions. Restricting the model's output to a fixed set of options enables convenient and near-accurate evaluation protocol. However, the relatively small scale of these benchmarks (less than 3.5K samples) results in incomprehensive evaluation. These limitations reveal the need of an automated and comprehensive benchmark for the assessment of AVLLMs.

To this end, we present \ourbenchmark, a multi-dimensional benchmark suite to extensively evaluate AVLLMs  (Fig. \ref{fig:data_stats}). The benchmark comprises \textbf{\totalbenchmarksize} samples spanning over \textbf{\totaltaskcount} tasks to evaluate the audio-visual comprehension capabilities in AVLLMs. We design a semi-automatic annotation paradigm to generate multiple-choice QAs for each task by adapting public audio-visual datasets, making it cost-efficient in terms of human annotations and more objective compared to prior work. Using \ourbenchmark, we make a thorough evaluation of 13 state-of-the-art AVLLMs (11 open and 2 closed source) and present useful findings about them based on their performances. 
Additionally, we provide valuable insights for future work to improve the robustness and reasoning capabilities of these models.

To address the limitations of existing AVLLMs, we further propose a new model-agnostic training strategy---\textbf{CAVPref}, comprising of a calibrated AV preference optimization protocol with a robustness module. As opposed to state-of-the-art preference optimization models \cite{rafailov2024direct} (which favors text over other multi-modal information, leading to multi-modal hallucinations \cite{sarkar2024mitigating}), \textbf{CAVPref}, in its formulation, involves conditioning from all the multi-modal inputs (audio, video, text), thereby improving reliability of the AVLLMs (Fig. \ref{fig:main_teaser}). Furthermore, the robustness module renders the AVLLMs impervious to the distributional shifts present in the multi-modal preference datasets and thereby improve performances of AVLLMs across underrepresented categories (without compromising on other categories). 

To summarize, our \textbf{main contributions} are as follows:


       


\begin{figure*}
    \centering
    \includegraphics[width=0.99\textwidth]{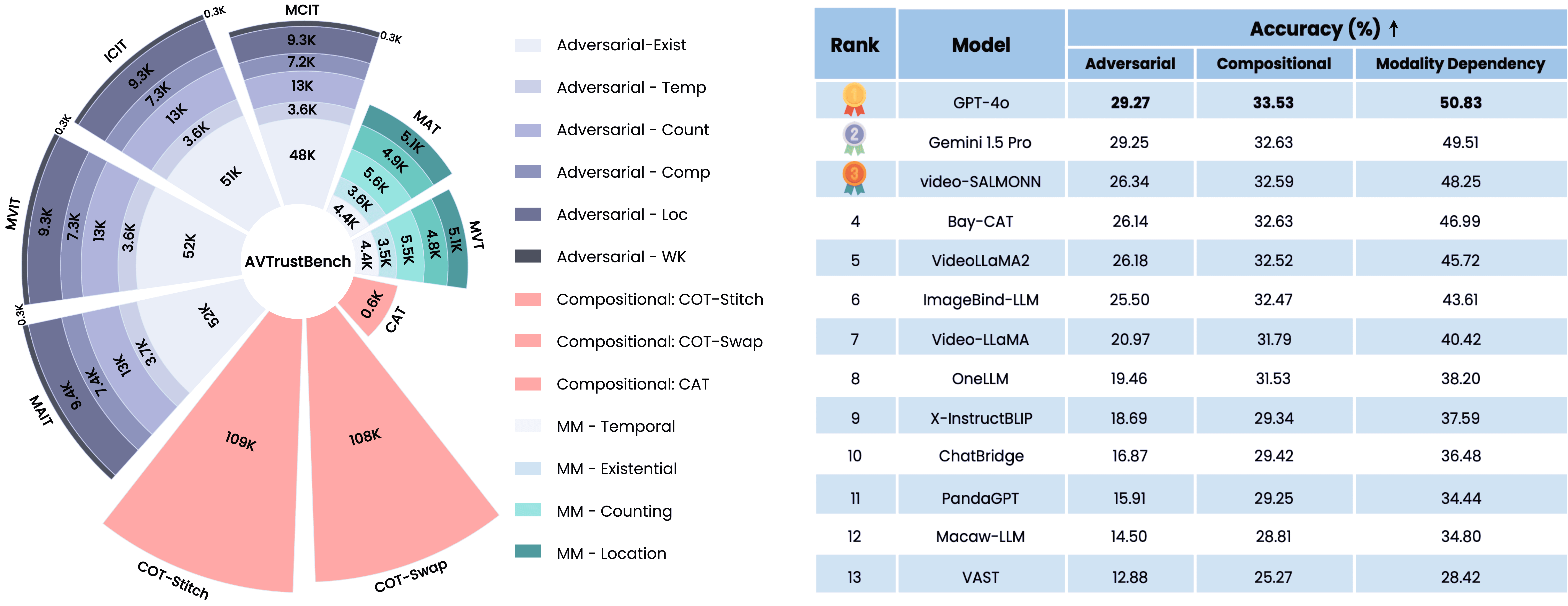}
   \vspace{-0.1in}
    \caption{\textbf{\ourbenchmark statistics and AVLLMs leaderboard.} (Left) Task-wise data distribution. Our benchmark comprises 9 diverse tasks spanning over 3 dimensions. (Right) Performance comparison on \ourbenchmark. Values represent dimension-wise averages.}
    \label{fig:data_stats}

\end{figure*}




\noindent{\textbf{(1)} \textit{We introduce \ourbenchmark}, the first comprehensive audio-visual benchmark that assesses the trustworthiness of AVLLMs. 
It evaluates existing AVLLMs under \textit{three} critical dimensions:  \textit{\adversarialsuite}, \textit{\compositionalsuite}, and \textit{\missingmodalitysuite}.}    



\noindent{\textbf{(2)} We \textit{extensively evaluate 13 state-of-the-art AVLLMs} under our benchmark, 
uncovering their major limitations and sharing our key observations on their performance.}

\noindent{\textbf{(3)} We \textit{introduce a novel model-agnostic training strategy}---\textbf{CAVPref}, comprising of a calibrated AV preference optimization with a robustness module. Our proposed approach achieves up to \textit{30.19\%} improvement across all 9 tasks.}
\section{Related Work}

\noindent\textbf{Building Multi-modal LLMs.} Inspired by the success of large language models \cite{instructgpt, vicuna, alpaca}, recent work has expanded LLMs to multi-modal understanding, leveraging high-quality multi-modal instructional data \cite{minigpt4, llava, otter, pandagpt, xllm, kosmos, shikra, vistallm, gpt4roi, flamingo, llamaadapter}. Video-LLMs \cite{macawllm, xinstructblip, pandagpt, videollama, onellm, timechat, vast} extend LLMs \cite{llama, llama2} and image-based LLMs \cite{flamingo, openflamingo, llava, florence} to handle additional modalities such as audio and subtitles. ChatBridge \cite{chatbridge} uses Perceiver \cite{perceiver} for modality alignment with LLMs, while PandaGPT and ImageBind-LLM \cite{imagebind, imagebindllm} naturally integrate multi-modal inputs. X-LLM \cite{xllm} applies Q-Former with modality-specific adapters to combine image, audio, and video with LLMs, and Video-LLaMA \cite{videollama} incorporates temporal embeddings via ImageBind. Bay-CAT \cite{baycat} is a recent AVLLM trained with an ambiguity-aware DPO strategy. Despite these advancements, none of these studies on AVLLMs address the challenges of AV consistency.

\vspace{0.05in}

\noindent{\textbf{Evaluating Multi-modal LLMs.}} With rapid advances in multi-modal LLMs, various benchmarks \cite{videollmsurvey, lamm, lvlmehub, mmbench} have been proposed for their evaluation. GVT \cite{gvtmakes} combines semantic (VQA, image captioning) and fine-grained tasks (object counting), while LVLM-eHub \cite{lvlmehub} aggregates benchmarks using human annotation. LAMM \cite{lamm} evaluates open-form predictions on images and point clouds with GPT, though this LLM-based evaluation may affect reliability. MME \cite{videollmsurvey} and MMBench \cite{mmbench} introduce multiple-choice QAs across diverse dimensions. Other benchmarks like AI2 Reasoning \cite{clark2018think}, HellaSwag \cite{hellaswag}, MMLU \cite{hendrycks2020measuring}, and TruthfulQA \cite{truthfulqa} assess reasoning, knowledge, and misinformation. SEED-Bench \cite{seedbench} adds temporal tasks with a quality-assured pipeline. While some benchmarks \cite{seedbench, videochatgpt, funqa} evaluate MLLM's temporal perception, they either work on primitive video tasks \cite{seedbench} or focus on particular domains (e.g., funny clips \cite{funqa}), thereby limiting their practical applicability. Besides, they involve labor-intensive annotations which introduce subject bias and are cost-ineffective. Recently, VideoBench \cite{videobench} and HallusionBench \cite{hallusionbench} investigated decision-making capabilities and visual illusions for videos and images. To address these limitations, we present a \textit{comprehensive} benchmark to evaluate the \textit{trustworthiness} of MLLMs under \textit{audio-visual} events.

\vspace{0.05in}

\noindent{\textbf{Multi-modal Preference Optimization.}} Recent works in multimodal scenarios focus on creating multimodal preference data \cite{silkie, zhao2023beyond, xiao2024detecting, zhou2024aligning, pi2024strengthening, yu2024rlaif, deng2024enhancing}. These
efforts include collecting human preference \cite{sun2023aligning, yu2024rlhf}, preference from
advanced multimodal LLMs \cite{silkie, yu2024rlaif}, and preference from the model to align itself \cite{deng2024enhancing}. In terms of learning objectives, recent works mainly follow DPO for LLMs \cite{silkie, zhao2023beyond, zhou2024aligning}. Some also apply reinforcement learning \cite{sun2023aligning, jing2024fgaif} and contrastive learning \cite{sarkar2024mitigating, jiang2024hallucination}. However, preference optimization-based approaches disregard the importance of AV consistency, which we incorporate within our proposed objective.

\vspace{-0.05in}
\section{\ourbenchmark: Audio-Visual Trustworthiness Assessment Suite}
\label{benchmark section}

\customsubsection{\ourbenchmark Taxonomy and Task Definitions}
\label{taxonomy}



\begin{figure*}[!t]
    \centering
    \includegraphics[width=0.97\textwidth]{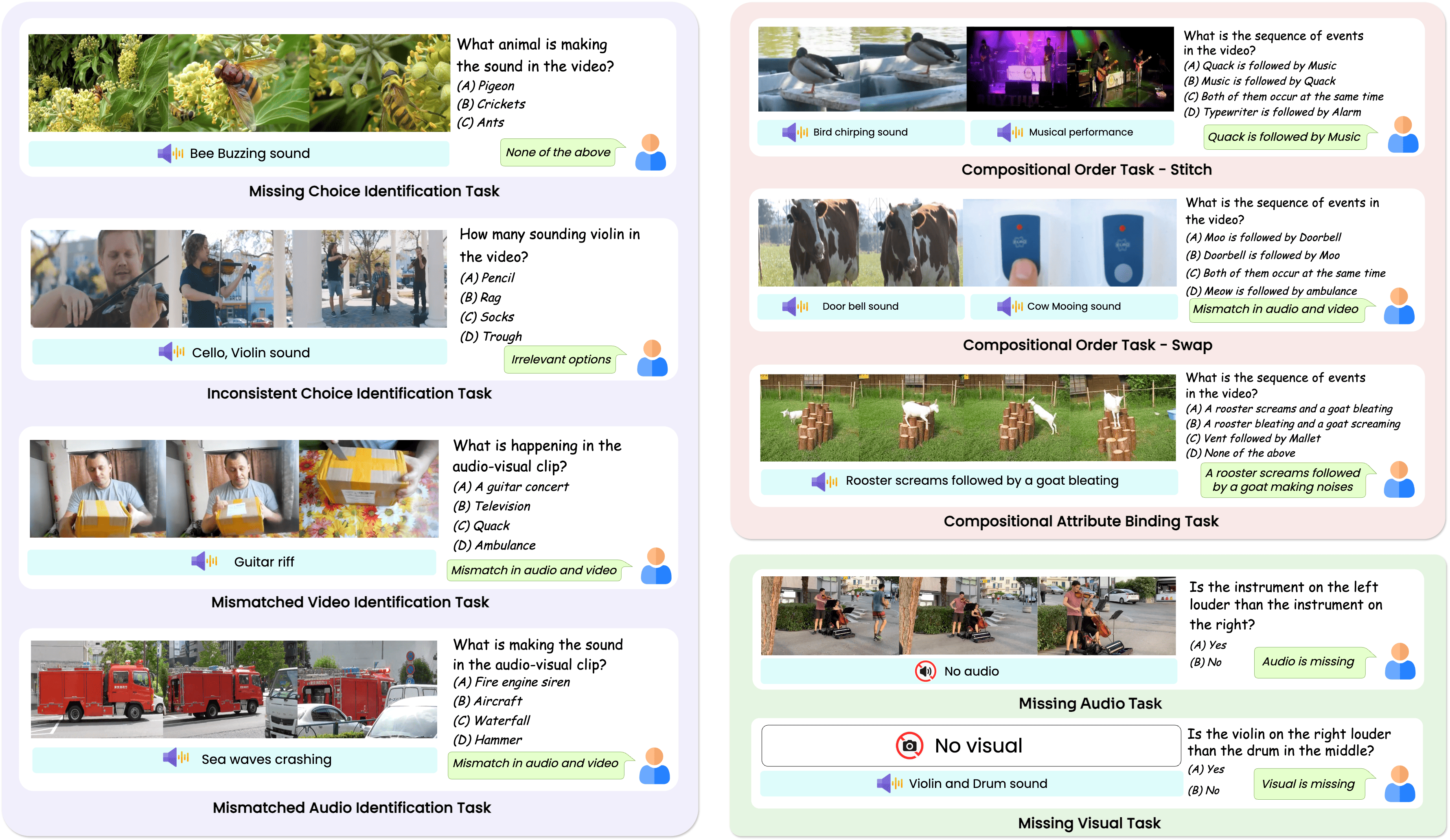}
    \vspace{-0.05in}
    \caption{\textbf{Task definitions:} \ourbenchmark comprises a total of \textbf{9 tasks} tasks MCIT, ICIT, MVIT and MAIT from \adversarialsuite, COT-Stitch, COT-Swap and CAT from \compositionalsuite and MAT and MVT from \missingmodalitysuite respectively. The goal of each dimension is to critically assess the robustness of existing AVLLMs under different modes of challenges. In each case, the AVLLMs are presented with a multiple-choice question setup. Refer to Sec. \ref{taxonomy} for task-specific details.}
    \label{fig:task_suites}
    \vspace{-4mm}
\end{figure*}

Our goal is to investigate the degree to which AVLLMs: \textit{accurately comprehend} the audio, visual, and textual inputs with correct semantics, \textit{rely} on individual modalities, and \textit{follow instructions}, even in the presence of inconsistencies in input signals. Accordingly, we design our study where we evaluate existing AVLLMs under \textbf{three} broad dimensions: \textbf{\adversarialsuite}, \textbf{\compositionalsuite}, and \textbf{\missingmodalitysuite}. Fig. \ref{fig:task_suites} depicts individual tasks with a representative example.

\vspace{0.02in}

\noindent\textbf{\adversarialsuite.} This suite comprises \textit{four} different tasks for evaluating AVLLMs' performance under adversarial problem settings. This collection of tasks either consists of incongruent audio-visual pairs or inconsistencies in the answer templates. \adversarialsuite\ suite includes:

\begin{itemize}[leftmargin=*]

\item\textbf{Missing Choice Identification Task (MCIT).} 
As the name suggests, this task analyzes whether the AVLLM can correctly discern that the appropriate answer is missing from the multiple-choice answer set. This task examines the model's capacity to restrain itself from responding with a choice from a plausible set of options when the correct choice is missing. Note in Fig. \ref{fig:task_suites} the model is presented with potential yet inaccurate options while asked to identify an audio-visual event. 

\item\textbf{Inconsistent Choice Identification Task (ICIT).} Unlike MCIT, in ICIT the answer set does not have any relevance to the question or audio-visual content. With entirely unrelated answer sets, ICIT assesses the extent of a model's propensity to force wrong answers with high confidence regardless of the semantic closeness to the provided choices.

\item\textbf{Mismatched Video Identification Task (MVIT).} MVIT assesses AVLLMs' ability to determine if a video and corresponding audio-question pairs are mismatched or incongruent. This evaluation examines the model's comprehension of the alignment between visual information with both textual (question + answer choices) and audio queries, with the objective of identifying cases where these combinations are incompatible. In Fig. \ref{fig:task_suites} the visual modality from the video of a man playing a guitar is replaced with a man unboxing a parcel. Despite one of the options in the answer set having `guitar', an intelligent system should ideally point out the inconsistency through its response.        

\item\textbf{Mismatched Audio Identification Task (MAIT).} Similar to MVIT, MAIT investigates the ability of AVLLMs to determine if the audio and corresponding visual + textual inputs are mismatched. The impractical example (Fig. \ref{fig:task_suites}) of a fire engine coupled with an audio track of pleasant sea waves with gulls squealing should trigger an ideal AVLLM to raise concern even in the presence of alluring options. 

\end{itemize}

\begin{table*}[!t]
\centering
\renewcommand{\arraystretch}{1.1}
\resizebox{\textwidth}{!}{%
\begin{tabular}{l | cccc | cccc | cccc | cccc}
\hline
\multirow{2}{*}{\textbf{Model}} & \multicolumn{4}{c|}{\textbf{MCIT}} & \multicolumn{4}{c|}{\textbf{ICIT}} & \multicolumn{4}{c|}{\textbf{MVIT}} & \multicolumn{4}{c}{\textbf{MAIT}} \\
\cline{2-17}
 &
  \textbf{E} &
  \textbf{L} &
  \textbf{T} &
  \textbf{WK} &
  \textbf{E} &
  \textbf{L} &
  \textbf{T} &
  \textbf{WK} &
  \textbf{E} &
  \textbf{L} &
  \textbf{T} &
  \textbf{WK} &
  \textbf{E} &
  \textbf{L} &
  \textbf{T} &
  \textbf{WK} \\ 
\hline
GPT-4o$^{\dagger}$ \cite{openai_gpt4o}         & \colorbox{ThemeColor}{36.28} & \colorbox{ThemeColor}{20.47} & \colorbox{ThemeColor}{15.87} & \colorbox{ThemeColor}{19.31} & \colorbox{ThemeColor}{50.97} & \colorbox{ThemeColor}{34.61} & \colorbox{ThemeColor}{28.89} & \colorbox{ThemeColor}{34.88} & \colorbox{ThemeColor}{43.65} & \colorbox{ThemeColor}{28.77} & \colorbox{ThemeColor}{22.94} & \colorbox{ThemeColor}{29.31} & \colorbox{ThemeColor}{40.27} & \colorbox{ThemeColor}{24.91} & \colorbox{ThemeColor}{18.76} & \colorbox{ThemeColor}{26.48} \\
Gemini 1.5 Pro$^{\dagger}$ \cite{gemini} & 33.94 & 18.64 & 13.32 & 17.96 & 48.66 & 32.25 & 27.19 & 33.01 & 41.29 & 26.43 & 21.72 & 27.66 & 39.19 & 23.76 & 18.13 & 25.05 \\
\hdashline
VideoLLaMA2 \cite{videollama2}   & 33.65 & 18.21 & 14.25 & 15.39 & 47.61 & 31.20 & 27.05 & 30.37 & 39.32 & 22.69 & 19.92 & 23.15 & 36.71 & 20.24 & 17.75 & 20.62 \\
Bay-CAT \cite{baycat}        & 33.41 & 18.03 & 14.29 & 15.23 & 47.38 & 31.14 & 26.79 & 30.02 & 39.97 & 23.47 & 20.63 & 24.03 & 37.42 & 20.88 & 17.93 & 21.55 \\
video-SALMONN \cite{videosalmonn}  & 33.19 & 17.85 & 13.98 & 14.64 & 47.16 & 30.87 & 26.84 & 29.76 & 40.81 & 25.31 & 20.85 & 25.78 & 37.68 & 21.05 & 17.88 & 21.67 \\
ImageBind-LLM \cite{imagebindllm}  & 30.52 & 15.38 & 10.84 & 12.11 & 44.36 & 29.65 & 26.31 & 27.54 & 38.49 & 21.86 & 19.47 & 22.62 & 35.15 & 18.31 & 17.16 & 19.73 \\
VideoLLaMA \cite{videollama}    & 27.43 & 11.96 & 5.62  & 7.38  & 41.62 & 25.87 & 19.23 & 22.91 & 35.26 & 16.82 & 13.21 & 15.64 & 32.15 & 14.27 & 11.44 & 13.36 \\
OneLLM  \cite{onellm}       & 25.77 & 9.63  & 4.86  & 7.97  & 38.37 & 24.28 & 15.04 & 22.33 & 31.65 & 16.81 & 9.88  & 16.76 & 29.29 & 13.36 & 7.97  & 14.51 \\
X-InstructBLIP \cite{xinstructblip} & 22.21 & 10.24 & 5.97  & 7.26  & 35.55 & 23.77 & 19.28 & 20.78 & 31.73 & 15.36 & 10.93 & 12.34 & 29.06 & 14.28 & 8.08  & 10.99 \\
ChatBridge \cite{chatbridge}     & 17.22 & 8.91  & 5.88  & 6.92  & 31.57 & 22.14 & 18.63 & 20.36 & 27.62 & 14.77 & 12.18 & 13.54 & 25.24 & 11.42 & 9.55  & 11.92 \\
PandaGPT \cite{pandagpt}      & 16.13 & 7.28  & 4.34  & 5.20  & 28.36 & 22.85 & 18.02 & 21.62 & 23.14 & 14.16 & 12.04 & 14.15 & 20.47 & 11.39 & 9.68  & 12.33 \\
Macaw-LLM  \cite{macawllm}    & 15.59 & 8.64  & 3.59  & 4.13  & 29.25 & 21.09 & 15.21 & 19.07 & 23.36 & 11.34 & 7.34  & 12.47 & 21.43 & 9.78  & 6.83  & 10.58 \\
VAST  \cite{vast}         & 13.59 & 7.31  & 1.80  & 2.43  & 27.22 & 20.29 & 13.44 & 17.60 & 18.84 & 14.25 & 6.31  & 10.74 & 16.62 & 11.79 & 4.95  & 8.34  \\
 
\hline
\end{tabular}%
}
\vspace{-0.1in}
\caption{\textbf{ZS evaluation results of AVLLMs for \adversarialsuite suite on AVQA dataset under \textit{instruction} setting}. E: Existential, L: Localization, T: Temporal, WK: World Knowledge. $\dagger$ represents closed-source models. Best results are highlighted.}
\label{tab:zs_adversarial_avqa}
\vspace{-4mm}
\end{table*}




\vspace{0.02in}
\noindent\textbf{\compositionalsuite.} This collection of tasks consists of multi-event audio-visual inputs where the sequence of event occurrences as well as their corresponding attribute binding may be distorted. The fundamental goal of multi-modal processing is to comprehend how the linguistic component aligns with the contents of the audio-video input pairs. Therefore, it is pivotal for AVLLMs to acknowledge that disparate word arrangements in a sentence can yield different multimodal perceptions. \compositionalsuite\ suite includes the following set of tasks:

\begin{itemize}[leftmargin=*]

\vspace{-1mm}
\item\textbf{Compositional Order Task (COT).} In a multi-event audio-video sequence, the order of occurrences of the events plays an important role in describing the entire semantical context. In particular, an audio-visual event may either precede, succeed, or simultaneously co-exist with another event. Therefore, we introduce \textit{order stitching task} as \textbf{COT-Stitch}, where we specifically \textit{stitch} two separate videos along with their corresponding audios one after the other and ask the model to comment on the order of events (Fig. \ref{fig:task_suites}). We also introduce  \textit{order swapping task} as \textbf{COT-Swap}, where we \textit{swap} the order of audio events, keeping the video events unaltered (or vice-versa) and verify if the model can recognize this anomaly (Fig. \ref{fig:task_suites}).  


\item\textbf{Compositional Attribute Binding Task (CAT).} Compositional understanding is not only restricted to comprehending the order of event occurrences but also understanding \textit{attribute-binding} of these disparate events. We are particularly inspired by the Winoground dataset \cite{winoground} built for evaluating vision-linguistic compositional reasoning. In this task, each audio-video pair contains two separate events which are associated with two different attributes. In Fig. \ref{fig:task_suites}, \textit{`a goat is bleating'} and a \textit{`rooster screaming'}. Note the answer choices contain the exact same words but in a different sequence. An AVLLM needs to have a strong audio-visual-linguistic understanding to comprehend the constituent modalities and semantically align them with the correct attribute.



\end{itemize}

\vspace{0.02in}

\noindent{\textbf{\missingmodalitysuite.}} This suite consists of tasks aimed at understanding AVLLM's dependency on the \textit{constituent} input modalities of a video.
Note that we consider only those instances where both modalities are \textit{essential} to answer a question, i.e., audio and visual modalities contain nuanced and complementary information. For instance, given the question in Fig. \ref{fig:task_suites} "\textit{Is the violin on the right louder than the drum in the middle?}", it is important to not only understand the audio content but also inspect the visual stream to gather information about its spatial orientation for a correct answer. We divide \missingmodalitysuite\ suite into the following categories:

\begin{itemize}[leftmargin=*]

\vspace{-1mm}
\item\textbf{Missing Audio Detection Task (MAT).} In this setting we remove the audio content from the input. Through this task we want to infer the dependency of the current AVLLMs on audio modality provided the video input is shown. 

\item\textbf{Missing Video Detection Task (MVT).} We remove the video content and keep the audio intact. We want to investigate how much the AVLLMs rely on visual inputs.

\end{itemize}

\begin{table*}[!t]
    \centering
    \begin{minipage}{0.34\textwidth}
        \centering
\renewcommand{\arraystretch}{1.12}
\resizebox{\columnwidth}{!}{%
\begin{tabular}{l|ccc}
\hline
\textbf{Model} & \textbf{COT-Stitch} & \textbf{COT-Swap} & \textbf{CAT} \\ \hline
GPT-4o & \colorbox{ThemeColor}{38.41} & 30.66 & \colorbox{ThemeColor}{31.52} \\
Gemini 1.5 Pro   & 37.19 & \colorbox{ThemeColor}{30.69} & 30.37 \\
\hdashline
VideoLLaMA2      & 36.45               & 30.52             & 30.59        \\
Bay-CAT      & 36.71               & 30.41             & 30.77        \\
video-SALMONN      & 36.93               & 30.37             & 30.48        \\
ImageBind-LLM    & 36.28               & 30.69 & 30.45 \\
VideoLLaMA      & 35.24               & 29.81             & 30.33        \\
OneLLM          & 33.55               & 29.45             & 30.35        \\
X-InstructBLIP   & 32.57               & 26.18             & 29.35        \\
ChatBridge       & 32.03               & 27.32             & 28.94        \\
PandaGPT         & 31.94               & 26.44             & 29.42        \\
Macaw-LLM        & 30.66               & 27.35             & 28.47        \\
VAST             & 25.19               & 25.52             & 25.11        \\
\hline
\end{tabular}%
}
\vspace{-0.1in}
\caption{\textbf{ZS evaluation under \compositionalsuite tasks.} The overall suboptimal performance of AVLLMs underlines their lack of strong compositional understanding.}
\label{tab:zs_compositional}
    \end{minipage}
\hfill
\begin{minipage}{0.64\textwidth}
        \centering
        \renewcommand{\arraystretch}{1.1}
\resizebox{\columnwidth}{!}{%
\begin{tabular}{l|ccccc|ccccc}
\hline
\multirow{2}{*}{\textbf{Model}} & \multicolumn{5}{c|}{\textbf{MVT}} & \multicolumn{5}{c}{\textbf{MAT}} \\
\cline{2-11}
 &
  \textbf{E} &
  \textbf{L} &
  \textbf{Cn} &
  \textbf{T} &
  \textbf{Co} &
  \textbf{E} &
  \textbf{L} &
  \textbf{Cn} &
  \textbf{T} &
  \textbf{Co} \\ 
\hline
GPT-4o & \colorbox{ThemeColor}{57.82} & \colorbox{ThemeColor}{51.63} & \colorbox{ThemeColor}{48.11} & \colorbox{ThemeColor}{41.77} & \colorbox{ThemeColor}{63.18} & \colorbox{ThemeColor}{54.26} & \colorbox{ThemeColor}{47.90} & \colorbox{ThemeColor}{45.39} & \colorbox{ThemeColor}{39.24} & \colorbox{ThemeColor}{58.95} \\
Gemini 1.5 Pro & 56.90 & 50.67 & 47.23 & 41.22 & 61.93 & 52.71 & 46.28 & 43.64 & 37.16 & 57.34 \\
\hdashline
VideoLLaMA2  & 51.44 & 46.92 & 43.15 & 38.71 & 57.98 & 48.22 & 42.97 & 39.42 & 34.66 & 53.71 \\
Bay-CAT  & 52.91 & 47.68 & 44.57 & 39.85 & 59.03 & 49.89 & 44.16 & 40.94 & 36.10 & 54.69 \\
video-SALMONN  & 54.12 & 48.81 & 45.62 & 41.05 & 60.11 & 51.52 & 45.49 & 42.16 & 37.80 & 55.76 \\
ImageBind-LLM  & 49.33 & 44.28 & 41.29 & 36.24 & 55.52 & 46.61 & 41.55 & 37.19 & 32.83 & 51.32 \\
Video LLaMA     & 46.39 & 41.45 & 38.48 & 32.91 & 51.17 & 43.58 & 38.77 & 34.11 & 28.44 & 48.92 \\
One LLM        & 44.99 & 39.38 & 36.75 & 29.58 & 50.28 & 40.39 & 36.32 & 32.57 & 25.62 & 46.15 \\
X-InstructBLIP & 44.22 & 38.03 & 37.39 & 27.58 & 49.31 & 41.23 & 34.12 & 33.49 & 24.16 & 46.33 \\
ChatBridge     & 44.93 & 36.23 & 35.45 & 26.47 & 47.93 & 40.38 & 33.55 & 32.54 & 23.22 & 44.19 \\
PandaGPT       & 41.59 & 34.68 & 34.52 & 24.35 & 45.12 & 38.25 & 31.47 & 30.16 & 21.93 & 42.46 \\
Macaw-LLM      & 40.50 & 33.44 & 35.86 & 25.11 & 47.41 & 37.25 & 30.44 & 31.28 & 22.43 & 44.27 \\
VAST           & 33.52 & 28.88 & 27.81 & 20.20 & 41.59 & 29.46 & 24.82 & 24.06 & 16.39 & 37.48 \\
\hline
\end{tabular}%
}
\vspace{-0.1in}
\caption{\textbf{ZS evaluation results on \missingmodalitysuite suite for MUSIC-AVQA dataset under \textit{instruction} setting.} Results show that this is the \textit{easiest} of the three presented dimensions with the highest average accuracy reported by GPT-4o across the subtasks. E: Existential, L: Localization, Cn: Count, T: Temporal, Co: Comparative.}
\label{tab:zs_mvt}
\end{minipage}
\end{table*}




\begin{figure*}[!t]
    \centering
    \includegraphics[width=\textwidth]{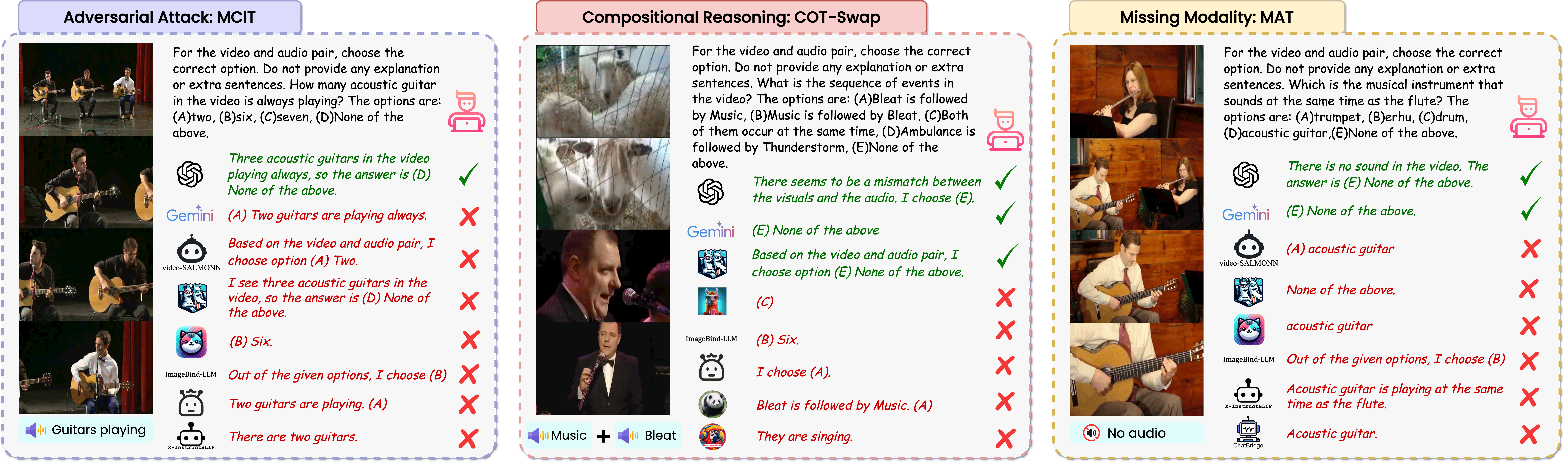}
    \vspace{-0.25in}
    \caption{\textbf{Qualitative results.} We report top 8 models' performance on three representative tasks MCIT, COT-Swap and MAT. GPT-4o consistently outperforms open-source models. Under \textit{instruction} setting we append the phrase \texttt{``If the correct answer is not present respond with None of the above''}. More qualitative results can be found in the supplementary.}
    \label{fig:qual_main_paper}
    \vspace{-5mm}
\end{figure*}

\vspace{0.05in}

\customsubsection{\ourbenchmark Statistics}

A comprehensive task-wise dataset statistics is illustrated in Fig. \ref{fig:data_stats}. The \adversarialsuite suite contains $\sim$ 350K samples and is adapted from the AVQA \cite{avqadataset} and MUSIC-AVQA \cite{musicavqadataset} datasets. We curate the \compositionalsuite suite containing $\sim$ 218K samples carefully chosen from AudioSet \cite{audioset} while $\sim$ 42K samples for \missingmodalitysuite suite are curated again from MUSIC-AVQA \cite{musicavqadataset} dataset. We retain the original category labels (\textit{`Existential', `Temporal', `Count', `Localisation', `Comparison'}) from the MUSIC-AVQA dataset while forming the QA pairs. To get similar insights within the AVQA dataset, we categorize every sample into one of the \textit{`Existential', `Temporal', `Localisation'} and \textit{`World Knowledge'} categories. We define these categories taking inspiration from MUSIC-AVQA and assign each sample into one of them using a carefully designed semi-automated (lookup + prefix matching) strategy (details in supplementary). For all our evaluations we use the \ourbenchmark\texttt{-test} set comprising \totalbenchmarktestsize samples.

\vspace{-1mm}
\section{Evaluating AVLLMs on \ourbenchmark}
\vspace{-1mm}


\customsubsection{Model Selection and Evaluation Metric}


We choose 11 open-source \cite{pandagpt, videollama, xinstructblip, macawllm, chatbridge, onellm, vast, imagebindllm, videollama2, baycat, videosalmonn} and 2 closed-source \cite{gemini, openai_gpt4o} AVLLMs that support video and open-world audio We post-process the models' output to extract its choice.

For QA pairs with no correct choice standard accepted answers are \textit{`None of the above'}, \textit{`The choices are irrelevant'}, \textit{`the video and question are mismatched'} and their variants (in the base setting), and \textit{`None of the above'} as a dedicated \textit{option} when it is explicitly provided in the answer set and instruction (refer to supplementary for more details on base and instruction settings). We choose Top-1 accuracy as the measure for evaluating all the models by extracting answers from model outputs using a \textit{choice extraction strategy} outlined in the supplementary.

\vspace{0.02in}

\customsubsection{Multi-dimensional Analysis and Key Takeaways}

\vspace{0.05in}


Fig. \ref{fig:qual_main_paper} illustrates the responses from the different AVLLMs for three representative tasks MCIT, COT-Swap and MAT. While models such as VAST demonstrate an overall poor performance across all the dimensions, due to its design choice (maps every modality to text), GPT-4o demonstrates an overall edge over other open-source models (see Tabs. \ref{tab:zs_adversarial_avqa} - \ref{tab:zs_mvt}). \textbf{Our key observations are summarized below}:

\noindent \textbf{Impact of different model architectures.} Bridge networks are responsible for mitigating the gap between the text and other modalities by transforming multi-modal features into tokens consistent with the LLM's embedded space (more discussion in supplementary). Tabs. \ref{tab:zs_adversarial_avqa} - \ref{tab:zs_mvt} show that VAST with the simplest bridge performs the worst as compared to advanced models (e.g., Bay-CAT, video-SALMONN, VideoLLaMA) which use Q-former-based bridges. However, despite Q-former-based bridges showing flexibility in handling the resulting number of AV tokens, they struggle to preserve the local context. Developing a perceiver network with deformable attention \cite{xia2022vision} preserving local information in the resampler while keeping its flexibility, may be useful. Moreover, we empirically find that pre-alignment aids in obtaining superior multi-modal features which are fed to LLM. For instance, VideoLLaMA2, Bay-CAT, and ImageBind-LLM use ImageBind encoders which are extensively pre-trained on multi-modal datasets and show superior performance compared to Macaw-LLM (Whisper and CLIP-ViT encoders) and ChatBridge (CLIP-ViT and BEATS) where the modality-encoders are not pre-aligned.

\vspace{0.02in}

\noindent \textbf{Lack of compositional understanding in AVLLMs.} We observe that AVLLMs act as bag of words model. Tab. \ref{tab:zs_compositional} shows that AVLLMs perform only marginally better than random chance on compositional tasks. Moreover, performance gaps between open and closed-source models are the least in \compositionalsuite in comparison to the other two suites. Additionally, increasing the size of LLM backbone leads to marginal improvements in \compositionalsuite as compared to tasks in other suites (see supplementary), suggesting that a bigger LLM variant does not substantially enhance AV compositional reasoning.

\vspace{0.02in}

\noindent \textbf{Comparison of dependency on the constituent modalities.} Results in Tabs. \ref{tab:zs_adversarial_avqa} - \ref{tab:zs_mvt} indicate an inclination of existing AVLLMs towards being more vulnerable to visual content moderation over the audio counterpart. The average category-wise accuracy in MVIT is higher than MAIT denoting that typically the AVLLMs are better equipped to detect the anomaly in the visual modality as compared to the audio modality. Additionally, the aggregated performance of all the models in MVT is higher than MAT indicating the effect of distorting the visual modality has a stronger effect as compared to the audio modality.

\begin{figure}[!t]
    \includegraphics[width=\columnwidth]{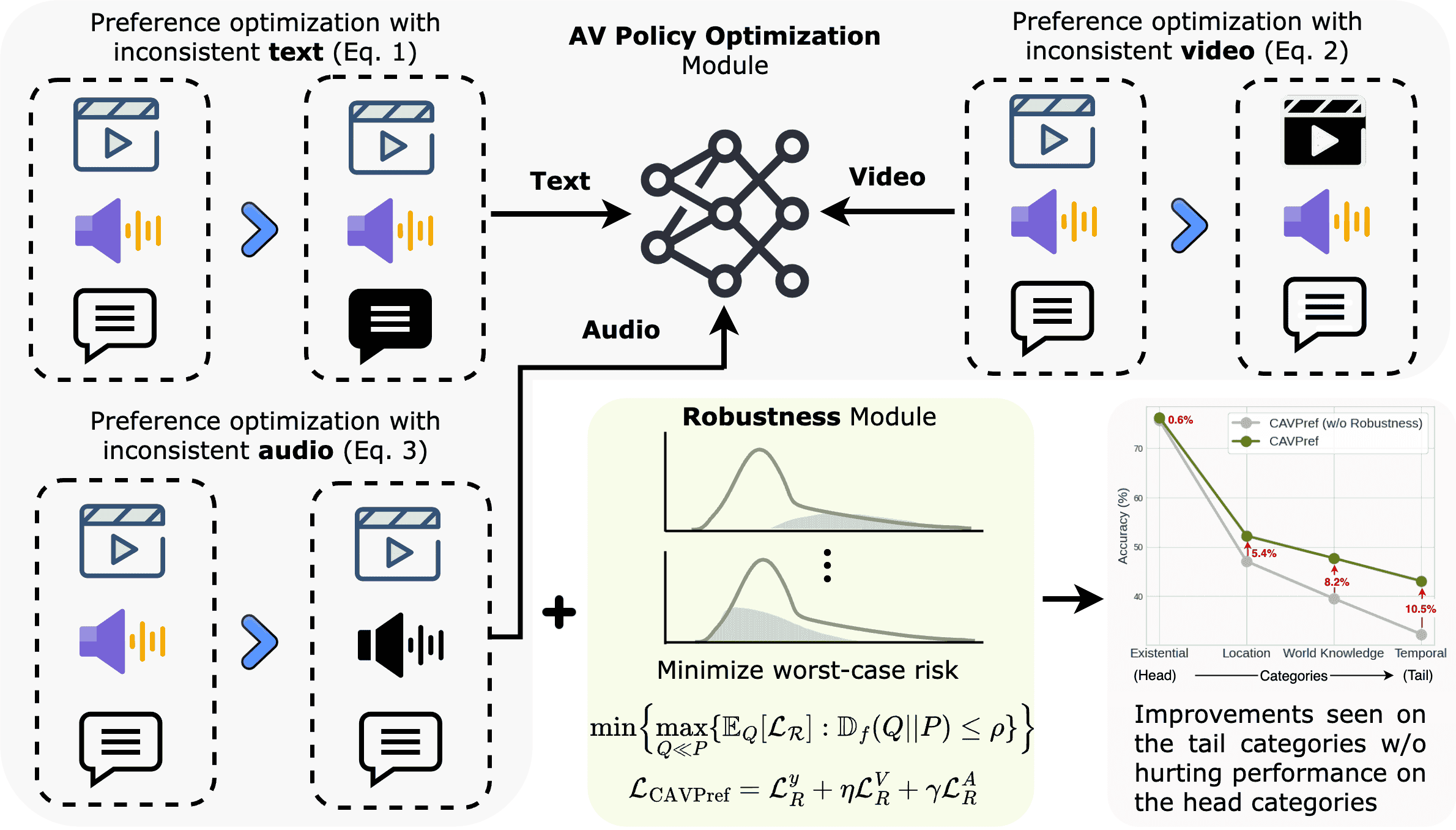}
    \vspace{-0.2in}
    \caption{\textbf{Overview of CAVPref.} We formulate a distributionally robust AV preferential optimization objective to incorporate the multi-modal relationships across different modalities and counter the tailing effect across diverse categories in the dataset.}
    \label{fig:mitigation_diag}
    \vspace{-0.2in}
\end{figure}

\vspace{0.02in}

\noindent \textbf{Performance on commonsense reasoning tasks.} For more reliable interaction between AVLLMs and humans, AVLLMs should comprehend AV scenes with human-like social and contextual reasoning capabilities. Furthermore, open-source AVLLMs tend to respond affirmatively even when presented with ambiguous questions from incompatible AV events. They struggle with counterfactual examples, exposing vulnerabilities and risks for real-world use (see supplementary for detailed discussion and examples). We attribute this limitation to their training dataset and the lack of negative instruction tuning.

\begin{table*}[!t]
\centering
\renewcommand{\arraystretch}{1.1}
\resizebox{\textwidth}{!}{%
\begin{tabular}{l|llll|lll|ll}
\hline
\multirow{2}{*}{\textbf{Mitigation Strategy}} &
  \multicolumn{4}{c|}{\textit{Adversarial Attack}} &
  \multicolumn{3}{c|}{\textit{Compositional Understanding}} &
  \multicolumn{2}{c}{\textit{Modality Dependency}} \\
\cline{2-10}
&
  \textbf{MCIT} &
  \textbf{ICIT} &
  \textbf{MVIT} &
  \textbf{MAIT} &
  \textbf{COT-Stitch} &
  \textbf{COT-Swap} &
  \textbf{CAT} &
  \textbf{MVT} &
  \textbf{MAT} \\
\hline
\rowcolor[HTML]{F8F8F8}
\multicolumn{10}{c}{\textit{VideoLLaMA2}} \\
SFT  & 25.68$^{\demph{+5.30\%}}$ & 39.91$^{\demph{+5.85\%}}$ & 35.27$^{\demph{+9.00\%}}$ & 31.18$^{\demph{+7.35\%}}$ & 42.06$^{\demph{+5.61\%}}$ & 35.26$^{\demph{+4.74\%}}$ & 35.13$^{\demph{+4.54\%}}$ & 52.92$^{\demph{+5.28\%}}$ & 48.52$^{\demph{+4.72\%}}$
 \\
DPO \cite{rafailov2024direct} & 35.82$^{\demph{+15.44\%}}$ & 48.64$^{\demph{+14.58\%}}$ & 36.53$^{\demph{+10.26\%}}$ & 32.16$^{\demph{+8.33\%}}$ & 50.15$^{\demph{+13.70\%}}$ & 36.72$^{\demph{+6.20\%}}$ & 39.45$^{\demph{+8.86\%}}$ & 53.86$^{\demph{+6.22\%}}$ & 49.91$^{\demph{+6.11\%}}$
 \\
\hdashline
\rowcolor{ThemeColor}
CAVPref (w/o Robustness)  & 36.11$^{\textcolor{ForestGreen}{+15.73\%}}$ & 48.95$^{\textcolor{ForestGreen}{+14.89\%}}$  & 48.65$^{\textcolor{ForestGreen}{+22.38\%}}$ & 46.51$^{\textcolor{ForestGreen}{+22.68\%}}$ & 50.97$^{\textcolor{ForestGreen}{+14.52\%}}$ & 46.88$^{\textcolor{ForestGreen}{+16.36\%}}$ & 40.13$^{\textcolor{ForestGreen}{+9.54\%}}$ & 65.42$^{\textcolor{ForestGreen}{+17.78\%}}$ & 64.77$^{\textcolor{ForestGreen}{+20.97\%}}$ \\
\rowcolor{ThemeColor}
\textbf{CAVPref}  & \textbf{41.45}$^{\textcolor{ForestGreen}{\textbf{+21.07\%}}}$ & \textbf{53.61}$^{\textcolor{ForestGreen}{\textbf{+19.55\%}}}$  & \textbf{54.83}$^{\textcolor{ForestGreen}{\textbf{+28.56\%}}}$ & \textbf{53.57}$^{\textcolor{ForestGreen}{\textbf{+29.74\%}}}$ & \textbf{53.06}$^{\textcolor{ForestGreen}{\textbf{+16.61\%}}}$ & \textbf{49.27}$^{\textcolor{ForestGreen}{\textbf{+18.75\%}}}$ & \textbf{43.64}$^{\textcolor{ForestGreen}{\textbf{+13.05\%}}}$ & \textbf{69.81}$^{\textcolor{ForestGreen}{\textbf{+22.17\%}}}$ & \textbf{69.12}$^{\textcolor{ForestGreen}{\textbf{+25.32\%}}}$ \\
\hline
\rowcolor[HTML]{F8F8F8}
\multicolumn{10}{c}{\textit{Bay-CAT}} \\
SFT  & 25.36$^{\demph{+5.12\%}}$ & 39.47$^{\demph{+5.64\%}}$ & 34.56$^{\demph{+7.53\%}}$ & 29.98$^{\demph{+5.54\%}}$ & 42.75$^{\demph{+6.04\%}}$ & 35.04$^{\demph{+4.63\%}}$ & 35.88$^{\demph{+5.11\%}}$ & 53.68$^{\demph{+4.87\%}}$ & 49.14$^{\demph{+4.06\%}}$
 \\
DPO \cite{rafailov2024direct}  & 37.29$^{\demph{+17.05\%}}$ & 51.81$^{\demph{+17.98\%}}$ & 35.14$^{\demph{+8.11\%}}$ & 30.21$^{\demph{+5.78\%}}$ & 53.03$^{\demph{+16.32\%}}$ & 36.95$^{\demph{+6.54\%}}$ & 42.86$^{\demph{+12.09\%}}$ & 54.15$^{\demph{+5.34\%}}$ & 51.44$^{\demph{+6.28\%}}$
 \\
\hdashline
\rowcolor{ThemeColor}
CAVPref (w/o Robustness)  & 37.52$^{\textcolor{ForestGreen}{+17.28\%}}$ & 52.06$^{\textcolor{ForestGreen}{+18.23\%}}$  & 46.27$^{\textcolor{ForestGreen}{+19.24\%}}$ & 45.13$^{\textcolor{ForestGreen}{+20.69\%}}$ & 53.17$^{\textcolor{ForestGreen}{+16.46\%}}$ & 46.92$^{\textcolor{ForestGreen}{+16.51\%}}$ & 43.38$^{\textcolor{ForestGreen}{+12.61\%}}$ & 63.57$^{\textcolor{ForestGreen}{+14.76\%}}$ & 62.89$^{\textcolor{ForestGreen}{+17.73\%}}$ \\
\rowcolor{ThemeColor}
\textbf{CAVPref}  & \textbf{41.95}$^{\textcolor{ForestGreen}{\textbf{+21.71\%}}}$ & \textbf{54.87}$^{\textcolor{ForestGreen}{\textbf{+21.04\%}}}$  & \textbf{49.39}$^{\textcolor{ForestGreen}{\textbf{+22.36\%}}}$ & \textbf{49.46}$^{\textcolor{ForestGreen}{\textbf{+25.02\%}}}$ & \textbf{55.79}$^{\textcolor{ForestGreen}{\textbf{+19.08\%}}}$ & \textbf{49.61}$^{\textcolor{ForestGreen}{\textbf{+19.20\%}}}$ & \textbf{45.78}$^{\textcolor{ForestGreen}{\textbf{+15.01\%}}}$ & \textbf{66.94}$^{\textcolor{ForestGreen}{\textbf{+18.13\%}}}$ & \textbf{66.25}$^{\textcolor{ForestGreen}{\textbf{+21.06\%}}}$ \\
\hline
\rowcolor[HTML]{F8F8F8}
\multicolumn{10}{c}{\textit{video-SALMONN}} \\
SFT & 24.84$^{\demph{+4.92\%}}$ & 38.29$^{\demph{+4.63\%}}$ & 38.13$^{\demph{+9.94\%}}$ & 34.40$^{\demph{+9.82\%}}$ & 42.11$^{\demph{+5.18\%}}$ & 33.97$^{\demph{+3.61\%}}$ & 35.28$^{\demph{+4.80\%}}$ & 55.12$^{\demph{+5.17\%}}$ & 50.35$^{\demph{+3.82\%}}$
 \\
DPO \cite{rafailov2024direct}  & 32.70$^{\demph{+12.78\%}}$ & 45.62$^{\demph{+11.96\%}}$ & 39.25$^{\demph{+11.06\%}}$ & 35.18$^{\demph{+10.61\%}}$ & 49.82$^{\demph{+12.89\%}}$ & 34.85$^{\demph{+4.48\%}}$ & 40.62$^{\demph{+10.14\%}}$ & 56.44$^{\demph{+6.50\%}}$ & 51.65$^{\demph{+5.11\%}}$
 \\
\hdashline
\rowcolor{ThemeColor}
CAVPref (w/o Robustness)  & 33.14$^{\textcolor{ForestGreen}{+13.22\%}}$ & 46.05$^{\textcolor{ForestGreen}{+12.39\%}}$  & 50.47$^{\textcolor{ForestGreen}{+22.28\%}}$ & 49.12$^{\textcolor{ForestGreen}{+24.55\%}}$ & 49.91$^{\textcolor{ForestGreen}{+12.98\%}}$ & 46.15$^{\textcolor{ForestGreen}{+15.78\%}}$ & 40.11$^{\textcolor{ForestGreen}{+9.63\%}}$ & 67.28$^{\textcolor{ForestGreen}{+17.34\%}}$ & 66.24$^{\textcolor{ForestGreen}{+19.69\%}}$ \\
\rowcolor{ThemeColor}
\textbf{CAVPref}  & \textbf{36.87}$^{\textcolor{ForestGreen}{\textbf{+16.95\%}}}$ & \textbf{50.91}$^{\textcolor{ForestGreen}{\textbf{+17.25\%}}}$  & \textbf{54.92}$^{\textcolor{ForestGreen}{\textbf{+26.73\%}}}$ & \textbf{54.77}$^{\textcolor{ForestGreen}{\textbf{+30.19\%}}}$ & \textbf{51.87}$^{\textcolor{ForestGreen}{\textbf{+14.94\%}}}$ & \textbf{49.96}$^{\textcolor{ForestGreen}{\textbf{+19.59\%}}}$ & \textbf{42.89}$^{\textcolor{ForestGreen}{\textbf{+12.41\%}}}$ & \textbf{70.86}$^{\textcolor{ForestGreen}{\textbf{+20.92\%}}}$ & \textbf{70.35}$^{\textcolor{ForestGreen}{\textbf{+23.80\%}}}$ \\
\hline
\end{tabular}%
}
\vspace{-0.05in}
\caption{\textbf{VideoLLaMA2, Bay-CAT and video-SALMONN on \ourbenchmark after applying different model-agnostic mitigation strategies.} \textbf{CAVPref} outperforms SFT and DPO by substantial margins. Accuracy differences with respect to ZS values are shown.}
\label{tab:mitigation}
\vspace{-2mm}
\end{table*}

\vspace{0.1in}
\section{Improving AVLLM through CAVPref}

Zero-shot evaluation results indicate the need to: \textit{(i)} create a preference dataset and perform negative instruction tuning to enhance compositional awareness and commonsense reasoning in AVLLMs, \textit{(ii)} ensure equal emphasis on both audio and video modalities.
Therefore, to improve the performances of AVLLMs on \ourbenchmark, we present a model-agnostic, robust $\beta$-\underline{C}alibrated \underline{A}udio-\underline{V}isual \underline{Pref}erence Optimization method (\textbf{CAVPref}). We compare our proposed method with widely adopted model-agnostic approaches such as Supervised Fine Tuning (SFT) and Direct Preference Optimization (DPO) \cite{rafailov2024direct}.

\vspace{0.05in}

\customsubsection{CAVPref.} With the rise of DPO \cite{rafailov2024direct}, it is possible to align LLMs with human preferences. However, utilizing multi-modal preference data may aggravate hallucination issues as opposed to alleviating them, as found in VLLMs \cite{li2023silkie}. Utilizing non-linguistic information indirectly may lead to a preferential focus on the linguistic counterpart, resulting in sub-optimal performances \cite{sarkar2024mitigating}. Therefore, it is important to have a direct conditioning of the non-linguistic information (e.g., video/audio) while implementing DPO-based approaches. Inspired by this, we propose a model-agnostic solution in an audio-visual setting.

In general, for all task categories in \ourbenchmark, considering textual response, video input, audio input, and question as $y_w$, $y_l$, $V$, $A$, and $q$ respectively, we define:
\vspace{-5mm}

\begin{equation}
\small
    \mathcal{L}^{y} = \log \sigma \left( \beta_y\log\frac{\pi_{\theta}(y_{w}|V,A,q)}{\pi_{\text{ref}}(y_{w}|V,A,q)} - \beta_y\log\frac{\pi_{\theta}(y_{l}|V,A,q)}{\pi_{\text{ref}}(y_{l}|V,A,q)}\right)
    \label{eq:Ly}
\end{equation}

In \ourbenchmark, task categories \adversarialaad, \adversarialiasd, \compositionalstitching, and \compositionalattribute comprise of cases where inconsistencies are only kept in the linguistic counterpart, i.e., the response. However, irregularities occur in video input in \adversarialivqd, \missingvideo, and \compositionalswapping, and in audio input in \adversarialiaqd, and \missingaudio. In particular, in these tasks, audio-visual consistency is absent, i.e., audio and video are either unrelated or one of the modality is missing. In such a scenario, considering only a conventional DPO formulation (Eq. \ref{eq:Ly}) is not only insufficient but also misleading since it only computes reward differences between winning and losing responses. However, reward differences must also be computed between the winning responses in the presence and absence of correct audio-visual conditioning to ensure that the AVLLM understands the correct associations (Fig. \ref{fig:mitigation_diag}). Hence, we define:
\vspace{-0.15in}
\begin{equation}
\scriptsize
    \mathcal{L}^{V} = \log \sigma \left( \beta_V\log\frac{\pi_{\theta}(y_{w}|V_w,A_w,q)}{\pi_{\text{ref}}(y_{w}|V_w,A_w,q)} - \beta_V\log\frac{\pi_{\theta}(y_{w}|V_l,A_w,q)}{\pi_{\text{ref}}(y_{w}|V_l,A_w,q)}\right)
    \label{eq:LV}
\vspace{-0.15in}
\end{equation}

\begin{equation}
\scriptsize
    \mathcal{L}^{A} = \log \sigma \left( \beta_A\log\frac{\pi_{\theta}(y_{w}|V_w,A_w,q)}{\pi_{\text{ref}}(y_{w}|V_w,A_w,q)} - \beta_A\log\frac{\pi_{\theta}(y_{w}|V_w,A_l,q)}{\pi_{\text{ref}}(y_{w}|V_w,A_l,q)}\right)
    \label{eq:LA}
\end{equation}

A critical aspect of DPO formulation (Eqs. \ref{eq:Ly} - \ref{eq:LA}) is its dependency on $\beta$. Specifically, DPO loss can be shown as $\log \left(1 + \frac{f_l}{f_w}^{\beta} \right)$ where $f_w = \frac{\pi_{\theta}(y_w)}{\pi_{\text{ref}}(y_w)}$ and $f_l = \frac{\pi_{\theta}(y_l)}{\pi_{\text{ref}}(y_l)}$ (see supplementary). Thus, in cases where winning and losing responses are semantically close, $\beta$ values should be small and vice-versa. For automatic selection of $\beta$, we propose $\beta$ as an increasing function of (batch) normalized similarity score difference $\Delta S$ between winning and losing scenarios: $\beta = g(\Delta S) = 0.9 \Delta S + 0.1$. For $\beta_y$ (Eq. \ref{eq:Ly}), we use CLAP score differences, and for $\beta_V$ and $\beta_A$ (Eqs. \ref{eq:LV} - \ref{eq:LA}), we use AV Similarity Metric (AVSM) \cite{chowdhury2024melfusion, chowdhury2024meerkat} differences as $\Delta S$.

Additionally, DPO formulation waives the need for a separate reward model by directly learning a policy from collected preference data \cite{rafailov2024direct}. Consequently, such an approach is reliant on the quality of the preference data \cite{liustatistical} which are vast in quantity and collected from multiple sources with diverse distributions. In addition to such distributional shifts, there exist under-represented categories and classes in the datasets, i.e., tail categories and classes (as also in our case, see supplementary). Optimizing the overall expected performance often deteriorates on these tail instances of the population \cite{duchi2021learning}.  To this end, we aim to improve the robustness of policy optimization in an AV setting. Instead of minimizing the average loss, we minimize the worst-case risk (worst-case expected loss) across a set of distributions $Q$ which remain $\rho$-close to the data generating distribution $P$. This not only provides a distributionally robust formulation but also evidently optimizes the tail performance, given as:
\vspace{-6mm}

\begin{equation}
    \text{minimize} \Bigl\{ \underset{Q \ll P}{\max}\{\mathbb{E}_{Q}[\mathcal{L_R}] : \mathbb{D}_{f}(Q||P) \leq \rho \} \Bigr\}
    \vspace{-0.05in}
\end{equation}

With a simplified form (derivation in supplementary) for the above expression and plugging $\mathcal{L}^{y}$, $\mathcal{L}^{V}$, and $\mathcal{L}^{A}$, respectively in place of $\mathcal{L}_R$, we obtain:

\vspace{-4mm}
\begin{equation}
    \mathcal{L}_{R}^{i} = - \lambda_{i}\log\left(\mathbb{E}_{P}\Bigl[e^{\frac{\mathcal{L}^{i}}{\lambda_{i}}}\Bigr]\right), i \in \{y, V, A\}
    \label{eq:L_R}
\end{equation}

Combining the above formulations, we obtain a unified expression for CAVPref:

\vspace{-4mm}
\begin{equation}
    \mathcal{L}_{\text{CAVPref}} = \mathcal{L}_{R}^{y} + \eta \mathcal{L}_{R}^{V} + \gamma \mathcal{L}_{R}^{A}
\end{equation}

Here, $\eta$ and $\gamma$ act as respective binary switching parameters. $\eta = 1$ for \adversarialivqd, \missingvideo, and \compositionalswapping, and $\gamma = 1$ for \adversarialiaqd, and \missingaudio, and 0 otherwise, respectively.

\begin{figure}[!t]
\centering      
\includegraphics[height=0.35\columnwidth, width=\columnwidth]{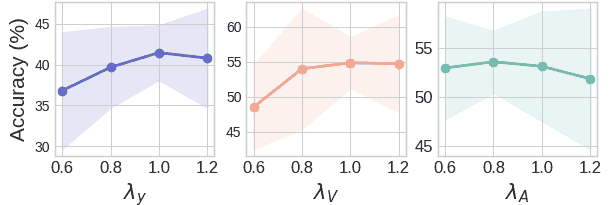}
\vspace{-0.2in}
   \caption{\textbf{Performance changes with varying values of $\lambda_y$, $\lambda_V$ and $\lambda_A$} on MCIT, MVIT and MAIT tasks respectively.}
\label{fig:avllm_ablation}
\vspace{-5.95mm}
\end{figure}

\vspace{0.05in}

\customsubsection{Results and Observations.} In Tab. \ref{tab:mitigation}, we report performances of VideoLLaMA2 \cite{videollama2}, Bay-CAT \cite{baycat}, and video-SALMONN \cite{videosalmonn} upon employing different mitigation techniques (remaining AVLLMs are in supplementary). We make the following observations: (\textit{i}) 
we obtain substantial performance improvements across all tasks (up to 30.19\%) with respect to zero-shot values using \textbf{CAVPref}; (\textit{ii}) the compositional awareness of AVLLMs have improved substantially; (\textit{iii}) The significance of AV conditioning over DPO is particularly evident in tasks like MVIT, MAIT, COT-Swap, MVT, and MAT, where DPO shows only marginal improvement over SFT; (\textit{iv}) The performance gap between MVIT and MAIT, as well as between MVT and MAT, has significantly narrowed, demonstrating that with \textbf{CAVPref}, AVLLMs now give equal importance to all modalities; (\textit{v}) the robustness module significantly improves tail categories without compromising others (refer to Fig. \ref{fig:mitigation_diag}).



\customsubsection{Ablations.} We systematically ablate the values of $\lambda_{y}$, $\lambda_{V}$ and $\lambda_{A}$ in the Eq. \ref{eq:L_R} and assess the performance on MCIT, MVIT and MAIT tasks respectively (Fig. \ref{fig:avllm_ablation}). We observe that a value of 1.0 is the best for both $\lambda_{y}$ and $\lambda_{V}$ whereas for $\lambda_{A}$ the best performance was obtained for a value of 0.8.

\vspace{-1.5mm}
\section{Conclusion}
\vspace{-2mm}
We presented \ourbenchmark, the first multi-dimensional and holistic benchmark suite
that analyses the reliability and robustness of AVLLMs. Through extensive evaluation of a series of SOTA AVLLMs under three critical yet unexplored dimensions: \adversarialsuite, \compositionalsuite, and \missingmodalitysuite, we identify critical findings on the strengths and weaknesses of existing models. Additionally, to improve performances of AVLLMs, we also presented a model-agnostic solution, \textbf{CAVPref}, which leads to substantial improvements. We hope our benchmark will facilitate future development of AVLLMs.

\noindent \textbf{Limitations and Future Work.} Although \textbf{CAVPref} incorporates AV associations, it is essentially a preference-based optimization strategy and is therefore sensitive to the quality of preference data. Moreover, it is yet to be tested whether such an approach can yield promising results for other axes of evaluation and/or fine-grained tasks. \ourbenchmark currently contains coarse-grained samples e.g., QA tasks. Future work can extend this for detection/segmentation. 

\medskip

{
\small




}



{
    \small
    \bibliographystyle{ieeenat_fullname}
    \bibliography{main}
}


\newpage
\appendix



\twocolumn[
    \centering
    \Large
    \textbf{\ourbenchmark: Assessing and Enhancing Reliability and Robustness in Audio-Visual LLMs}\\
    \vspace{0.5em}\textcolor{blue}{Supplementary Material} \\
    \vspace{1.0em}
]





\makeatletter


\appendix


\maketitle







\noindent{\textbf{The supplementary is organised as follows}:



\ref{details of data appendix} More Details about the Data 

\ref{details on eval settings appendix} Additional Details on Evaluation Settings 

\ref{results on zero shot appendix} Additional Results on Zero-Shot Evaluation 

\ref{training_details_appendix} Additional Details on Training 

\ref{bridging_appendix} Discussion on Bridging Networks 

\ref{model variants performance appendix} Performance with Different Model Variants 

\ref{related works appendix} More Related Works 

\ref{implementation details appendix} Implementation Details 

\ref{common sense reasoning appendix} Common Sense Reasoning

\ref{qual results appendix} More Qualitative Examples

\ref{failure cases appendix} Failure Cases

\ref{supp video appendix} Supplementary Video Examples 

\ref{societal impact appendix} Societal Impact 

\ref{human study appendix} Human Study Details



\section{More Details About the Data}
\label{details of data appendix}


\subsection{Exclusion of single modality questions.}
In the original AVQA \cite{avqadataset}, MUSIC-AVQA \cite{musicavqadataset} a subset of the questions were agnostic either of visual or the audio modality, which can be answered with only one modality. However, while forming the QA pairs, we perform a careful inspection to eliminate such samples. To ensure the validity of the \ourbenchmark benchmark, we carefully excluded these questions. we removed $\sim$ 10\% of samples from MUSIC-AVQA for the \adversarialsuite and $\sim$ 50\% for the \missingmodalitysuite respectively. For \compositionalsuite we carefully choose the samples that encompass both the modalities from the AudioSet dataset following a semi-automated strategy. Nearly 30\% of the samples are synthetically generated.



\begin{table*}[!t]
\renewcommand{\arraystretch}{0.9}
\resizebox{\textwidth}{!}{%
\begin{tabular}{l|c|llll}
\cline{1-3}
\multicolumn{1}{c|}{Dimension} & Task & \multicolumn{1}{c}{Sample Question with Options} &  &  &  \\ \cline{1-3}
\multirow{40}{*}{Adversarial Attack} & \multirow{5}{*}{MCIT} & \begin{tabular}[c]{@{}l@{}}Is the ukulele on the left more rhythmic than the saxophone on the right?\\ A. Yes  B. No\end{tabular} &  &  &  \\ \cline{3-3}
 &  & \begin{tabular}[c]{@{}l@{}}Is the instrument on the right louder than the instrument on the left? \\ A. Yes  B. No\end{tabular} &  &  &  \\ \cline{3-3}
 &  & \begin{tabular}[c]{@{}l@{}}How many sounding erhu in the video? \\ A. Five  B. Six  C. More than ten  D. Three  E. None of the above\end{tabular} &  &  &  \\ \cline{3-3}
 &  & \begin{tabular}[c]{@{}l@{}}Where is the lowest instrument? \\ A. Guzheng  B. Middle  C. Bagpipe  D. Right  E. None of the above\end{tabular} &  &  &  \\ \cline{3-3}
 &  & \begin{tabular}[c]{@{}l@{}}What are the main sources of sound in the video? \\ A. Sound of wind  B. Water flow sound  C. Using a sewing machine  D. None of the above\end{tabular} &  &  &  \\ \cline{2-3}
 & \multirow{5}{*}{ICIT} & \begin{tabular}[c]{@{}l@{}}Is the instrument on the right louder than the instrument on the left? \\ A. Napkin  B. Container  C. Calculator  D. Stool  E. None of the above\end{tabular} &  &  &  \\ \cline{3-3}
 &  & \begin{tabular}[c]{@{}l@{}}Is the first sound coming from the middle instrument? \\ A. Book  B. Chair  C. Wok  D. Tree  E. None of the above\end{tabular} &  &  &  \\ \cline{3-3}
 &  & \begin{tabular}[c]{@{}l@{}}Is the xylophone in the video always playing? \\ A.Blanket  B. Cloud  C. Computer  D. Door  E. None of the above\end{tabular} &  &  &  \\ \cline{3-3}
 &  & \begin{tabular}[c]{@{}l@{}}Is the flute in the video more rhythmic than the cello? \\ A.Calculator  B. Statue  C. Rag  D. Kiln  E. None of the above\end{tabular} &  &  &  \\ \cline{3-3}
 &  & \begin{tabular}[c]{@{}l@{}}Is there a voiceover? \\ A. Table  B. Stapler  C. Bag  D. Blanket\end{tabular} &  &  &  \\ \cline{2-3}
 & \multirow{5}{*}{MAIT} & \begin{tabular}[c]{@{}l@{}}Is the first sound coming from the middle instrument? \\ A. Yes B. No\end{tabular} &  &  &  \\ \cline{3-3}
 &  & \begin{tabular}[c]{@{}l@{}}Is the instrument on the right louder than the instrument on the left? \\ A. Yes B. No\end{tabular} &  &  &  \\ \cline{3-3}
 &  & \begin{tabular}[c]{@{}l@{}}Is the xylophone in the video always playing? \\ A. Yes  B. No\end{tabular} &  &  &  \\ \cline{3-3}
 &  & \begin{tabular}[c]{@{}l@{}}Where is the performance? \\ A. Tube  B. Trumpet  C. Flute  D. Indoor  E. None of the above\end{tabular} &  &  &  \\ \cline{3-3}
 &  & \begin{tabular}[c]{@{}l@{}}What is the first instrument that comes in? \\ A. Pipa  B Trumpet  C. Congas  D. Violin\end{tabular} &  &  &  \\ \cline{2-3}
 & \multirow{5}{*}{MVIT} & \begin{tabular}[c]{@{}l@{}}Is the saxophone in the video always playing? \\ A. Yes  B. No\end{tabular} &  &  &  \\ \cline{3-3}
 &  & \begin{tabular}[c]{@{}l@{}}Is the instrument on the right louder than the instrument on the left? \\ A. Yes  B. No\end{tabular} &  &  &  \\ \cline{3-3}
 &  & \begin{tabular}[c]{@{}l@{}}Which is the musical instrument that sounds at the same time as the pipa? \\ A. Flute  B. Guzheng  C. Middle  D. Acoustic guitar  E. None of the above\end{tabular} &  &  &  \\ \cline{3-3}
 &  & \begin{tabular}[c]{@{}l@{}}How many sounding flute in the video? \\ A. Zero  B. Three C. No  D. One\end{tabular} &  &  &  \\ \cline{3-3}
 &  & \begin{tabular}[c]{@{}l@{}}Is the clarinet on the right louder than the accordion on the left? \\ A. Yes  B. No\end{tabular} &  &  &  \\ \cline{1-3}
\multirow{18}{*}{Compositional Reasoning} & \multirow{2}{*}{COT-Stitch} & \begin{tabular}[c]{@{}l@{}}What is the sequence of events in the video? \\ A. Speech is followed by Meow.  B. Meow is followed by Speech. \\ C. Both of them occur at the same time  D. Toilet flush is followed by Toilet flush.\end{tabular} &  &  &  \\ \cline{3-3}
 &  & \begin{tabular}[c]{@{}l@{}}What is the sequence of events in the video? \\ A. Speech is followed by Meow.  B. Meow is followed by Speech \\ C. Both of them occur at the same time  D. Whistle is followed by Helicopter.\end{tabular} &  &  &  \\ \cline{2-3}
 & \multirow{2}{*}{COT-Swap} & \begin{tabular}[c]{@{}l@{}}What is the sequence of events in the video? \\ A. Speech is followed by Meow.  B. Meow is followed by Speech. C. Both of them occur at the same time.  \\ D. Ambulance (siren) is followed by Music. E. None of the above\end{tabular} &  &  &  \\ \cline{3-3}
 &  & \begin{tabular}[c]{@{}l@{}}What is the sequence of events in the video? \\ A. Speech is followed by Meow.  B. Meow is followed by Speech. C. Both of them occur at the same time.  \\ D. Doorbell is followed by Moo. E. None of the above\end{tabular} &  &  &  \\ \cline{2-3}
 & \multirow{2}{*}{CAT} & \begin{tabular}[c]{@{}l@{}}What is the sequence of events in the video? \\ A. A crowd cheers and a man speaks.  B. A crowd speaks and a man cheers. \\ C. Door followed by book\end{tabular} &  &  &  \\ \cline{3-3}
 &  & \begin{tabular}[c]{@{}l@{}}What is the sequence of events in the video? \\ A. A man is speaking, and a crowd applauds.  B. A man is applauding, and a crowd speaks. \\ C. Boots followed by Ring.\end{tabular} &  &  &  \\ \hline
\multirow{16}{*}{Missing Modality} & \multirow{4}{*}{MAT} & \begin{tabular}[c]{@{}l@{}}How many types of musical instruments sound in the video? \\ A. Seven  B. No  C. Three  D. Two  E. None of the above\end{tabular} &  &  &  \\ \cline{3-3}
 &  & \begin{tabular}[c]{@{}l@{}}Is there a voiceover? \\ A. Yes  B. No\end{tabular} &  &  &  \\ \cline{3-3}
 &  & \begin{tabular}[c]{@{}l@{}}Which is the musical instrument that sounds at the same time as the violin? \\ A. Suona  B. Trumpet  C. Middle  D. Accordion  E. None of the above\end{tabular} &  &  &  \\ \cline{3-3}
 &  & \begin{tabular}[c]{@{}l@{}}Is the instrument on the right more rhythmic than the instrument in the middle? \\ A. Yes.  B. No\end{tabular} &  &  &  \\ \cline{2-3}
 & \multirow{4}{*}{MVT} & \begin{tabular}[c]{@{}l@{}}How many sounding flute in the video? \\ A. Zero  B. Three C. No  D. One  E. None of the above\end{tabular} &  &  &  \\ \cline{3-3}
 &  & \begin{tabular}[c]{@{}l@{}}Is the instrument on the left louder than the instrument on the right? \\ A. Yes  B. No\end{tabular} &  &  &  \\ \cline{3-3}
 &  & \begin{tabular}[c]{@{}l@{}}Is the first sound coming from the left instrument? \\ A. Yes  B. No\end{tabular} &  &  &  \\
 \cline{3-3}
 &  & \begin{tabular}[c]{@{}l@{}}What is the first instrument that comes in? \\ A. Acoustic guitar  B. Congas  C. Banjo  D. Violin\end{tabular} &  &  &  \\ \cline{1-3}
\end{tabular}%
}
\caption{Task-wise sample templates with potential options.}
\label{tab:task_wise_propmpts}
\end{table*}

\subsection{Construction of \ourbenchmark}
\label{data statistics appendix}

Tab. \ref{tab:task_wise_propmpts} contains the task-wise question and instruction templates for each task. We carefully construct up to $\sim$ 5 different prompts for each task type. Next, we elaborate on the data preparation strategy for each task.

\paragraph{\adversarialsuite.} For \adversarialsuite we consider the AVQA \cite{avqadataset} and MUSIC-AVQA dataset \cite{musicavqadataset}. We retain the original labels from the MUSIC-AVQA dataset (`Existential', `Localization', etc.) and annotate samples from AVQA with one of the `Existential', `Temporal', `Localisation' and `World Knowledge' categories depending on the QA pair. For AVQA, we prepare two sets that act as look-up tables while forming the options in the below-mentioned cases. The first one (\textbf{T1}) contains a mapping between a given sounding object class of interest and other classes which are not associated with this class \textit{in any way}. This mapping is done through careful manual annotation. The other table (\textbf{T2}) contains category-wise groupings for sounding objects for example `musical instruments', `animal sound', `vehicles' etc. which are the most common supercategories observed in the AVQA dataset. For MUSIC-AVQA, note that the audio files are mostly restricted to music instrument classes. Subsequently, we prepare a Table (\textbf{T3}) mapping the category information (i.e., Existential, Localization, etc.) with all the available Ground Truth answers in the MUSIC-AVQA dataset. For example, the `Existential' category may be mapped to `Flute', `Piano', etc., whereas the `Localization' category may be mapped to `Left', `Right', etc.

\noindent{\textit{\underline{\adversarialaad:}}} For this task we prepare an automated script to first extract the correct response for a given question and replace that with another option from the same category. For example: if the question is `What is the colour of the instrument at the left of the sounding object?' the correct answer `Brown' is replaced with `Black' which is chosen from the previously defined look-up (T2). For the AVQA dataset, we directly adapt its original options before removing the correct choice, while for MUSIC-AVQA we add the options from T3 (as defined above) depending on the question category.

\noindent{\textit{\underline{\adversarialiasd:}}} In this task, we ensure the options provided to the AVLLMs have no relevance at all to the semantics of the question. For AVQA, we again sample the options from a pre-built look-up containing category-wise object/entity names (T1). For example, the category `animal' contains the names of all the animals from the datasets we are dealing with. So while preparing the options for this task we ensure to choose samples from non-overlapping categories. For MUSIC-AVQA, we follow a similar strategy where we sample options based on T3 from the question categories other than the actual category under consideration.

\noindent{\textit{\underline{\adversarialivqd:}}} While preparing the samples for this task, we replace the visual content with completely unrelated visual events. We ensure that this video clip which is used to replace the original video snippet is taken from T1 containing the mapping of this category with other non-overlapping categories for AVQA. For MUSIC-AVQA, we choose options from T3 depending on the question category.

\noindent{\textit{\underline{\adversarialiaqd:}}} Lastly, for AVQA we again employ T1 to find samples which are non-correlated with a sample under consideration and replace its audio content using the latter. For MUSIC-AVQA, we again select options from T3 depending on the question category.    

\paragraph{\compositionalsuite.} We leverage the AudioSet \cite{audioset} dataset to prepare the samples for this task. Below we elaborate on the data preparation strategy.

\noindent{\textit{\underline{COT-Stitch:}}} We carefully choose two semantically separate audio events and concatenate them in the time dimension. The options are prepared by extracting the audio event class. For example, if a \textit{aeroplane engine sound} is concatenated with \textit{a person playing the guitar}, the correct option is: `Aeroplane followed by guitar'. The remaining options are generated using LLM (e.g., GPT-4) where we ask it to swap the ordering of acoustic events, replace the preposition, or swap noun-verb associations. Consequently, the generated options serve as negatives with similar contexts but different compositions which make the task even more challenging. Such generated options in the context of the above example are:`Guitar followed by aeroplane' and `Both events occur simultaneously'. 


\noindent{\textit{\underline{COT-Swap:}}} For this task, the option preparation strategy remains the same as above while the audio components of two dissimilar videos are swapped. We pick the two samples for each case from non-overlapping sets of audio events which we prepare beforehand.

\noindent{\textit{\underline{CAT:}}} For CAT, we first create a collection of several unique audio snippets and their labels where each consists of a single audio event. Using the snippet and label corresponding to the audio events we concatenate or overlay one audio over the other. Additionally, to assure high quality we don't concatenate or overlay random events but ask an LLM to create unique audio scenes. We prepare the options in a similar fashion as described above.

\paragraph{\missingmodalitysuite.} We consider a subset of the MUSIC-AVQA dataset and only consider samples that have a dependency on both audio and visual modalities.

\noindent{\textit{\underline{MVT:}}} We systematically eliminate the video modality from each video in this task. We keep the original answer and add the remaining options by choosing entries from T3 based on the question category under consideration.     

\noindent{\textit{\underline{MAT:}}} We follow the same strategy as MVT except here the audio component is eliminated.

\subsection{Diversity in the data samples.}

Our dataset contains samples from a variety of datasets, e.g., AVQA, MUSIC-AVQA, and AudioSet, eventually making the data points belong to diverse distributions and categories. While our selection of AudioSet contains samples from 190 different categories, AVQA comprises 165 classes (compared to MUSIC-AVQA which comprises samples from 22 musical instruments) - which spans 355 out of a total of 377 categories making the collection of samples considerably diverse. These datasets are widely used in the majority of audio-visual tasks which lead to generalizable models due to the varied categories of events present in them. Additionally, we argue that datasets employed (e.g., CC3M, SBU, TextVQA, Kinetics, etc.) in some of the existing benchmarks do not contain meaningful audio information and hence are not suitable for our study. Finally, the size of our dataset is ~40X larger than recent video benchmarks (SEED-Bench and VideoBench, etc) making it comprehensive and well round.
We provide a comparison on the category-wise diversity of AVTrustBench with other existing benchmarks in the Tab. \ref{tab:diverse_categories_supp}.

\begin{table*}[!t]
    \centering
    \renewcommand{\arraystretch}{1.5}
    \resizebox{0.4\textwidth}{!}{%
    \begin{tabular}{ccc|c}
    \hline
        \bf MSR-VTT & \bf LUMA & \bf SSV2 & \bf \ourbenchmark \\
        \hline
         20 & 50 & 174 & \bf 377 \\
        \hline
    \end{tabular}
    }
    \caption{Comparison of various benchmarks with \ourbenchmark on number of categories.}
    \label{tab:diverse_categories_supp}
\end{table*}

\section{Additional Details on Evaluation Settings}
\label{details on eval settings appendix}

\subsection{Evaluation Settings}
\label{base vs instruction appendix}
Unless stated otherwise, all results presented in this paper adhere to the conventional zero-shot evaluation setting. Below we provide different evaluation settings for the AVLLMs on \ourbenchmark.

\begin{itemize}[leftmargin=*]

    \item \textbf{Base setting.} In this setting, neither additional instructions are provided to the model to withhold answers nor choices such as \textit{None of the above} are provided. This setting represents the most common environment for using and the hardest scenario for evaluating AVLLMs on \adversarialsuite and \missingmodalitysuite suites.

    \item \textbf{Instruction setting.} In this setting, additional options such as "None of the above" and/or additional instruction such as "If all the options are incorrect, answer (D) None of the above." are provided to explicitly drive the model towards acknowledging the inconsistencies in the tasks present in \adversarialsuite, \compositionalsuite, and \missingmodalitysuite suites.

\end{itemize}

\subsection{More Details on LLM-based Choice Extraction}

\paragraph{Choice extraction strategy.}
\label{choice extraction strategy appendix}
We employ a two-step choice extraction strategy which we explain next. Extracting choices from free-form predictions is straightforward for human beings, but might be difficult with rule-based matching. To this end, we design a universal evaluation strategy for all AVLLMs with different instruction-following capabilities:

\noindent{\textit{\textbf{Step 1.} Prediction matching:}}
Initially, we attempt to extract choices from AVLLM predictions using heuristic matching. We aim to extract the choice label (e.g., `A', `B', `C', `D') from the AVLLM’s output. If successful, we use this as the prediction. If not, we attempt to extract the choice label using GPT-4. 

\noindent{\textit{\textbf{Step 2.} GPT-4 processing:}}
Previous evaluation benchmarks \cite{mmbench} establish the effectiveness of GPT-4 as a choice extractor. If step 1 fails, we provide GPT-4 with the question, choices, and model prediction. and instruct it to align the prediction with one of the given choices and produce the label. If there is no match found, GPT-4 returns `No match found’.

We also employ the CircularEval strategy \cite{mmbench} to ensure a rigorous evaluation and effectively demonstrate the performance gap across various models.

\paragraph{Response matching.}
To apply the matching algorithm to the options we maintain the following: when an option is denoted simply by a letter such as `A’ or expressed as `A) \texttt{<response>}’, `A. ’, `A, \texttt{<response>}’, `(A) \texttt{<response>}’ without the inclusion of other choices within the `<\texttt{response}>’ portion, it is considered that option `A’ is being predicted.

\paragraph{Where does heuristic matching fail?}

The heuristic matching strategy typically fails in one of the following cases (i) when the AVLLM is not able to respond with any answer and asks for further clarification `Apologies, can you please clarify ...' or its variants. (ii) when the AVLLMs respond with more than one option choice (A, B, C, etc.). In these cases we move on to Step 2 -- GPT-4 based choice extraction. We provide a sample of how GPT-4 is prompted below.


\begin{tcolorbox}[float, width=\columnwidth, colback=white, colframe=ThemeColor, title=\textcolor{black}{Choice extraction prompt for GPT-4} ] 

Can you help me match an answer with a set of options for a single correct answer type question? I will provide you with a question, a set of options, and a response from an agent. You are required to map the agent's response to the most similar option from the set. You should respond with a single uppercase character in `A', `B', `C', `D', and `E' depending on the choice you feel is the most appropriate match. If there are no similar options you might output `No match found'. Please refrain from being subjective while matching and do not use any external knowledge. Below are some examples:\\
Example 1: \\
Question: What color is the man's shirt who is sitting left of the object making this sound? \\
Options: A. Green B. Red C. Yellow D. Black \\
Answer: The person sitting next to the record player is wearing a black color shirt   \\
Your output: D \\
Example 2: \\
Question: What does the audio-visual event constitute? \\
Options: A. A dog barking at a cat  B. A dog barking on being hit by a stick C. The dog is hungry D. The dog is chasing another dog \\
Answer: It is a wolf  \\
Your output: No match found \\

\end{tcolorbox}

\paragraph{Change in template for GPT-4 evaluation.}

Next, to identify the model prediction, we leverage GPT-4 following MMBench \cite{mmbench}. We query it with the template, including the question, options, and the corresponding AVLLM’s prediction. As for options, we add task-specific options to recognize the model predictions.

For \adversarialaad, we add two options: a masked correct option and the option of `None of the above', `Provided options are incorrect', and `I cannot answer' and its variants.

For \adversarialiasd, we add two options: a masked correct option, and the option of `None of the above', `No option is correct', `Irrelevant options', `I cannot answer.' etc. 

For \adversarialiaqd and \adversarialivqd, we add an option of `The visual/audio is incompatible with the question', or `I cannot answer.'


For COT-Swap, we add an option of `The visual/audio is incompatible', or `I cannot answer.' and its variants.

Finally, for MAT and MVT we add an option of `The audio is missing' and `The video is missing' respectively or `I cannot answer.' and its different variants to handle similar responses from AVLLMs.


\begin{table*}[!t]
\centering
\renewcommand{\arraystretch}{1.5}
\resizebox{\textwidth}{!}{%
\begin{tabular}{l|ccccccccccccc}
\hline
\textbf{Task} &
  \textbf{Video LLaMA} &
  \textbf{Macaw-LLM} &
  \textbf{PandaGPT} &
  \textbf{ChatBridge} &
  \textbf{X-InstructBLIP} &
  \textbf{One LLM} &
  \textbf{VAST} &
  \textbf{ImageBind- LLM} &
  \textbf{Gemini 1.5 Pro} &
  \textbf{VideoLLaMA 2} &
  \textbf{Bay-CAT} &
  \textbf{video-SALMONN} &
  \textbf{GPT-4o} \\
\hline
MCIT & 13.1 / 15.9 & 7.99 / 10.94 & 8.24 / 10.98 & 9.73 / 12.63 & 11.42 / 12.62 & 12.06 / 13.8 & 6.28 / 8.28 & 17.21 / 19.76 & 20.38 / 22.31 & 20.24 / 22.19 & 19.92 / 21.09 & 20.97 / 22.06 & \bf 22.98 / 25.93\\
ICIT & 27.41 / 28.74 & 21.16 / 23.48 & 22.71 / 23.9 & 23.18 / 24.54 & 24.85 / 27.68 & 25.01 / 26.3 & 19.64 / 21.55 & 31.96 / 34.91 & 34.06 / 35.32 & 33.83 / 35.19 & 33.66 / 35.92 & 35.28 / 38.11 & \bf 37.34 / 40.33\\
MVIT & 20.23 / 22.12 & 13.63 / 16.33 & 15.87 / 17.61 & 17.03 / 19.38 & 17.59 / 20.08 & 18.78 / 21.22 & 12.54 / 15.16 & 25.61 / 26.64 & 26.27 / 28.68 & 27.03 / 28.71 & 28.19 / 30.25 & 29.28 / 30.9 & \bf 31.17 / 33.39\\
MAIT & 17.81 / 20.35 & 12.16 / 13.16 & 13.47 / 14.99 & 14.53 / 16.94 & 15.6 / 17.72 & 16.28 / 18.86 & 10.43 / 12.87 & 22.59 / 23.77 & 23.83 / 26.53 & 24.44 / 25.84 & 24.57 / 27.31 & 26.53 / 29.39 & \bf 27.61 / 30.43\\
\hline
COT-Stitch & 35.24 / 36.86 & 30.66 / 32.69 & 31.94 / 34.15 & 32.03 / 33.82 & 32.57 / 34.34 & 33.55 / 36.41 & 25.19 / 27.48 & 36.28 / 38.13 & 36.45 / 37.98 & 36.71 / 37.98 & 36.93 / 39.03 & 37.19 / 39.62 & \bf 38.41 / 40.59\\
COT-Swap & 29.81 / 31.47 & 27.35 / 30.14 & 26.44 / 28.17 & 27.32 / 29.83 & 26.18 / 27.24 & 29.45 / 31.14 & 25.52 / 28.25 & 30.69 / 32.1 & 30.52 / 33.36 & 30.41 / 32.96 & 30.37 / 32.15 & 30.69 / 32.22 & \bf 30.66 / 31.72\\
CAT & 30.33 / 32.05 & 28.47 / 31.46 & 29.42 / 31.73 & 28.94 / 31.29 & 29.35 / 30.4 & 30.35 / 32.62 & 25.11 / 27.63 & 30.45 / 31.88 & 30.59 / 33.43 & 30.77 / 32.12 & 30.48 / 32.87 & 30.37 / 32.29 & \bf 31.52 / 33.14\\
\hline
MVT & 42.08 / 44.16 & 36.46 / 39.4 & 36.05 / 38.77 & 38.2 / 41.05 & 39.31 / 41.66 & 40.2 / 42.29 & 30.4 / 31.86 & 45.33 / 47.88 & 47.64 / 49.98 & 48.81 / 50.07 & 49.94 / 51.27 & 51.59 / 54.05 & \bf 52.5 / 55.36\\
MAT & 38.76 / 40.93 & 33.13 / 34.14 & 32.85 / 34.03 & 34.78 / 36.21 & 35.87 / 37.23 & 36.21 / 38.63 & 26.44 / 28.7 & 41.9 / 43.93 & 43.8 / 45.79 & 45.16 / 47.72 & 46.55 / 49.03 & 47.43 / 48.97 & \bf 49.15 / 50.39\\
\hline
\end{tabular}%
}
\caption{Average accuracy of each model in Circular vs Vanilla Evaluation (given as Circular / Vanilla values).}
\label{tab:circular_vs_vanilla}
\end{table*}

\subsection{Ensuring Robust Evaluation}
\label{robust evaluation appendix}
Inspired by MMBench \cite{mmbench} we employ a CircularEval strategy to ensure robust evaluation. In \ourbenchmark, the problems are presented as multiple-choice questions. Such formulation poses an evaluation challenge: random guessing will lead to $\sim 25\%$ Top-1 accuracy for 4-choice questions. We notice the AVLLMs are prone to predict a certain choice more often introducing bias in the evaluation. Following \cite{mmbench} we feed each question \textit{N} times to the AVLLMs where \textit{N} is the number of choices by making a circular shift to the choices. We attribute the AVLLM to successfully solving a question if it correctly predicts the answer in all circular passes. Once an AVLLM fails in any of the passes there is no need to infer the remaining passes ensuring a good balance between model robustness and cost. 

\subsection{CircularEval vs. VanillaEval}
We first compare the evaluation results under CircularEval (infer a question over multiple passes) with VanillaEval (infer a question only once) and report the average accuracy in Tab. \ref{tab:circular_vs_vanilla} on \ourbenchmark-\texttt{test}. We note, that for most AVLLMs switching from VanillaEval to CircularEval leads to a drop in model accuracy. In general, comparisons under CircularEval reveal a significant performance gap between different AVLLMs. The results as reported in Tab. \ref{tab:circular_vs_vanilla} offer valuable insights, as we find the propensity in current AVLLMs to predict a certain choice when presented with a multiple-choice setup. 



\begin{figure*}[h]
    \centering
    \includegraphics[width=\textwidth]{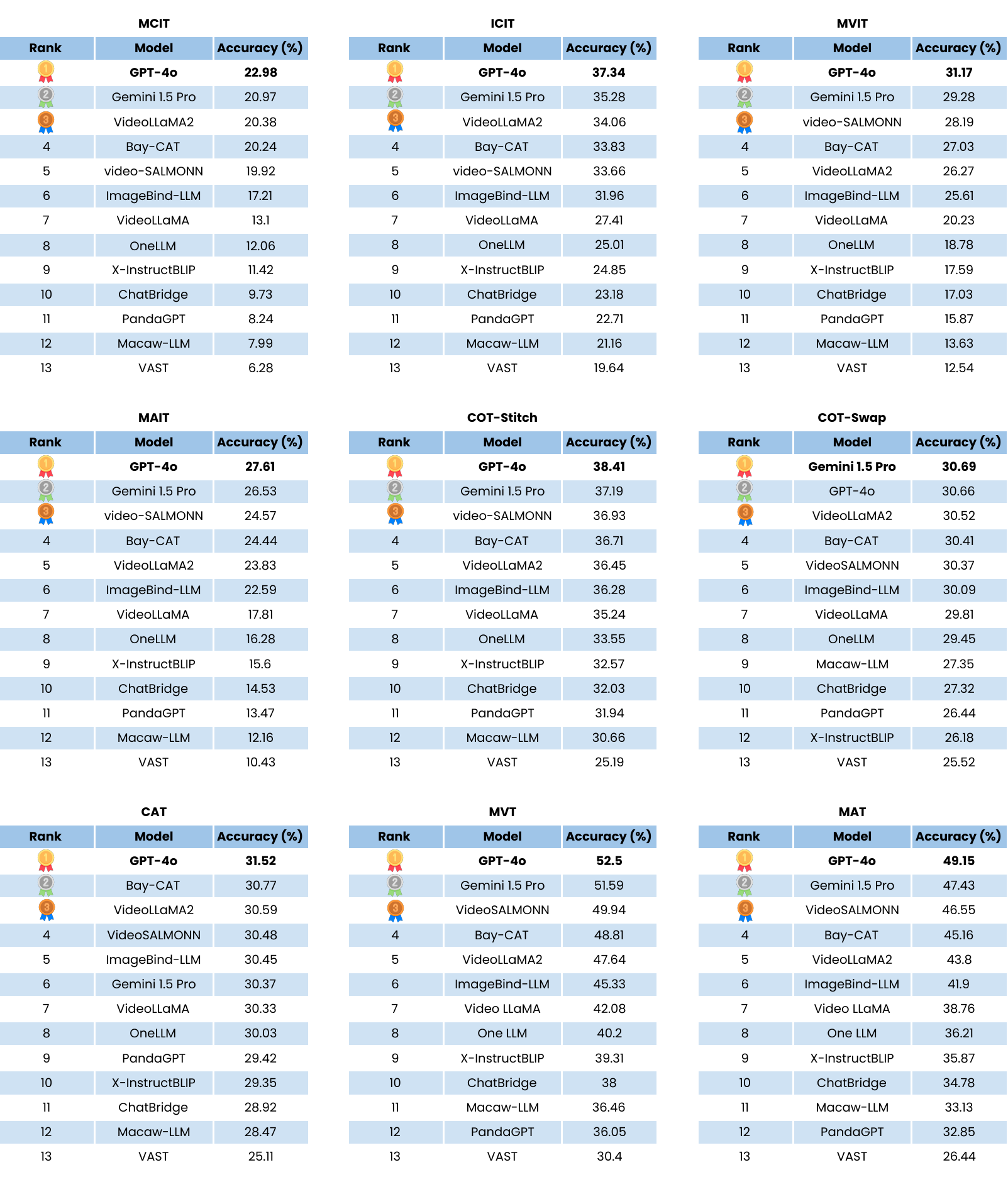}
    \caption{Leaderboards for zero-shot evaluation on 9 different tasks in \ourbenchmark.}
    \label{fig:leaderboard_3x3}
    \vspace{-0.5mm}
\end{figure*}

\subsection{Human Evaluation} 
\label{human_ealuation_appendix}

We manually selected 50 successful and 50 failed cases from the GPT-4o evaluation for each of the 9 tasks and conducted a manual assessment to estimate the upper bound of performance. The average accuracy we achieved was \textbf{91.27\%}, suggesting that the designed tasks are synchronous to human cognition and are relatively straightforward for human subjects. This highlights the significant disparity between the current performance of the benchmark AVLLM and human capabilities. 

\section{Additional Results on Zero-Shot Evaluation}
\label{results on zero shot appendix}

Considering 13 AVLLMs, we provide a leaderboard separately across all the task categories for \ourbenchmark in Fig. \ref{fig:leaderboard_3x3}. Furthermore, we provide additional results on zero-shot evaluations under \textit{base} and \textit{instruction} settings in Tabs. \ref{tab:avqa_base_adversarial} - \ref{tab:adversarial_musicavqa_appendix}. We observe that for all the models the performance in the instruction setting improved considerably. However, the performance of these models is still far from satisfactory.

\subsection{Comparison with different prompts.}

In Tab. \ref{tab:prompts_supp}, we report results of zero-shot evaluation with Video-LLaMA2 on 8 additional prompts, for all the three dimensions of evaluation. We observe that the performance of the AVLLM is sensitive to the prompt used within considerable limits.




\begin{table*}[!t]
\centering
\renewcommand{\arraystretch}{1.5}
\resizebox{\textwidth}{!}{%
\begin{tabular}{l | ccccc cccccccc}
\hline
\textbf{Category} &
  \textbf{Video LLaMA} &
  \textbf{Macaw-LLM} &
  \textbf{PandaGPT} &
  \textbf{ChatBridge} &
  \textbf{X-InstructBLIP} &
  \textbf{OneLLM} &
  \textbf{VAST} &
  \textbf{ImageBind-LLM} &
  \textbf{Gemini 1.5 Pro} &
  \textbf{VideoLLaMA 2} &
  \textbf{Bay-CAT} &
  \textbf{video-SALMONN} &
  \textbf{GPT-4o}  \\ 
\hline
\multicolumn{14}{c}{\textit{Missing Choice Identification Task (MCIT)}} \\
\hline
Existential     & 1.54 & 0.32 & 0.44 & 0.63 & 0.73 & 0.77 & 0.12 & 1.90 & 8.76 & 3.11 & 3.18 & 2.56 & \bf 10.03 \\
Localization    & 0.63 & 0.21  & 0.27  & 0.41  & 0.48  & 0.37  & 0.0  & 0.98 & 5.83 & 1.62 & 2.09 & 1.35 & \bf 6.12 \\
Temporal        & 0.55 & 0.36   & 0.35   & 0.53  & 0.56   & 0.49  & 0.01   & 1.20 & 3.11 & 1.61 & 1.78 & 2.20 & \bf 4.18\\
World knowledge & 0.94 & 0.67   & 0.76   & 0.91  & 0.98   & 0.98  & 0.09   & 1.35 & 6.18 & 2.96 & 2.65 & 1.98 & \bf 6.95 \\
\hline
\multicolumn{14}{c}{\textit{Inconsistent Choice Identification Task (ICIT)}} \\
\hline
Existential     & 3.24  & 2.44 & 2.94 & 2.57 & 4.32 & 3.26 & 1.28 & 4.85 & 11.03 & 5.98 & 5.45 & 5.61 & \bf 12.15\\
Localization    & 3.17 & 2.19 & 2.86 & 2.99 & 3.51 & 3.24 & 0.88 & 4.78 & 9.14 & 5.96 & 5.11 & 5.67 & \bf 9.16 \\
Temporal        & 4.14  & 3.13  & 3.82  & 4.92  & 2.05  & 2.98  & 0.46  & 5.23 & 5.62 & 5.27 & 5.31 & 5.40 & \bf 5.91 \\
World knowledge & 4.49  & 3.39  & 2.82 & 3.16 & 3.60 & 3.48 & 0.65   & 4.57 & 9.06 & 5.78 & 5.91 & 6.22 & \bf 9.42 \\
\hline
\multicolumn{14}{c}{\textit{Mismatched Video Identification Task (MVIT)}} \\
\hline
Existential     & 4.88 & 4.43 & 5.11 & 3.81 & 5.98 & 4.95 & 3.34 & 6.73 & 14.33 & 7.11 & 7.23 & 7.97 & \bf 14.82 \\
Localization    & 5.27 & 3.78 & 4.77 & 3.94 & 5.72 & 4.62 & 2.75 & 5.80 & 11.50 & 6.95 & 6.10 & 7.26 & \bf 12.11 \\
Temporal        & 5.94 & 4.86  & 5.27 & 6.56 & 3.89 & 3.36  & 2.45  & 6.66 & 6.16 & 6.71 & 6.18 & 6.95 & \bf 7.20 \\
World knowledge & 6.58 & 3.96 & 3.76 & 4.93 & 5.64 & 4.97 & 2.90 & 5.82 & 12.53 & 5.97 & 6.42 & 6.55 & \bf 15.12 \\
\hline
\multicolumn{14}{c}{\textit{Mismatched Audio Identification Task (MAIT)}} \\
\hline
Existential     & 3.71 & 3.29 & 3.91 & 2.93 & 4.96 & 3.68 & 2.11 & 5.45 & 12.85 & 7.11 & 6.28 & 7.21 & \bf 13.05 \\
Localization    & 3.46 & 2.64  & 3.42 & 2.71 & 3.58 & 3.74 & 1.49 & 4.11 & 9.78 & 5.89 & 5.62 & 6.24 & \bf 10.12 \\
Temporal        & 4.89 & 3.98  & 4.19  & 3.94  & 3.81  & 2.79  & 1.04  & 4.50 & 5.92 & 4.98 & 4.65 & 5.11 & \bf 6.23 \\
World knowledge & 5.33 & 2.84 & 2.32 & 3.76 & 4.22 & 3.92 & 1.13  & 5.31 & 9.57 & 5.76 & 5.92 & 5.98 & \bf 10.11 \\ 
\hline
\end{tabular}%
}
\caption{\textbf{Zero shot evaluation results of AVLLMs under \adversarialsuite suite on AVQA dataset under \textit{base} setting}. Models are required to demonstrate strong audio-visual comprehension capabilities to withhold answers when presented with perturbed questions/answers/input signals.
}
\label{tab:avqa_base_adversarial}
\end{table*}

\begin{table*}[!t]
\centering
\renewcommand{\arraystretch}{1.5}
\resizebox{\textwidth}{!}{%
\begin{tabular}{l|ccccccccccccc}
\hline
\textbf{Category} &
  \textbf{Video LLaMA} &
  \textbf{Macaw-LLM} &
  \textbf{PandaGPT} &
  \textbf{ChatBridge} &
  \textbf{X-InstructBLIP} &
  \textbf{One LLM} &
  \textbf{VAST} &
  \textbf{ImageBind- LLM} &
  \textbf{Gemini 1.5 Pro} &
  \textbf{VideoLLaMA 2} &
  \textbf{Bay-CAT} &
  \textbf{video-SALMONN} &
  \textbf{GPT-4o} \\
\hline
\multicolumn{14}{c}{\textit{Missing Video Identification Task (MVT)}} \\
\hline
Existential  & 7.58 & 5.24  & 6.31 & 6.27 & 6.36 & 6.44 & 3.59 & 9.25 & 12.48 & 10.65 & 11.51 & 10.97 & \bf 13.77 \\
Localization & 4.22 & 2.30  & 2.20 & 3.51  & 4.43  & 3.27  & 2.42 & 6.50  & 8.74 & 6.91 & 7.01 & 7.13 & \bf 9.22 \\
Count & 4.46 & 2.35 & 2.88  & 2.21 & 1.78 & 2.99 & 1.97 & 5.56 & 8.48 & 6.88 & 6.13 & 6.45 & \bf 10.08 \\
Temporal & 3.37  & 2.23 & 3.36  & 3.46 & 3.15 & 3.67 & 2.76  & 3.44 & 6.19 & 4.98 & 4.87 & 4.91 & \bf 7.55 \\
Comparison & 8.23 & 5.62 & 6.04  & 6.26  & 7.61 & 7.58 & 3.78 & 8.77 & 12.28 & 9.87 & 9.91 & 8.72 & \bf 12.96 \\
\hline
\multicolumn{14}{c}{\textit{Missing Audio Identification Task (MAT)}} \\
\hline
Existential & 6.39 & 4.56 & 4.78 & 5.54 & 5.98 & 5.21 & 2.70 & 7.17 & 8.24 & 7.54 & 7.23 & 7.98 & \bf 9.06 \\
Localization & 3.71 & 1.54 & 1.88 & 2.04 & 2.35 & 2.98 & 1.04  & 5.03 & 7.57 & 7.54 & 7.23 & 8.11 & \bf 8.95 \\
Count & 3.29 & 1.08 & 1.73 & 1.79  & 2.56 & 2.75 & 0.79  & 4.24 & 7.13 & 6.56 & 5.12 & 7.11 & \bf 8.78 \\
Temporal & 2.51 & 1.65 & 2.13 & 2.36 & 2.81 & 2.49 & 1.35 & 2.90 & 3.46 & 2.98 & 3.02 & 3.11 & \bf 3.67 \\
Comparison & 7.71 & 4.84 & 5.34 & 5.72 & 6.26 & 6.91 & 2.47 & 7.46 & 9.84 & 8.52 & 8.76 & 9.03 & \bf 10.15 \\
\hline
\end{tabular}%
}
\caption{Comparison of zero-shot evaluation results on \missingmodalitysuite suite for MUSIC-AVQA dataset under \textit{base} setting.}
\label{tab:musicavqa_base_missing}
\end{table*}





\begin{table*}[!t]
\centering
\renewcommand{\arraystretch}{1.5}
\resizebox{\textwidth}{!}{%
\begin{tabular}{l | ccccc cccccccc}
\hline
\textbf{Category} &
  \textbf{Video LLaMA} &
  \textbf{Macaw-LLM} &
  \textbf{PandaGPT} &
  \textbf{ChatBridge} &
  \textbf{X-InstructBLIP} &
  \textbf{OneLLM} &
  \textbf{VAST} &
  \textbf{ImageBind-LLM} &
  \textbf{Gemini 1.5 Pro} &
  \textbf{VideoLLaMA 2} &
  \textbf{Bay-CAT} &
  \textbf{video-SALMONN} &
  \textbf{GPT-4o} \\ 
\hline
\multicolumn{14}{c}{\textit{Missing Choice Identification Task (MCIT)}} \\
\hline
Existential     & 1.16 / 26.72 & 0.31 / 14.22 & 0.41 / 15.34 & 0.62 / 16.65 & 0.79 / 21.59 & 0.73 / 23.30 & 0.19 / 12.36 & 1.45 / 27.38 & 8.10 / 31.88 & 4.12 / 29.58 & 5.01 / 30.01 & 3.62 / 30.18 & \bf 10.61 / 33.96 \\
Localization    & 0.59 / 10.26 & 0.27 / 7.99  & 0.29 / 7.96 & 0.40 / 8.44 & 0.53 / 9.80 & 0.39 / 9.88 & 0.21 / 7.22 & 0.97 / 13.14 & 5.55 / 19.39 & 2.16 / 16.51 & 3.96 / 18.11 & 2.67 / 18.76 & \bf 7.41 / 21.90 \\
Temporal        & 0.51/5.29 & 0.39 / 3.31 & 0.38 / 5.42 & 0.57 / 6.27 & 0.53 / 5.90 & 0.57 / 4.90 & 0.13 / 1.20 & 1.16 / 11.66 & 3.00 / 12.44 & 1.91 / 10.91 & 2.61 / 11.42 & 1.99 / 11.20 & \bf 5.91 / 14.93 \\
Count & 0.82/7.10 & 0.65 / 4.35 & 0.77 / 5.45 & 1.04 / 7.36 & 0.84 / 7.87 & 0.95 / 7.51 & 0.20 / 3.78 & 1.27 / 13.70 & 6.02 / 17.10 & 3.61 / 15.71 & 4.89 / 15.98 & 3.90 / 14.64 & \bf 8.11 / 19.61\\
Comparative & 1.41 / 27.28 &  0.48 / 15.65 & 0.56 / 17.89 &  0.85 / 18.33 & 0.91 / 23.57 &  0.85 / 26.72 &  0.30 / 14.80 &  3.56 / 31.76 & 11.34 / 34.48 & 6.57 / 32.86 & 7.11 / 32.67 & 6.42 / 32.19 & \bf 12.91 / 36.75\\
\hline
\multicolumn{14}{c}{\textit{Inconsistent Choice Identification Task (ICIT)}} \\
\hline
Existential &
3.43 / 40.33 &
2.40 / 28.38 &
2.96 / 26.91 &
3.01 / 32.65 &
3.51 / 37.59 &
3.65 / 39.11 &
1.12 / 25.19 &
4.11 / 42.36 &
9.57 / 48.89 &
5.82 / 44.85 &
6.01 / 46.48 &
5.42 / 45.53 &
\bf 10.13 / 49.65\\
Localization &
3.12 / 27.11 &
2.02 / 22.61 &
2.11 / 23.01 &
2.82 / 21.88 &
3.24 / 22.96 &
3.21 / 24.18 &
0.49 / 18.42 &
4.05 / 28.78 &
9.31 / 32.06 &
6.15 / 29.18 &
6.89 / 29.64 &
5.92 / 28.57 &
\bf 10.76 / 34.66\\
Temporal &
2.98 / 20.27 &
2.38 / 13.88 &
2.52 / 18.87 &
2.91 / 19.92 &
2.97 / 20.05 &
3.28 / 14.85 &
0.41 / 14.16 &
3.92 / 27.10 &
6.12 / 28.14 &
4.95 / 27.61 &
4.68 / 27.67 &
4.15 / 27.11 &
\bf 7.44 / 30.61\\
Count &
3.13 / 21.76 &
2.76 / 18.54 &
2.79 / 20.42 &
3.06 / 21.03 &
3.21 / 20.83 &
3.09 / 24.62 &
0.67 / 18.80 &
3.86 / 26.24 &
9.02 / 32.55 &
5.64 / 28.55 &
5.98 / 29.41 &
5.75 / 29.62 &
\bf 11.41 / 34.56\\
Comparative &
4.31 / 43.54 &
3.16 / 29.67 &
3.09 / 28.26 &
4.15 / 34.32 &
3.89 / 39.44 &
4.41 / 40.66 &
1.98 / 27.22 &
6.78 / 44.63 &
11.45 / 50.90 &
7.23 / 46.75 &
8.11 / 47.11 &
7.86 / 49.17 &
\bf 12.71 / 51.89\\
\hline
\multicolumn{14}{c}{\textit{Mismatched Video Identification Task (MVIT)}} \\
\hline
Existential &
4.20 / 34.80 &
4.03 / 22.36 &
5.90 / 22.14 &
3.64 / 26.27 &
5.66 / 30.37 &
4.48 / 30.58 &
3.30 / 18.27 &
6.47 / 37.93 &
13.98 / 39.77 &
8.42 / 38.42 &
8.77 / 38.91 &
8.18 / 39.11 &
\bf 15.71 / 41.02\\
Localization &
5.42 / 15.33 &
3.31 / 11.39 &
5.34 / 13.48 &
3.38 / 14.34 &
5.21 / 14.91 &
4.56 / 16.31 &
2.04 / 13.56 &
5.98 / 20.00 &
11.28 / 25.25 &
6.11 / 21.84 &
6.87 / 21.91 &
6.45 / 20.96 &
\bf 12.88 / 27.60\\
Temporal &
5.34 / 12.80 &
4.28 / 8.72 &
5.69 / 12.60 &
6.16 / 12.14 &
4.20 / 10.58 &
3.79 / 10.46 &
3.20 / 7.90 &
6.47 / 18.28 &
6.70 / 22.97 &
5.57 / 16.51 &
5.94 / 16.68 &
5.13 / 17.41 &
\bf 7.19 / 23.96\\
Count &
6.12 / 14.28 &
4.62 / 12.19 &
4.65 / 15.14 &
5.75 / 14.73 &
5.40 / 11.03 &
4.24 / 17.20 &
2.42 / 11.25 &
5.49 / 21.74 &
12.01 / 26.20 &
8.32 / 22.76 &
8.67 / 23.13 &
7.18 / 23.57 &
\bf 13.87 / 27.96\\
Comparative &
4.47 / 35.87 &
5.12 / 24.46 &
6.11 / 23.88 &
3.96 / 27.90 &
6.17 / 32.39 &
4.79 / 32.51 &
4.04 / 19.32 &
7.43 / 38.67 &
14.28 / 41.34 &
9.65 / 39.87 &
9.88 / 39.29 &
8.74 / 38.56 &
\bf 16.41 / 42.98\\
\hline
\multicolumn{14}{c}{\textit{Mismatched Audio Identification Task (MAIT)}} \\
\hline
Existential &
4.68 / 31.51 &
3.88 / 20.67 &
3.47 / 21.77 &
2.52 / 24.24 &
4.62 / 28.20 &
3.63 / 28.35 &
2.35 / 15.51 &
5.21 / 34.34 &
13.61 / 38.29 &
6.75 / 35.78 &
7.42 / 36.17 &
6.57 / 35.57 &
\bf 15.08 / 39.46 \\
Localization &
3.15 / 13.44 &
2.03 / 9.37 &
4.21 / 12.48 &
2.33 / 11.03 &
4.36 / 14.36 &
3.37 / 14.00 &
1.03 / 11.18 &
4.36 / 17.76 &
10.38 / 23.22 &
6.44 / 21.76 &
7.12 / 22.58 &
6.38 / 21.69 &
\bf 11.32 / 24.89 \\
Temporal &
4.32 / 11.68 &
3.46 / 5.46 &
4.77 / 9.50 &
5.70 / 9.89 &
2.27 / 8.67 &
2.84 / 7.25 &
1.56 / 4.44 &
5.90 / 17.92 &
5.66 / 19.72 &
5.61 / 18.71 &
5.65 / 19.02 &
5.43 / 18.76 &
\bf 7.11 / 19.89\\
Count &
5.88 / 13.00 &
2.97 / 9.39 &
2.53 / 12.01 &
4.30 / 11.27 &
4.72 / 10.46 &
3.41 / 14.76 &
1.83 / 8.21 &
4.44 / 19.95 &
10.28 / 24.82 &
6.96 / 21.67 &
7.24 / 22.71 &
6.34 / 20.98 &
\bf 12.16 / 22.58\\
Comparative &
4.92 / 33.90 &
4.56 / 22.72 &
3.77 / 22.32 &
3.15 / 26.29 &
5.27 / 29.81 &
4.45 / 30.77 &
2.78 / 17.99 &
6.37 / 37.75 &
15.29 / 41.66 &
8.16 / 39.58 &
8.78 / 39.90 &
7.61 / 38.66 &
\bf 16.11 / 42.71 \\
\hline
\end{tabular}%
}
\caption{\textbf{Zero shot evaluation results of AVLLMs under \adversarialsuite suite on MUSIC-AVQA dataset under both \textit{base} and \textit{instruction} settings} Results are reported in \textit{base/instruction} format.
}
\label{tab:adversarial_musicavqa_appendix}
\end{table*}

\begin{table*}[!t]
\centering
\renewcommand{\arraystretch}{1.1}
\resizebox{\textwidth}{!}{%
\begin{tabular}{c|c|c|c}
\hline
\bf Prompts & \bf Adversarial & \bf Compositional & \bf Missing Modality \\
\hline
If the correct choice is not provided, reply with "None of the above." & 23.36 & 30.28 & 43.72\\
If none of the options are correct, respond with "None of the above." & 23.55 & 31.80 & 43.14\\
If the right option is not included in the list, use "None of the above." & 24.82 & 31.35 & 44.97\\
If none of the listed options is correct, reply with "None of the above." & 22.48 & 30.71 & 42.33\\
If the right answer is missing from the options, use "None of the above" as your response. & 22.97 & 32.42 & 42.04\\
If the answer is not among the choices, reply with "None of the above." & 23.16 & 31.02 & 42.81\\
If none of the answers are correct, choose "None of the above." & 25.79 & 31.98 & 43.56\\
If no listed option is accurate, respond with "None of the above." & 25.03 & 31.62 & 44.70\\
\hline
\bf If the correct answer is not present, respond with None of the above [reported in paper] & \bf 26.18 & \bf 32.52 & \bf 45.72\\
\hline
\end{tabular}%
}
\caption{\textbf{Comparison with different prompts with Video-LLaMA2.} Reported values are aggregated across tasks.}
\label{tab:prompts_supp}
\end{table*}


\begin{table*}[!t]
\centering
\renewcommand{\arraystretch}{1.1}
\resizebox{\textwidth}{!}{%
\begin{tabular}{l|cccc|ccc|cc}
\hline
\multirow{2}{*}{\textbf{Mitigation Strategy}} &
  \multicolumn{4}{c|}{\textit{Adversarial Attack}} &
  \multicolumn{3}{c|}{\textit{Compositional Understanding}} &
  \multicolumn{2}{c}{\textit{Modality Dependency}} \\
\cline{2-10}
&
  \textbf{MCIT} &
  \textbf{ICIT} &
  \textbf{MVIT} &
  \textbf{MAIT} &
  \textbf{COT-Stitch} &
  \textbf{COT-Swap} &
  \textbf{CAT} &
  \textbf{MVT} &
  \textbf{MAT} \\
\hline
\rowcolor[HTML]{F8F8F8}
\multicolumn{10}{c}{\textit{ImageBind-LLM}} \\
SFT  & 26.50 & 34.84 & 33.13 & 26.08 & 36.43 & 31.63 & 32.62 & 48.30 & 42.83
 \\
DPO \cite{rafailov2024direct} & 32.46 & 41.15 & 35.10 & 27.20 & 44.58 & 32.27 & 38.29 & 48.39 & 42.99
 \\
\hdashline
\rowcolor{ThemeColor}
CAVPref (w/o Robustness)  & 33.19 & 42.00 & 47.39 & 39.11 & 45.10 & 42.49 & 38.72 & 56.48 & 54.91 \\
\rowcolor{ThemeColor}
\textbf{CAVPref}  & \textbf{37.51} & \textbf{45.27} & \textbf{50.24} & \textbf{42.48} & \textbf{48.87} & \textbf{46.91} & \textbf{42.85} & \textbf{60.21} & \textbf{59.74} \\
\hline
\rowcolor[HTML]{F8F8F8}
\multicolumn{10}{c}{\textit{Video-LLaMA}} \\
SFT  & 20.36 &	33.13 &	30.28 &	25.46 &	39.83 &	35.43 &	32.44 &	47.48 &	42.43\\
DPO \cite{rafailov2024direct} & 28.41 &	39.76 &	30.56 &	26.70 &	47.84 &	36.72 &	37.67 &	48.03 &	43.09\\
\hdashline
\rowcolor{ThemeColor}
CAVPref (w/o Robustness)  & 29.08 &	40.57 &	36.19 &	35.41 &	48.01 &	44.13 &	37.93 &	56.31 &	55.49\\
\rowcolor{ThemeColor}
\textbf{CAVPref}  & \bf 32.44 &	\bf 44.53 &	\bf 40.86 &	\bf 38.74 &	\bf 50.22 &	\bf 47.66 &	\bf 41.95 &	\bf 60.08 &	\bf 60.29\\
\hline
\rowcolor[HTML]{F8F8F8}
\multicolumn{10}{c}{\textit{One-LLM}} \\
SFT  & 18.52 &	31.25 &	25.65 &	23.50 &	35.55 &	31.05 &	32.64 &	45.54 &	41.36\\
DPO \cite{rafailov2024direct} & 26.19 &	38.77 &	26.41 &	24.14 &	42.89 &	31.82 &	39.87 &	46.56 &	42.03\\
\hdashline
\rowcolor{ThemeColor}
CAVPref (w/o Robustness)  & 26.85 &	39.57 &	34.20 &	32.88 &	43.10 &	39.80 &	40.15 &	54.15 &	52.07\\
\rowcolor{ThemeColor}
\textbf{CAVPref}  & \bf 30.43 &	\bf 42.96 &	\bf 37.56 &	\bf 35.07 &	\bf 46.61 &	\bf 42.62 &	\bf 44.57 &	\bf 57.95 &	\bf 56.14\\
\hline
\rowcolor[HTML]{F8F8F8}
\multicolumn{10}{c}{\textit{X-InstructBLIP}} \\
SFT  & 15.67 &	30.02 &	26.06 &	20.18 &	37.35 &	31.07 &	33.67 &	43.82 &	39.37\\
DPO \cite{rafailov2024direct} & 24.03 &	38.26 &	26.77 &	21.35 &	45.49 &	32.79 &	39.76 &	45.31 &	40.07\\
\hdashline
\rowcolor{ThemeColor}
CAVPref (w/o Robustness) & 25.41 &	39.43 &	33.63 &	29.34 &	45.68 &	40.65 &	40.05 &	56.67 &	52.99\\
\rowcolor{ThemeColor}
\textbf{CAVPref} & \bf 29.20 &	\bf 41.99 &	\bf 37.05 &	\bf 34.16 &	\bf 48.94 &	\bf 43.97 &	\bf 44.70 &	\bf 58.79 &	\bf 55.07\\
\hline
\rowcolor[HTML]{F8F8F8}
\multicolumn{10}{c}{\textit{ChatBridge}} \\
SFT  & 14.09 &	28.48 &	25.09 &	19.23 &	34.39 &	30.37 &	31.8 &	41.67 &	38.78\\
DPO \cite{rafailov2024direct} & 22.34 &	37.04 &	26.69 &	19.86 &	42.79 &	31.04 &	37.11 &	42.25 &	39.84\\
\hdashline
\rowcolor{ThemeColor}
CAVPref (w/o Robustness) & 23.22 &	37.61 &	34.72 &	28.01 &	42.83 &	39.17 &	37.25 &	48.40 &	47.88\\
\rowcolor{ThemeColor}
\textbf{CAVPref} & \bf 26.23 &	\bf 41.5 &	\bf 37.08 &	\bf 33.59 &	\bf 46.85 &	\bf 41.97 &	\bf 40.06 &	\bf 51.86 &	\bf 50.13\\
\hline
\rowcolor[HTML]{F8F8F8}
\multicolumn{10}{c}{\textit{PandaGPT}} \\
SFT & 12.36 & 27.34 &	20.39 &	17.88 &	34.65 &	30.42 &	32.15 &	38.34 &	36.1\\
DPO \cite{rafailov2024direct} & 20.84 &	33.56 &	21.10 &	18.20 &	42.40 &	31.24 &	40.35 &	39.49 &	37.78\\
\hdashline
\rowcolor{ThemeColor}
CAVPref (w/o Robustness) & 21.56 &	34.13 &	30.42 &	26.30 &	42.78 &	39.37 &	40.86 &	46.33 &	45.65\\
\rowcolor{ThemeColor}
\textbf{CAVPref} & \bf 24.75 &	\bf 38.12 &	\bf 35.73 &	\bf 29.23 &	\bf 45.41 &	\bf 42.46 &	\bf 44.09 &	\bf 49.51 &	\bf 48.72\\
\hline
\rowcolor[HTML]{F8F8F8}
\multicolumn{10}{c}{\textit{Macaw-LLM}} \\
SFT  & 11.4 &	25.05 &	20.46 &	15.21 &	35.56 &	30.92 &	32.2 &	39.97 &	34.21\\
DPO \cite{rafailov2024direct} & 18.05 &	33.4 &	20.85 &	16.73 &	42.16 &	31.35 &	38.12 &	40.44 &	34.65\\
\hdashline
\rowcolor{ThemeColor}
CAVPref (w/o Robustness) & 19.36 &	34.42 &	31.44 &	24.3 &	42.87 &	40.97 &	38.83 &	49.36 &	45.52\\
\rowcolor{ThemeColor}
\textbf{CAVPref}  & \bf 23.05 &	\bf 37.03 &	\bf 33.81 &	\bf 28.77 &	\bf 45.94 &	\bf 43.77 &	\bf 40.28 &	\bf 51.82 &	\bf 48.96\\
\hline
\end{tabular}%
}
\caption{\textbf{ImageBind-LLM, Video-LLaMA, One-LLM, X-InstructBLIP, ChatBridge, PandaGPT, and Macaw-LLM on \ourbenchmark after applying different model-agnostic mitigation strategies.} \textbf{CAVPref} outperforms SFT and DPO by substantial margins.}


\label{tab:mitigation_supp}
\end{table*}

\begin{table*}[t!]
    \centering
    \renewcommand{\arraystretch}{1.5}
    \resizebox{0.6\textwidth}{!}{%
    \begin{tabular}{c|c|c}
       \hline
       \bf Tasks  & \bf Zero-shot Evaluation & \bf After training with CAVPref \\
       \hline
       \multicolumn{3}{c}{\textit{VideoBench}} \\
       \hline
        Prior knowledge-based QA & 27.80 & \bf 34.65 \\
        Comprehension decision making & 38.21 & \bf 47.68 \\
        Video exclusive understanding & 32.48 & \bf 40.71 \\
        \hline
       \multicolumn{3}{c}{\textit{MVBench}}\\
       \hline
       Average Accuracy (on 20 tasks) & 34.10 & \bf 42.38 \\
       \hline
    \end{tabular}}
    \caption{Results of Video-LLaMA2 on VideoBench and MVBench.}
    \label{tab:other_benchmarks_supp}
\end{table*}

\begin{table*}[!t]
\centering
\renewcommand{\arraystretch}{1.5}
\resizebox{0.8\textwidth}{!}{%
\begin{tabular}{l|cccc|ccc|cc}
\hline
\textbf{Model} &
  \textbf{MCIT} &
  \textbf{ICIT} &
  \textbf{MVIT} &
  \textbf{MAIT} &
  \textbf{COT-Stitch} &
  \textbf{COT-Swap} &
  \textbf{CAT} &
  \textbf{MVT} &
  \textbf{MAT} \\
\hline
Video LLaMA-7B & 11.25 & 25.9 & 18.74 & 16.57 & 32.33 & 27.5 & 28.13 & 40.64 & 36.71\\
\bf Video LLaMA-13B & \bf 13.1 & \bf 27.41 & \bf 20.23 & \bf 17.81 & \bf 35.24 & \bf 29.81 & \bf 30.33 & \bf 42.08 & \bf 38.76\\
\hline
PandaGPT-7B & 6.24 & 21.1 & 14.19 & 11.93 & 30.86 & 25.22 & 27.84 & 34.36 & 30.21\\
\bf PandaGPT-13B & \bf 8.24 & \bf 22.71 & \bf 15.87 & \bf 13.47 & \bf 31.94 & \bf 26.44 & \bf 29.42 & \bf 36.05 & \bf 32.85\\
\hline
X-InstructBLIP-7B & 10.41 & 21.92 & 15.6 & 14.0 & 30.95 & 23.94 & 27.67 & 37.58 & 34.78\\
\bf X-InstructBLIP-13B & \bf 11.42 & \bf 24.85 & \bf 17.59 & \bf 15.6 & \bf 32.57 & \bf 26.18 & \bf 29.35 & \bf 39.31 & \bf 35.87\\
\hline
\end{tabular}%
}
\caption{Performance comparison with 7B vs 13B models.}
\label{tab:7b_vs_13b}
\end{table*}

\section{Additional Details on Training}
\label{training_details_appendix}

\subsection{Under-represented categories.} We observe a non-uniformity in the distribution of categories across the AVQA and MUSIC-AVQA datasets. Such skewness leads to overemphasis of some categories on which the model's predictions are biased (as shown in Fig. \ref{fig:category_distribution}). To mitigate such issues, we incorporate a robustness module in the proposed CAVPref (details in the main text).

\begin{figure*}[!t]
    \centering
    \includegraphics[width=0.9\textwidth]{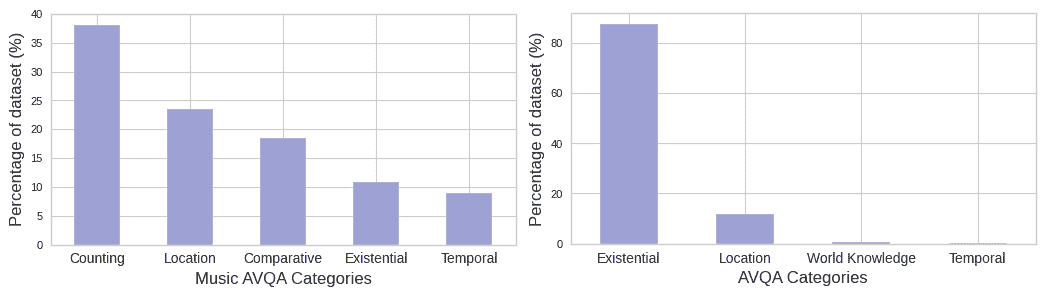}
    \caption{Distribution of different question categories across AVQA and MUSIC-AVQA datasets.}
    \label{fig:category_distribution}
\end{figure*}

\subsection{Proof for the final objective of CAVPref.}

\textbf{Theorem 1.} \textit{Considering KL divergence as the discrepancy measure between $Q$ and $P$, the closed-form objective becomes:}

\begin{equation}
    \mathcal{L}_{\text{closed-form}} = - \lambda \log \left( \mathbb{E}_{P}\Bigl[e^{\frac{\mathcal{L}}{\lambda}}\Bigr]\right)
\end{equation}

\noindent \textit{where $\lambda$ is a regularization hyperparameter.}

\noindent\textbf{Proof.} Considering the actual optimization problem:

\begin{equation}
    \underset{Q}{\max}\mathbb{E}_{Q}[\mathcal{L}]: \mathbb{D}_{KL}(Q||P) \leq \rho
\end{equation}

\noindent By method of Lagrangian multipliers, the problem becomes:

\begin{equation}
    \underset{Q}{\max}\mathbb{E}_{Q}[\mathcal{L}] - \lambda (\mathbb{D}_{KL}(Q||P) - \rho)
\end{equation}

\noindent Solving the saddle-point problem by taking partial derivative with respect to $Q$ and equating it to 0, we obtain:

\begin{gather}
    \frac{\partial}{\partial Q} \mathbb{E}_{Q}\Bigl[\mathcal{L} - \lambda \log \frac{Q}{P}\Bigr] = 0 \\ \nonumber
    Q^{*} \propto P e^{\frac{\mathcal{L}}{\lambda}}
\end{gather}

\noindent Since $Q^{*}$ is a probability distribution, we obtain:

\begin{equation}
    Q^{*} = \frac{P e^{\frac{\mathcal{L}}{\lambda}}}{Z}
\end{equation}

\noindent where $Z = \mathbb{E}_{P}\Bigl[e^{\frac{\mathcal{L}}{\lambda}}\Bigr]$ is a normalizing factor or partition function.

\noindent Substituting $Q^{*}$ back in the original objective, we obtain:

\begin{equation}
    \mathbb{E}_{Q}[\mathcal{L}] = \sum Q^{*} \mathcal{L} = \sum \frac{P e^{\frac{\mathcal{L}}{\lambda}}}{Z} \mathcal{L} = \frac{1}{Z}\mathbb{E}_{P}\Bigl[\mathcal{L}e^{\frac{\mathcal{L}}{\lambda}}\Bigr]
\end{equation}

\noindent Solving the dual problem by substituting the value of $Q^{*}$:

\begin{gather}
    \mathbb{D}_{KL}(Q^{*}||P) = \lambda \rho \\ \nonumber
    \mathbb{E}_{Q^{*}}\Bigl[\log\frac{Q^{*}}{P}\Bigr] = \lambda\rho \\ \nonumber
    \mathbb{E}_{Q^{*}}[\frac{\mathcal{L}}{\lambda} - \log Z] = \lambda \rho \\ \nonumber
    \frac{1}{\lambda}\mathbb{E}_{Q^{*}}[\mathcal{L}] - \log Z = \lambda \rho
\end{gather}

\noindent Therefore, the final closed-form objective is equivalent to minimizing:

\begin{gather}
    \mathcal{L}_{\text{closed-form}} = - \lambda \log Z \\ \nonumber
    \mathcal{L}_{\text{closed-form}} = - \lambda \log \left( \mathbb{E}_{P}\Bigl[e^{\frac{\mathcal{L}}{\lambda}}\Bigr]\right)
\end{gather}

\subsection{Simplification of the DPO objective.}

DPO objective is given as:

\begin{equation}
\footnotesize
    \mathcal{L}_{\text{DPO}} = - \mathbb{E}_{(x,y_w,y_l) \sim P}\Bigl[\log \sigma \Big(\beta \log \Big(\frac{\pi_{\theta}(y_w)}{\pi_{\text{ref}}(y_w)}\Big) - \beta \log \Big(\frac{\pi_{\theta}(y_l)}{\pi_{\text{ref}}(y_l)}\Big)\Big]
\end{equation}

Considering $f_w = \frac{\pi_{\theta}(y_w)}{\pi_{\text{ref}}(y_w)}$ and $f_l = \frac{\pi_{\theta}(y_l)}{\pi_{\text{ref}}(y_l)}$, putting $\sigma(x) = \frac{1}{1 + \exp(-x)}$the above equation can be rewritten and simplified as:

\begin{gather}
    \mathcal{L}_{\text{DPO}} = - \mathbb{E}_{(x,y_w,y_l) \sim P}\left[\log \left(\frac{1}{1 + exp\Big(-\log\Big(\frac{f_w}{f_l}\Big)^{\beta}\Big)}\right)\right] \\ \nonumber
    \mathcal{L}_{\text{DPO}} = - \mathbb{E}_{(x,y_w,y_l) \sim P}\left[\log \left(\frac{1}{1 + exp\Big(\log\Big(\frac{f_l}{f_w}\Big)^{\beta}\Big)}\right)\right]\\ \nonumber
    \mathcal{L}_{\text{DPO}} = - \mathbb{E}_{(x,y_w,y_l) \sim P}\left[\log \left(\frac{1}{1 + \Big(\frac{f_l}{f_w}\Big)^{\beta}}\right)\right]\\ \nonumber
    \mathcal{L}_{\text{DPO}} = \mathbb{E}_{(x,y_w,y_l) \sim P}\left[\log \left(1 + \Big(\frac{f_l}{f_w}\Big)^{\beta}\right)\right]
\end{gather}

\subsection{Pseudocode for CAVPref}

The training pseudocode for CAVPref is shown in \cref{algo:CAVPref}. We employ a multimodal DPO formulation and update the objective functions as outlined below.

\subsection{Results on other models}
In \cref{tab:mitigation_supp} we compare the performance of 7 other open source models upon employing supervised finetuning (SFT), DPO, and CAVPref. Experimental results demonstrate a steady boost in performance upon applying CAVPref across all the models over all 9 tasks. We note that the highest performance gains are observed in the modality dependency suite - as our proposed approach guides the models to ingest modality-specific information thereby making a holistic inference.     


\subsection{Results on other benchmarks}

We evaluate two different benchmarks, i.e., Video-Bench and MVBench before (zero-shot) and after training (following our proposed strategy - CAVPref) and report the values in Tab. \ref{tab:other_benchmarks_supp} (using Video-LLaMA2). We observed substantial improvements with our proposed training paradigm.

\section{Discussion on Bridging Networks}
\label{bridging_appendix}
Bridge networks are modules used to connect the modality-specific encoders with the LLM by transforming the information from multi-modal encoders' space to LLM embedding space. For instance, VAST \cite{vast} uses text converters as the most basic and simplest bridge. Macaw-LLM uses a customized bridge network with linear layers and cross-attention-based alignment modules. VideoLLaMA(-2), Bay-CAT, video-SALMONN and X-InstructBLIP use Q-former-based bridge networks, whereas ChatBridge uses a customized perceiver network shared across all the modalities. OneLLM uses a mixture of projection experts equipped with a modality routing module, and ImageBind-LLM uses sophisticated trainable bind networks as the bridging module. 



\section{Performance with Different Model Variants}
\label{model variants performance appendix}

We experiment with the 7B and 13B variants of VideoLLaMA, PandaGPT, and X-InstructBLIP (other models employ a single variant). Experimental results confirm the performance boost with the 13B variants. A key observation is increasing the model size from 7B to 13B doesn't help in obtaining significant gain in \compositionalsuite suite of tasks. We hypothesize that LLMs are not able to capture the attribute level binding information and often work as bag-of-word models. Tab. \ref{tab:7b_vs_13b} compares the two variants of the above-mentioned models.    



\section{More Related Works}
\label{related works appendix}

\paragraph{Audio-Visual QA datasets.}
Deep learning for video QA relies on diverse datasets such as MSRVTT-QA \cite{msrvtt}, and ActivityNet-QA \cite{activitynet}. MovieQA \cite{movieqa} and TVQA \cite{tvqa} add to the diversity of available scenario-specific datasets in this space. However, these datasets often focus on specific tasks and cannot amply evaluate the comprehensive reasoning capabilities of AVLLMs. Moreover, the majority of these datasets do not contain meaningful audio and QA pairs encompassing cross-modal understanding. To this end, we leverage three public audio-visual datasets AVQA \cite{avqadataset}, MUSIC-AVQA \cite{musicavqadataset} and AudioSet \cite{audioset} to form the QA pairs for all our tasks. These datasets can facilitate study on spatio-temporal reasoning for dynamic and long-term audio-visual scenes, complex audio-visual reasoning, multi-modal perception and granularity (\textit{existential}, \textit{location}, \textit{counting} etc.). In the face of a massive deluge of MLLMs, there is an acute shortage of benchmarks that can extensively evaluate the trustworthiness of these models. Our presented \ourbenchmark can bridge this gap by serving as a testbed to evaluate different dimensions of these models such as cross-modal comprehension, reasoning, and perception abilities.

\begin{algorithm*}
   \caption{PyTorch-style pseudocode for CAVPref.}
   \label{algo:CAVPref}
   
    \definecolor{codeblue}{rgb}{0.25,0.5,0.5}
    \definecolor{codekw}{rgb}{0.85, 0.18, 0.50}
    \newcommand{\algofontsize}{9.0pt}
    \lstset{
      backgroundcolor=\color{white},
      basicstyle=\fontsize{\algofontsize}{\algofontsize}\ttfamily\selectfont,
      columns=fullflexible,
      breaklines=true,
      captionpos=b,
      commentstyle=\fontsize{\algofontsize}{\algofontsize}\color{codeblue},
      keywordstyle=\fontsize{\algofontsize}{\algofontsize}\color{codekw},
    }
\begin{lstlisting}[language=python]
# pi_yw_logps: winning response logprobs (policy)
# pi_yl_logps: losing response logprobs (policy)

# pi_yw_Vw_logps: winning response with correct visual logprobs (policy)
# pi_yw_Vl_logps: winning response with incorrect visual logprobs (policy)

# pi_yw_Aw_logps: winning response with correct audio logprobs (policy)
# pi_yw_Al_logps: winning response with incorrect audio logprobs (policy)

# ref_yw_logps: winning response logprobs (reference model)
# ref_yl_logps: losing response logprobs (reference model)

# ref_yw_Vw_logps: winning response with correct visual logprobs (reference model)
# ref_yw_Vl_logps: winning response with incorrect visual logprobs (reference model)

# ref_yw_Aw_logps: winning response with correct audio logprobs (reference model)
# ref_yw_Al_logps: winning response with incorrect audio logprobs (reference model)

# beta_y, beta_V, beta_A: policy regularization coefficients

# lambda_y, lambda_V, lambda_A: robustness coefficients

def CAVPref:
    # linguistic component (Eq. 1)
    pi_logratios_y = pi_yw_logps - pi_yl_logps
    ref_logratios_y = ref_yw_logps - ref_yl_logps
    
    loss_y = F. logsigmoid ( beta_y * ( pi_logratios - ref_logratios ))

    # visual component (Eq. 2)
    pi_logratios_V = pi_yw_Vw_logps - pi_yw_Vl_logps
    ref_logratios_V = ref_yw_Vw_logps - ref_yw_Vl_logps
    
    loss_V = F. logsigmoid ( beta_V * ( pi_logratios_V - ref_logratios_V ))

    # audio component (Eq. 3)
    pi_logratios_A = pi_yw_Aw_logps - pi_yw_Al_logps
    ref_logratios_A = ref_yw_Aw_logps - ref_yw_Al_logps
    
    loss_A = F. logsigmoid ( beta_A * ( pi_logratios_A - ref_logratios_A ))

    # Eqs. 5 and 6 combined
    CAVPref_loss = - (lambda_y * torch.log(torch.mean(torch.exp(loss_y / lambda_y))) + lambda_V * torch.log(torch.mean(torch.exp(loss_V / lambda_V))) + lambda_A * torch.log(torch.mean(torch.exp(loss_A / lambda_A))))
            
    return CAVPref_loss
\end{lstlisting}
\end{algorithm*}

\section{Implementation Details}
\label{implementation details appendix}

For open-source models, we follow their default best inference settings and hyperparameters. To evaluate GPT-4o, Gemini 1.5 Pro we utilize their official APIs. Full videos are directly passed to Gemini 1.5 Pro, as its API (using Google Cloud vertexai framework) inherently supports video inputs. For each model under evaluation, we generate responses to the questions independently and without retaining the chat history. For evaluating all open-source AVLLMs on \ourbenchmark tasks, we use 1 A100 GPU. For training the open-source AVLLMs on \ourbenchmark tasks, we utilize 8 A100 GPUs and follow their respective training implementation details.

\section{Common Sense Reasoning}
\label{common sense reasoning appendix}
\cref{fig:common_sense} shows that the current AVLLMs \textit{lack} commonsense reasoning. There is evidence in animal study \cite{hoffman2017dogs} that it is a natural tendency of a dog to bark at an unknown cat. In this example (refer to video 7min 50sec) most AVLLMs fail to infer this and opts for incorrect response underlying their lack of commonsense reasoning skills.


\section{More qualitative Examples}
\label{qual results appendix}

We share more qualitative samples from each task in Fig. \ref{fig:qual_mcit_supp} - \ref{fig:qual_mat_supp}. As can be seen, closed-source models demonstrate an overall better performance compared to open-source counterparts with GPT-4o being the strongest performer across the majority of the tasks. We note that upon employing CAVPref, the responses of the AVLLMs improve as they tend to make fewer mistakes on the same QA pairs - which underlines the effectiveness of our proposed approach over DPO.      


\section{Failure Cases}
\label{failure cases appendix}

Fig. \ref{fig:failure_supp} illustrates the failure cases of our mitigation approach CAVPref while used with video-SALMONN, Video-LLaMA2, and Bay-CAT. In the first case, the models are unable to differentiate between `violin' in the video and `viola' in the audio since they are semantically closely associated. Therefore, although this is a task of MVIT, the models are unable to pick the correct answer, i.e., `(E) None of the above'. In the second case, the models are unable to see the speaker (on the left) who is facing their back (i.e., their face is not visible). Therefore, they are unable to understand that the correct answer, i.e., `left' which is not present in the set of options (MCIT task), and thus the ideal response would be `(E) None of the above'.

\section{Supplementary Video Examples}
\label{supp video appendix}
In the supplementary video, we add qualitative examples for each of the tasks of \ourbenchmark for each model. We find the MLLMs to produce free-form responses on many occasions. We employ our two-stage choice extraction strategy as explained in \cref{choice extraction strategy appendix} to obtain the AVLLMs responses and process them accordingly. The use of headphones is recommended for a better audio-visual QA experience.

\section{Societal Impact}
\label{societal impact appendix}
In this work, we perform an extensive analysis of existing state-of-the-art AVLLMs to study their failure modes. Our study reveals that models lack sufficient audio-visual comprehension skills and most often fail to address scenarios that require common sense reasoning. We believe our work can be useful to the community and our findings can reveal the potential threats associated with deploying these models in real-time or accuracy-critical setups. The users must recognize these limitations in the new generation models and proceed with caution, especially in scenarios where the precision and neutrality of results hold significant importance. Users are encouraged to thoroughly scrutinize and validate the outputs of the model to avoid the possibility of disseminating inaccurate information. We employ the existing public datasets to curate the benchmark and we don't collect or use any personal/human subject data without their consent during our data preparation and experiments stages.

\section{Human Study Details}
\label{human study appendix}
We conducted a small study involving 20 individuals to assess the difficulty of our proposed benchmark and estimate the upper bound for the tasks proposed. The user study protocol was approved by the Institutional Review Board and we do not collect, share or store any personal information of the participants.

\customsubsection{Data Collection and Quality Control}

We form Audio-Visual QAs in the format of multiple-choice problems for each task. A problem $P_i$ corresponds to $\left(Q_i, C_i, V_i, A_i, R_i\right)$. $Q_i$ denotes the question, $C_i$ represents a set with $n(2 \leq n \leq 5)$ choices $c_1, c_2, \ldots, c_n, V_i,$ and $A_i$ represents the input video and the audio respectively, and $R_i$ is the correct response. The number of choices varies depending on the task. For each task, we first prepare up to $\sim 5$ different question templates to ensure sufficient variations in the question formats. We carefully choose the questions from one of these templates. We add more details on the QA pair formation in the supplementary.

\begin{figure*}[!t]
    \centering
    \includegraphics[width=\textwidth]{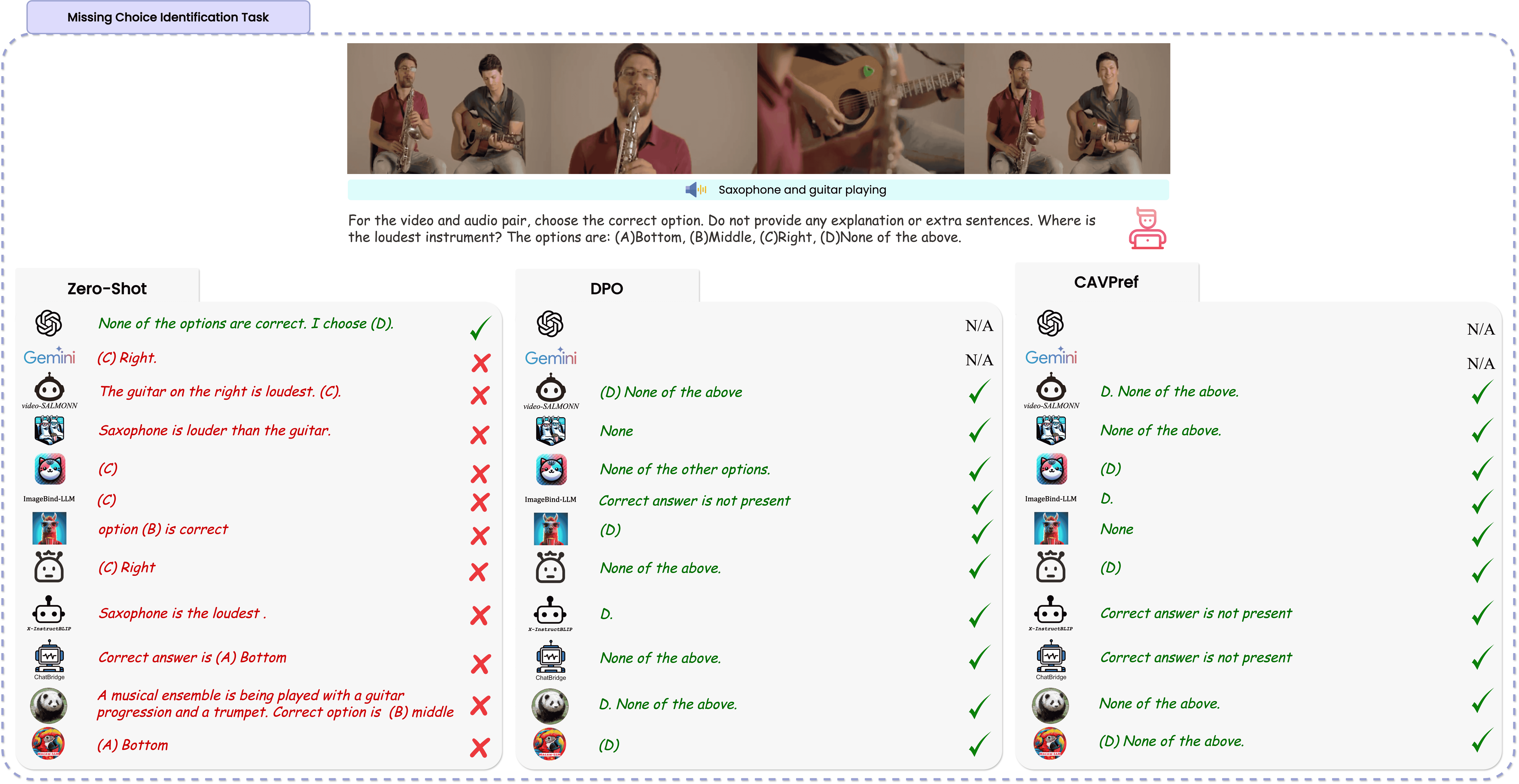}
    \caption{Performance comparison of all open source models on MCIT task under ZS, DPO and CAVPref.}
    \label{fig:qual_mcit_supp}
\end{figure*}

We collect the AV samples from benchmark datasets AVQA, MUSIC-AVQA, and AudioSet. While the QA pairs for AVQA and MUSIC-AVQA are adapted directly from those datasets, for AudioSet we obtain the QA pairs from a pre-designed template (Tab. \ref{tab:task_wise_propmpts}). Finally, while forming the mismatched pairs, we follow a semi-automated (heuristics + look-up table) approach. We apriori create a dictionary of mismatched pairs by careful manual inspection to ensure that the corresponding audio-visual pairs have no association between them. To further validate, we manually investigate randomly chosen 500 samples from each of the axes of evaluation. We compute the spearman correlation coefficient between the human labels and our curated data on those samples and we obtain a mean score of 0.979 (p < 0.05) - indicating a significantly strong correlation.

Kindly note that samples from AudioSet were only collected for the compositional understanding tasks. For the adversarial attack and missing modality tasks, the samples are curated from the AVQA and the MUSIC-AVQA datasets. Moreover, employing AudioSet for both fine-grained and coarse-grained audio-visual tasks has been explored by the community \cite{chowdhury2024meerkat, mercea2022audio, parida2020coordinated}.

AudioSet contains real-world samples under in-the-wild settings where we ensure that the constituent modalities (audio and visual) are aligned by adhering to the following strategy. We utilize the CLIP \cite{clip} and CLAP \cite{clap} scores by calculating $T_{\text{sim}} = \mathcal{S}_{\text{CLIP}} \mathcal{S}_{\text{CLAP}}^{T}$, where $\mathcal{S} \in \mathbb{R}^{N \times N}$ and denotes the pairwise cross-modal similarity scores for a batch of size $N$. The CLIP similarity is calculated between the chosen visual and the audio class label, similarly, the CLAP score is calculated between the audio class label and the audio snippet. The text modality acts as the bridging modality in this case. Note the range of the scores is normalized between [0,1] with 0 being the lowest. We don’t consider the samples having a $T_{\text{sim}}$ score of less than 0.70 to ensure a strong association between the two modalities. Notably, CLIP + CLAP based selection approach has been employed and accepted in the audio-visual community in recent literature \cite{chowdhury2024melfusion, chowdhury2024meerkat}.

\begin{figure*}[!t]
    \centering
    \includegraphics[width=\textwidth]{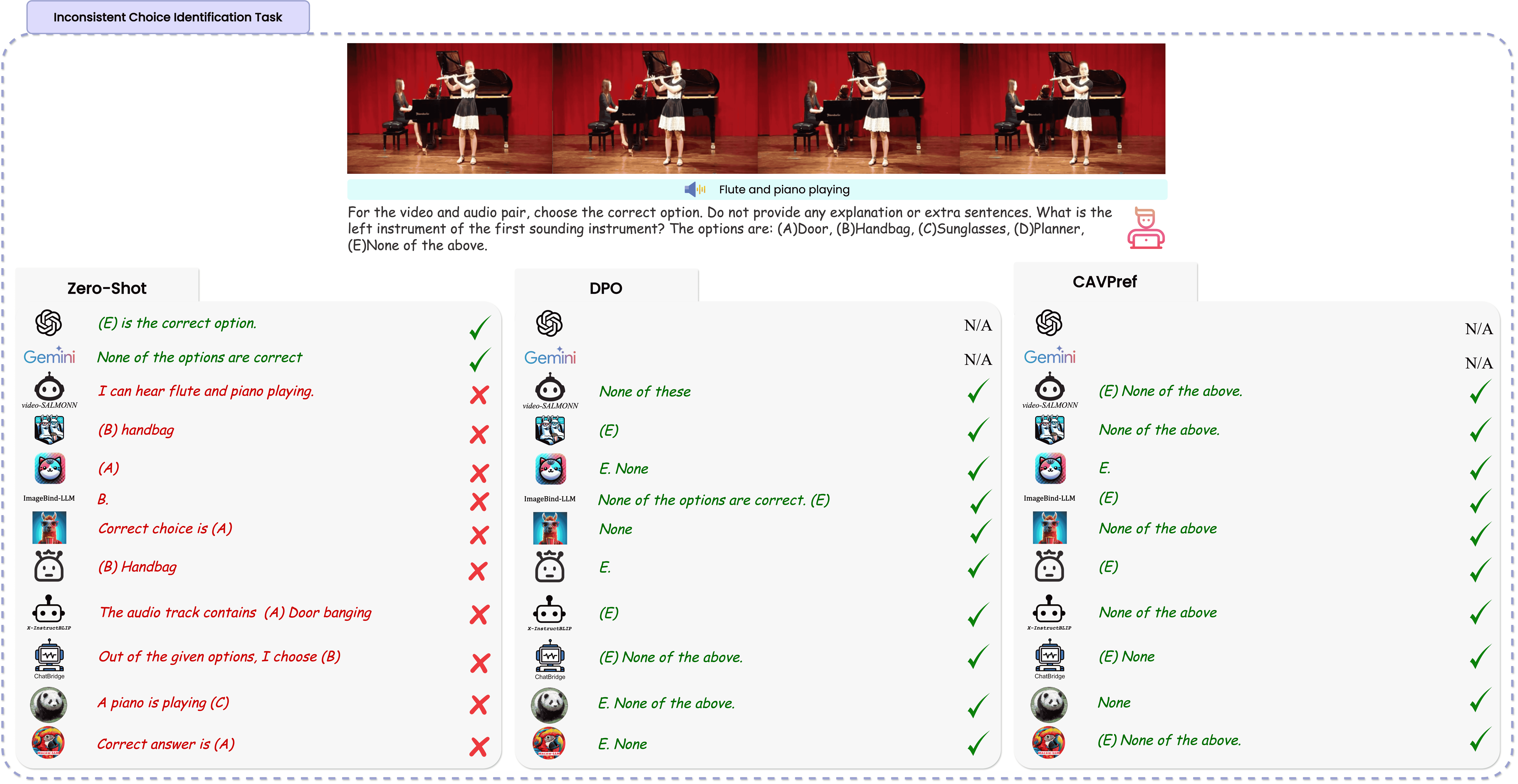}
    \caption{Performance comparison of all open source models on ICIT task under ZS, DPO, and CAVPref.}
    \label{fig:qual_icit_supp}
\end{figure*}

\begin{figure*}[!t]
    \centering
    \includegraphics[width=\textwidth]{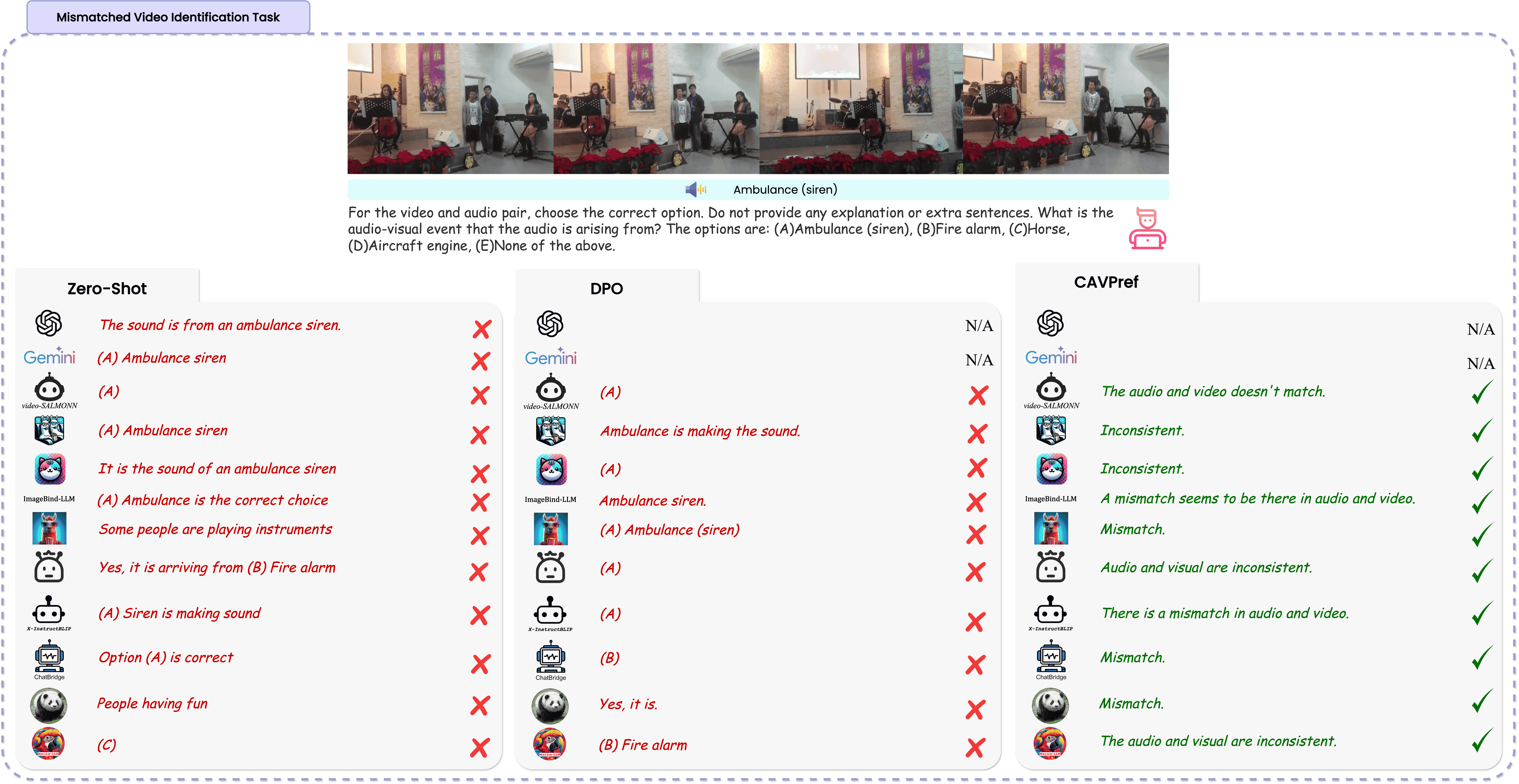}
    \caption{Performance comparison of all open source models on MVIT task under ZS, DPO, and CAVPref.}
    \label{fig:qual_mvit_supp}
\end{figure*}

\begin{figure*}[!t]
    \centering
    \includegraphics[width=\textwidth]{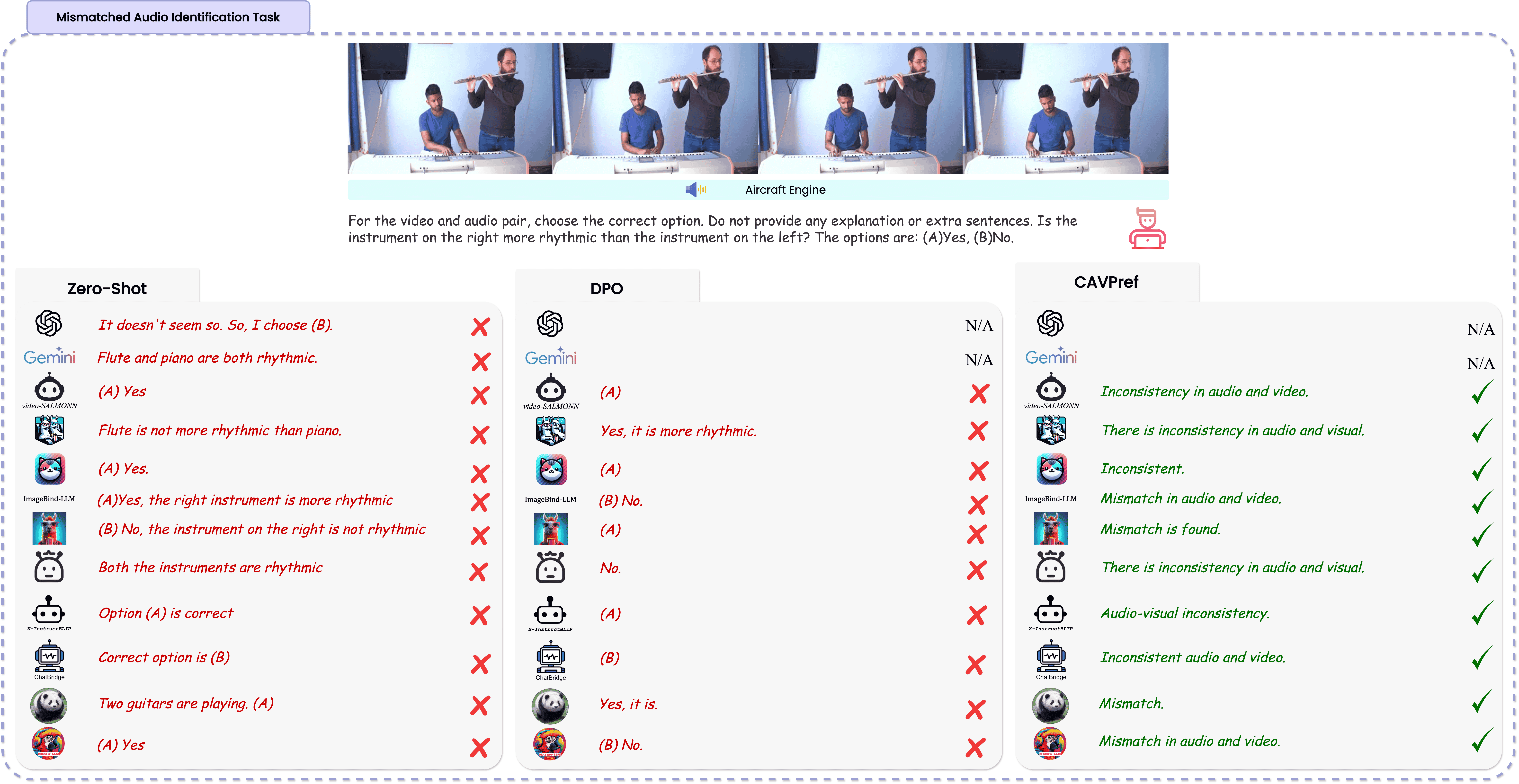}
    \caption{Performance comparison of all open source models on MAIT task under ZS, DPO, and CAVPref.}
    \label{fig:qual_mait_supp}
\end{figure*}

\begin{figure*}[!t]
    \centering
    \includegraphics[width=\textwidth]{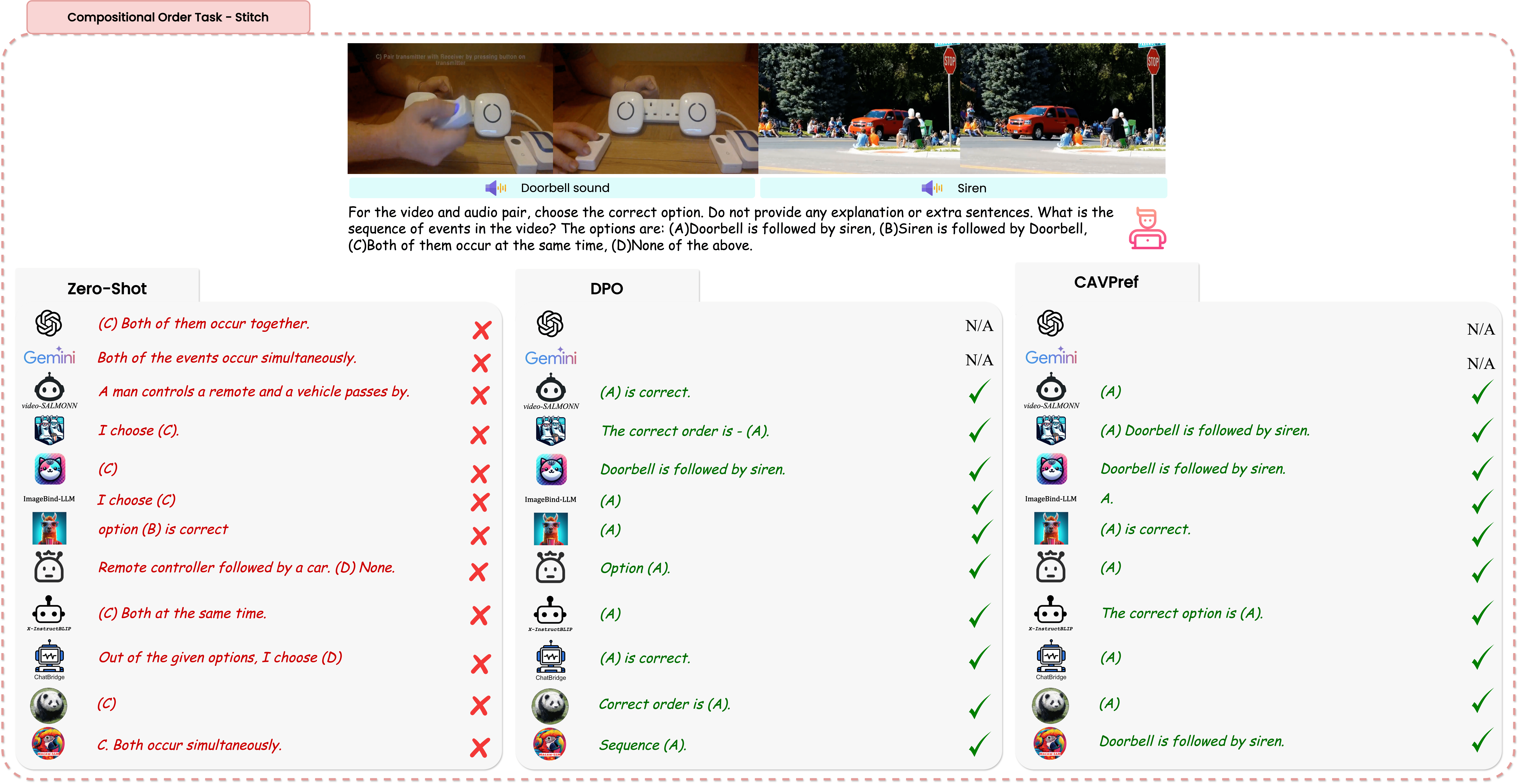}
    \caption{Performance comparison of all open source models on COT-Stitch task under ZS, DPO, and CAVPref.}
    \label{fig:qual_cot_stitch_supp}
\end{figure*}

\begin{figure*}[!t]
    \centering
    \includegraphics[width=\textwidth]{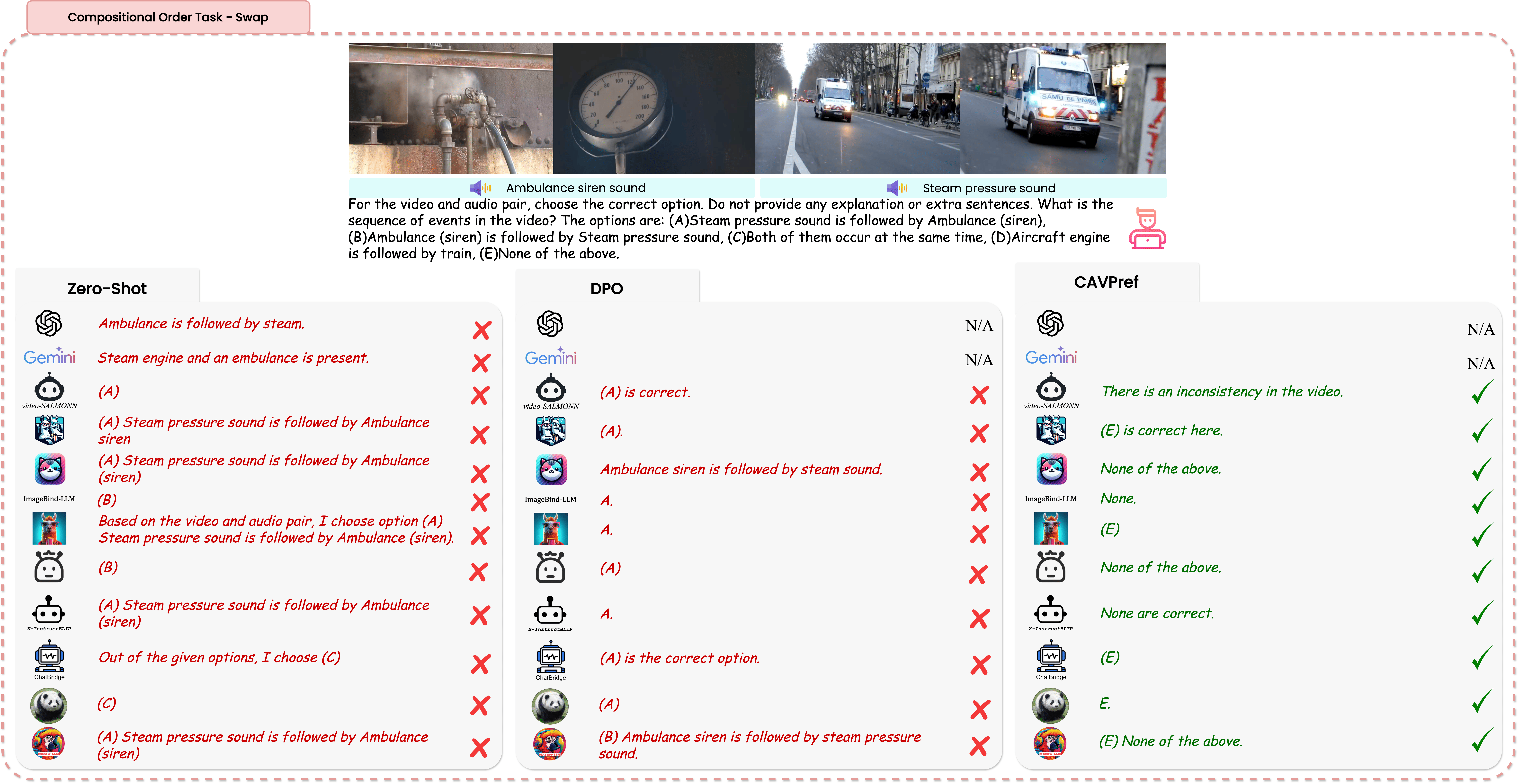}
    \caption{Performance comparison of all open source models on COT-Swap task under ZS, DPO, and CAVPref.}
    \label{fig:qual_cot_swap_supp}
\end{figure*}

\begin{figure*}[!t]
    \centering
    \includegraphics[width=\textwidth]{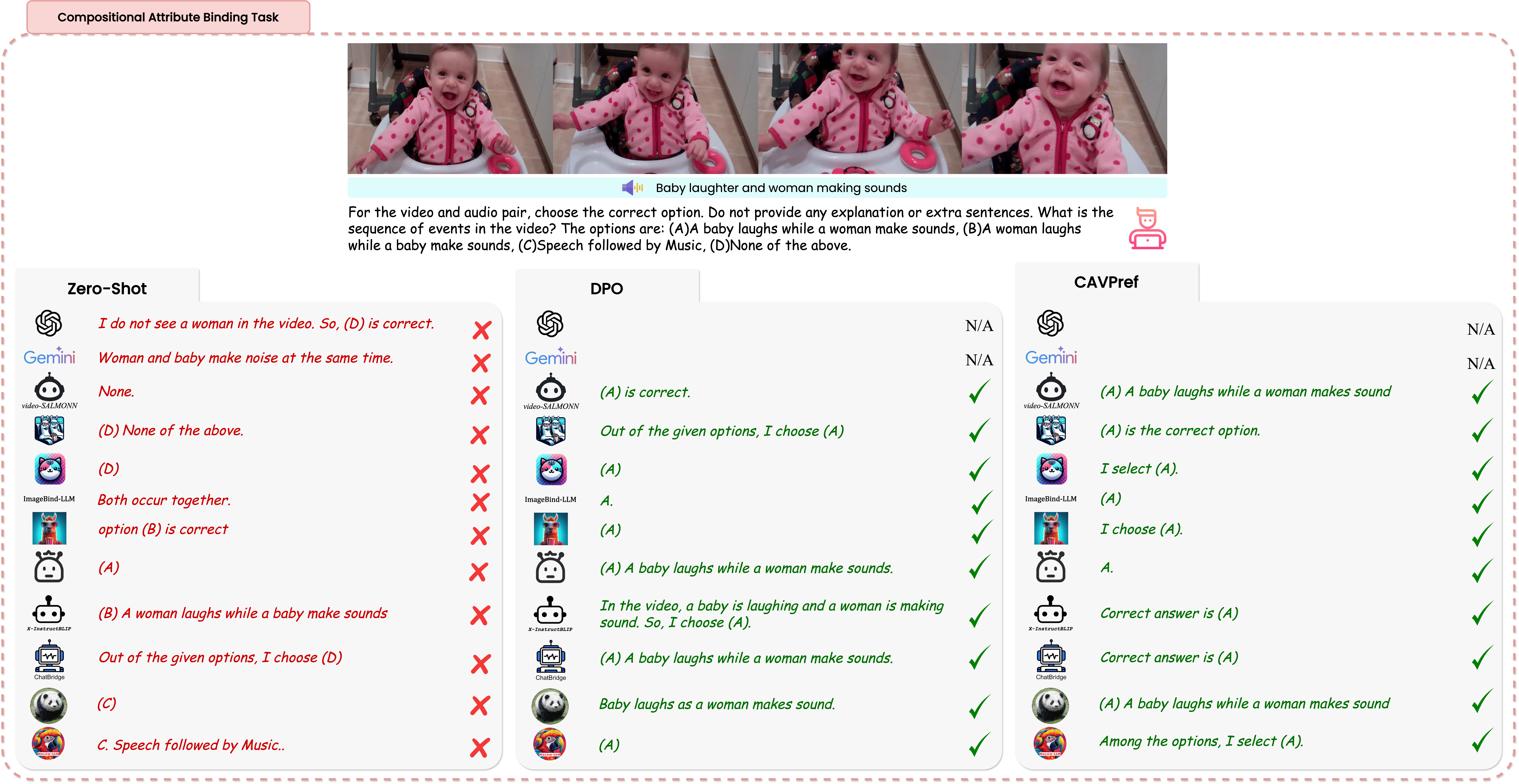}
    \caption{Performance comparison of all open source models on CAT task under ZS, DPO, and CAVPref.}
    \label{fig:qual_cat_supp}
\end{figure*}

\begin{figure*}[!t]
    \centering
    \includegraphics[width=\textwidth]{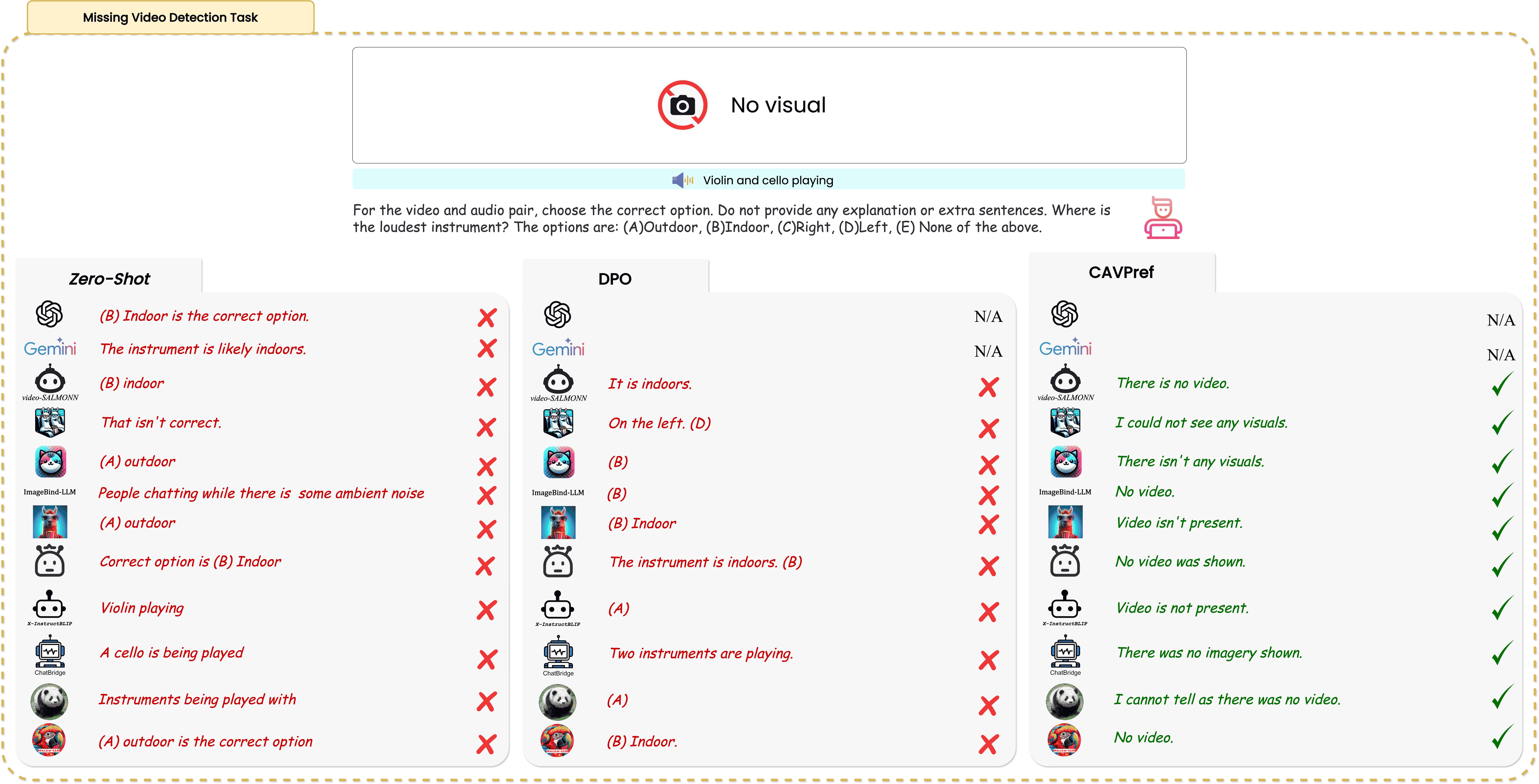}
    \caption{Performance comparison of all open source models on MVT task under ZS, DPO, and CAVPref.}
    \label{fig:qual_mvt_supp}
\end{figure*}

\begin{figure*}[!t]
    \centering
    \includegraphics[width=\textwidth]{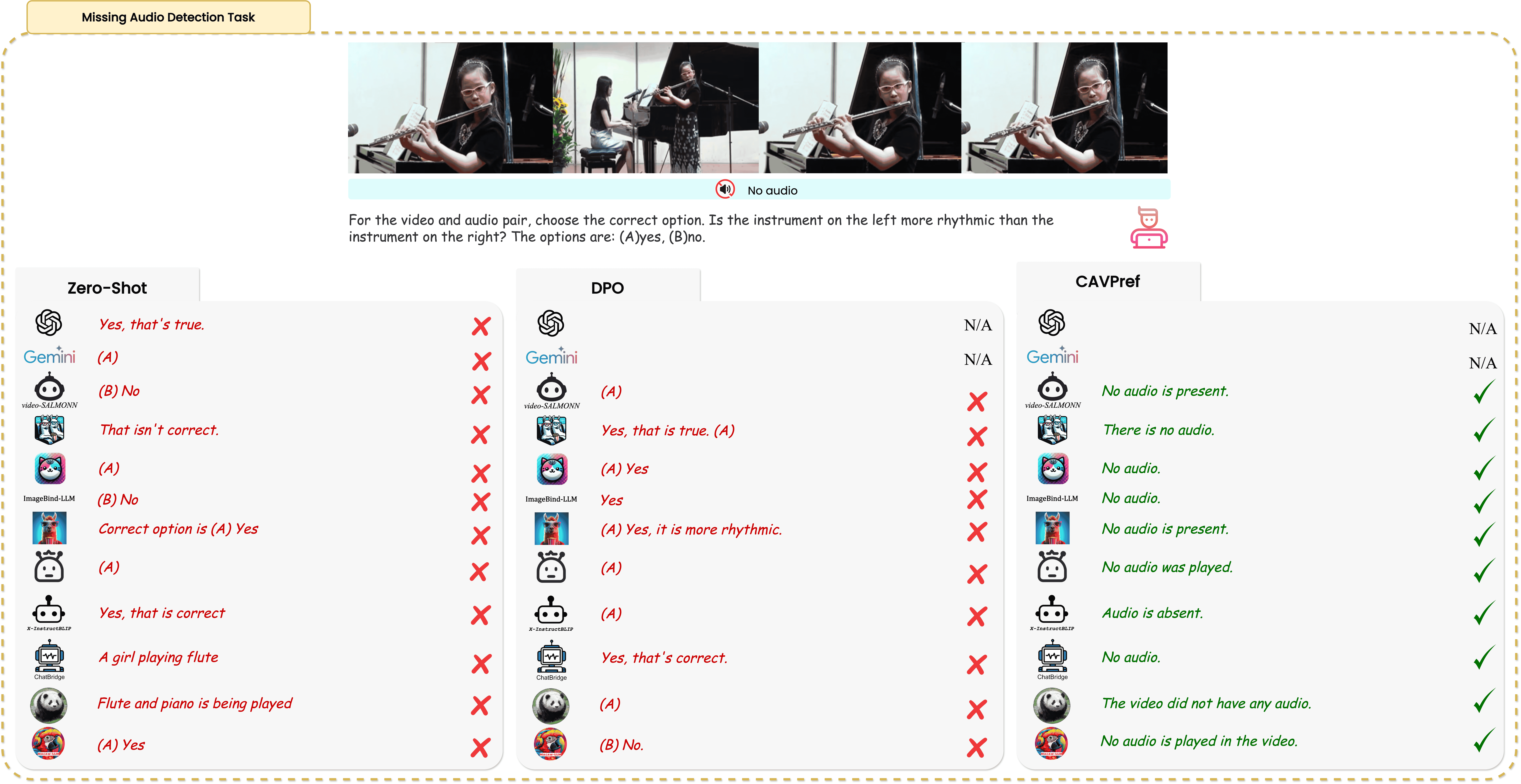}
    \caption{Performance comparison of all open source models on MAT task under ZS, DPO, and CAVPref.}
    \label{fig:qual_mat_supp}
\end{figure*}

\begin{figure*}[!t]
    \centering
    \includegraphics[width=0.5\textwidth]{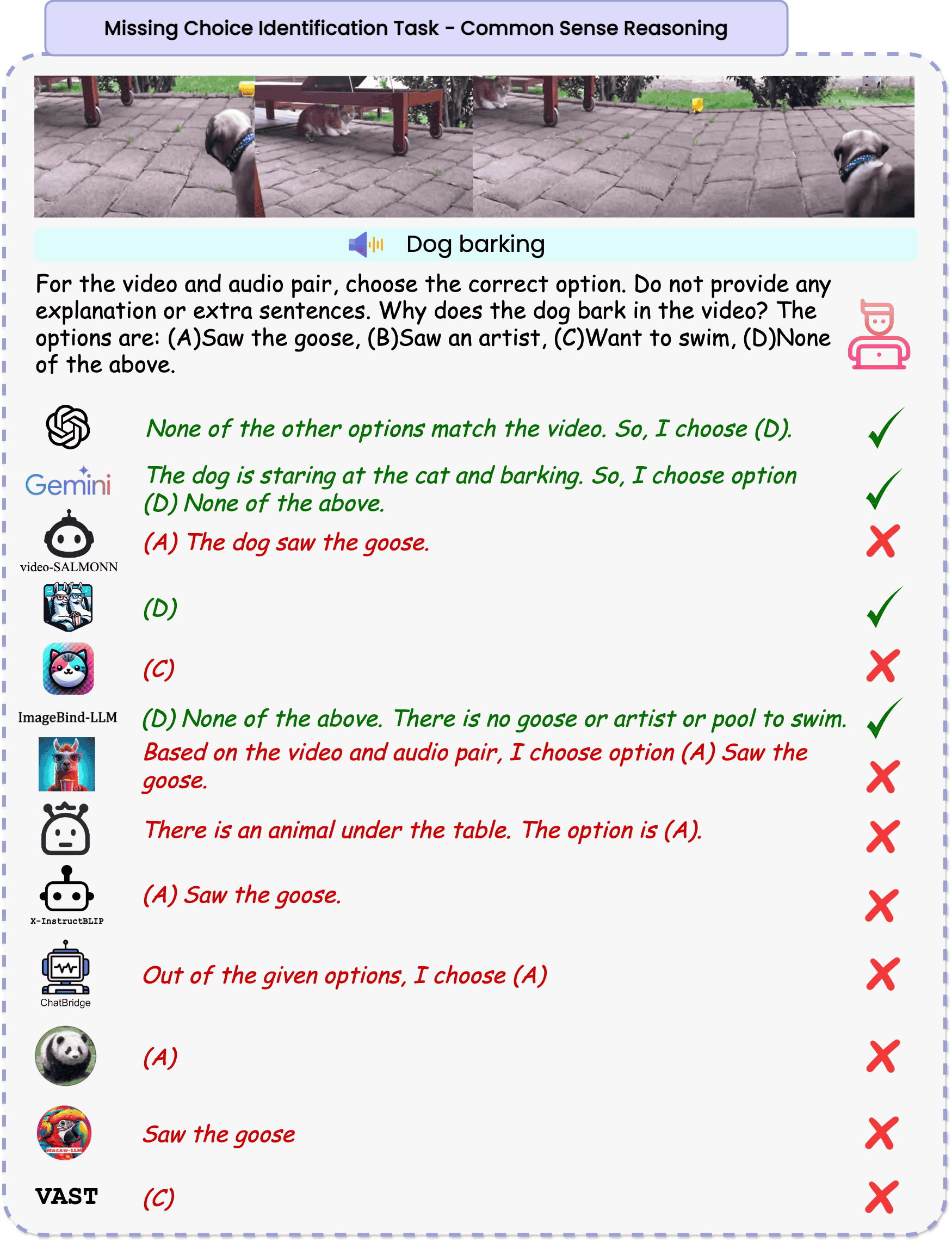}
    \caption{Example scenario depicting that most AVLLMs struggle in Common Sense Reasoning.}
    \label{fig:common_sense}
\end{figure*}

\begin{figure*}[ht]
    \centering
    \includegraphics[width=0.45\textwidth]{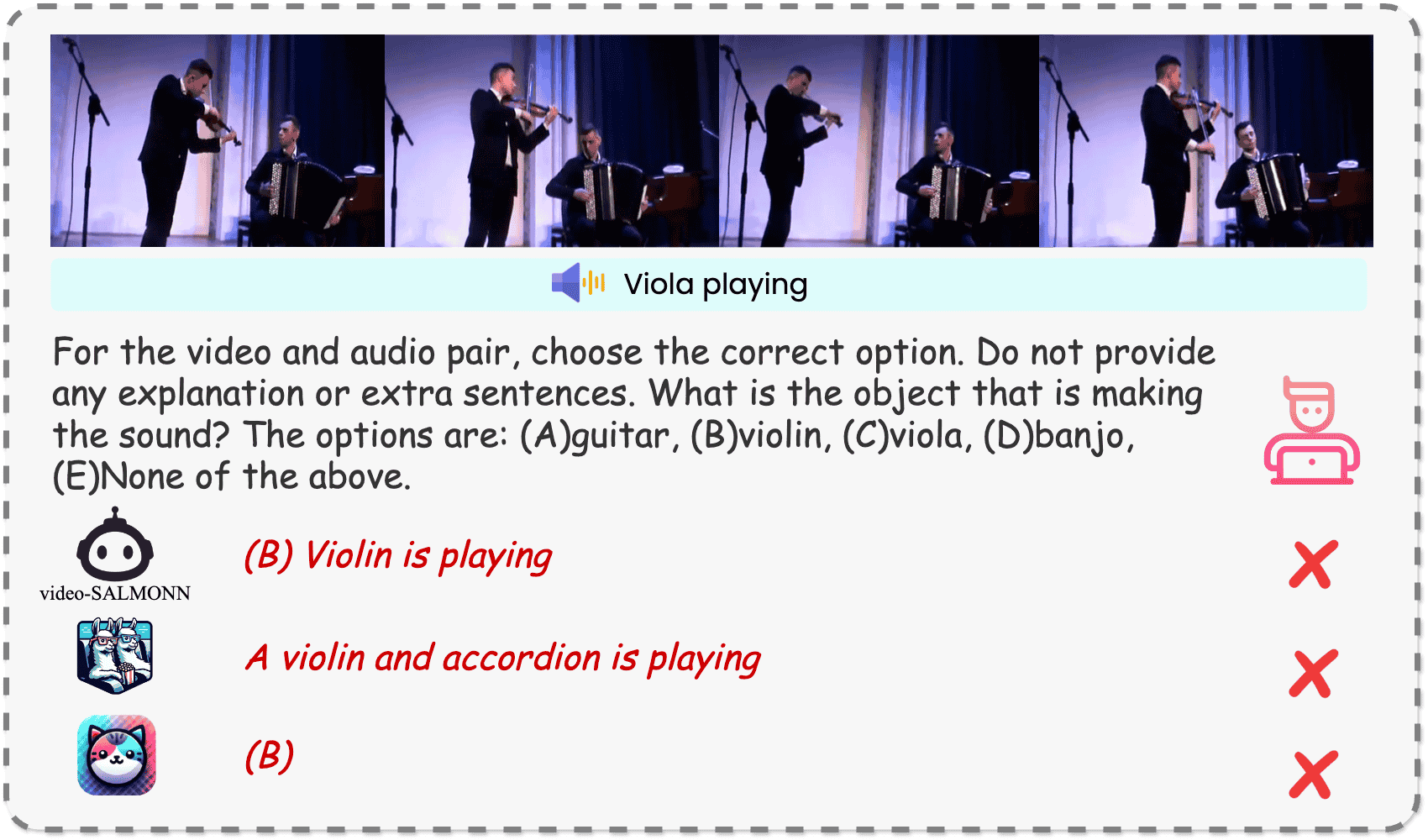} 
    \hfill
        \includegraphics[width=0.45\textwidth]{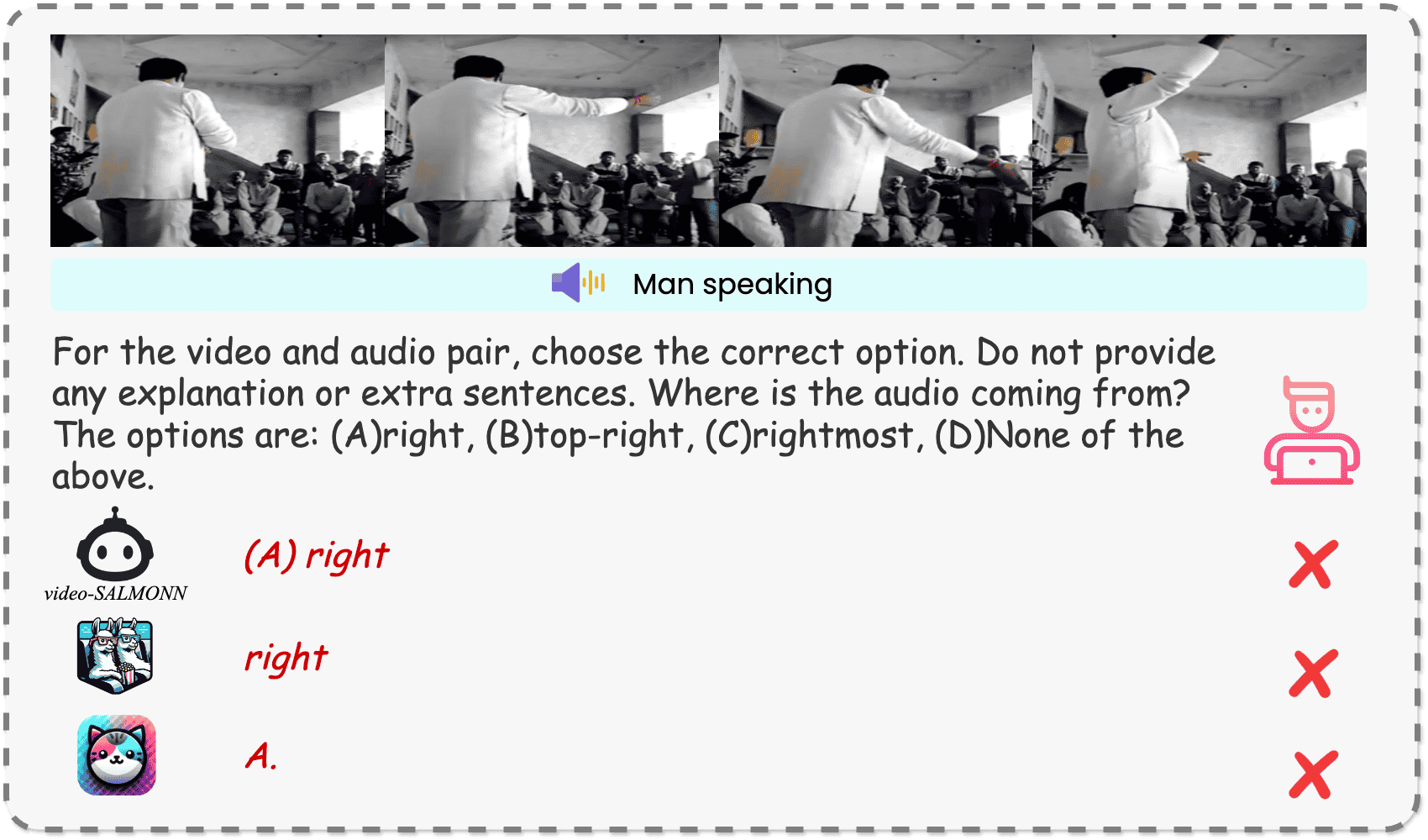} 
    \caption{Failure cases of video-SALMONN, Video-LLaMA2, and Bay-CAT after training with CAVPref.}
    \label{fig:failure_supp}
\end{figure*}

\end{document}